\documentclass[runningheads]{llncs}

\usepackage{eccv}

\usepackage{eccvabbrv}

\usepackage{graphicx}
\usepackage{booktabs}

\usepackage[accsupp]{axessibility}  

\definecolor{eccvblue}{rgb}{0.21,0.49,0.74}
\usepackage[breaklinks,colorlinks,citecolor=eccvblue,linkcolor=eccvblue,urlcolor=eccvblue]{hyperref}
\usepackage{orcidlink}  

\usepackage{fontawesome5}
\usepackage{pifont}
\usepackage{array}
\usepackage{makecell}
\usepackage{subcaption}
\usepackage{adjustbox}
\usepackage{float}
\usepackage{xspace}
\usepackage[normalem]{ulem}
\usepackage[table]{xcolor}
\usepackage{placeins}

\newcommand{\xmark}{\ding{55}}
\newcommand{\cmark}{\ding{51}}
\newcommand{\best}[1]{\textbf{#1}}

\newcommand{\method}{MMDiff\xspace}
\definecolor{row}{RGB}{235, 245, 251}

\newcommand{\papertitle}{\method: Extending Diffusion Transformers for Multi-Modal Generation}

\newcommand{\maketitlesupplementary}{
  \begin{center}
  {\Large\bfseries \papertitle \par}
  \vspace{0.3cm}
  {\large Supplementary Material\par}
  \end{center}
  \vspace{0.6cm}
}

\newlength{\cellw}
\newlength{\rowh}
\newcommand{\img}[1]{%
  \begin{minipage}[c][\rowh][c]{\cellw}
    \centering
    \includegraphics[width=\linewidth,height=\rowh,keepaspectratio]{#1}
  \end{minipage}
}

\begin{document}

\title{\papertitle}
\titlerunning{\method}

\author{Yagmur Akarken\inst{1} \and
Orest Kupyn\inst{1} \and
Christian Rupprecht\inst{1}}

\authorrunning{Y.~Akarken et al.}

\institute{University of Oxford, Visual Geometry Group}

\maketitle
\begin{center}
\includegraphics[width=\textwidth]{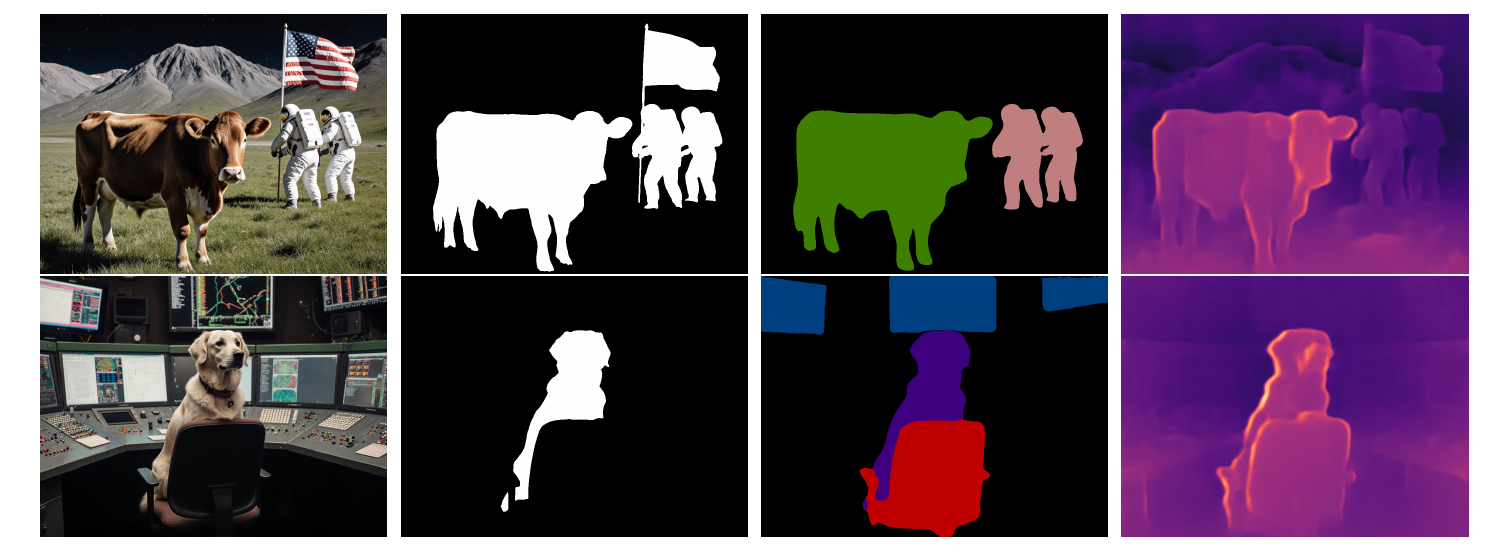}
    \vspace{-0.15cm}
   \captionsetup{type=figure}
   \captionof{figure}{
   \textbf{\method.}
   We present a method to decode multiple modalities from a frozen Diffusion Transformer. We show that information in DiT is distributed across the denoising trajectory, and introduce multi-timestep feature fusion to capture the complete perceptual information encoded across them. Lightweight decoder heads predict semantic segmentation, saliency, and depth with no modifications to the generative model.
   }
   \label{fig:teaser}
\end{center}

\begin{abstract}
Diffusion transformers have demonstrated remarkable generative capabilities, yet the rich perceptual representations computed across their denoising trajectory are discarded once the content is rendered. We present \method, a framework that transforms a frozen diffusion transformer into a multi-modal generative system that jointly produces images alongside any combination of dense perceptual modalities using lightweight decoder heads. Our central finding is that perceptual information is temporally distributed along the denoising trajectory, and that multi-timestep feature fusion with spatially varying aggregation weights is essential, improving semantic segmentation results by up to 28.7\% mIoU over single-timestep extraction. We further adopt concept-driven attention extraction for interpretable spatial guidance, and show that frozen diffusion features are competitive with and complementary to state-of-the-art encoders such as DINOv3. By training only lightweight decoder heads on a frozen backbone, we achieve strong performance in semantic segmentation, salient object detection, and depth estimation, and demonstrate that this framework enables effective synthetic data generation at scale. Code, models, and results are available \href{https://yagmurakarken.github.io/mmdiff/}{\texttt{here}}.
\end{abstract}
\section{Introduction}
\label{sec:intro}
Image and video generators are often not used solely to generate visual data but also require additional information about the image, such as segmentation or depth, for use in downstream applications. 
Diffusion transformers process thousands of tokens across dozens of layers and denoising timesteps to generate a single image or video, yet all intermediate representations are discarded once the final content is rendered.  Separate task-specific models are then used to extract semantic, geometric, and structural information from the generated pixels, thereby re-extracting information that may already have been computed during generation.
Extending diffusion models to predict new perceptual modalities has clear practical value. A model that generates images while simultaneously predicting their semantic labels, depth maps, or saliency masks can, for example, produce annotated training data at scale, thereby eliminating the need for expensive manual annotation. Similarly, such a pipeline is equally valuable for video and 3D generation, where geometric and semantic understanding must be consistent with the generated content. In both settings, the key challenge is the same: how to best extract perceptual information from a frozen generative backbone.
Several works have explored this direction. DatasetDM~\cite{wu2023datasetdmsynthesizingdataperception} and DiffuMask~\cite{wu2024diffumasksynthesizingimagespixellevel} extract annotations from Stable Diffusion's U-Net \cite{rombach2022high} during generation. In parallel, recent methods extend diffusion backbones to jointly predict scene geometry and video~\cite{zhang2024world,yang2025orchid,wang20254d}. However, all of these approaches extract features from the final denoised latent or the last denoising timestep, discarding all intermediate representations computed during generation.
The shift toward transformer-based diffusion models makes this particularly important. State-of-the-art models like FLUX~\cite{labs2023flux}, Stable Diffusion 3~\cite{esser2024sd3}, and PixArt-$\alpha$~\cite{chen2023pixart} adopt the Diffusion Transformer (DiT) architecture~\cite{peebles2023scalable}. Unlike U-Nets~\cite{unet}, which organize representations hierarchically through spatial downsampling, DiTs process all tokens at uniform resolution throughout the network~\cite{dosovitskiy2020image}. Without spatial hierarchies, perceptual information is spread more uniformly across layers and timesteps rather than concentrated at specific scales, meaning the final denoised latent alone is insufficient for dense prediction.

In this work, we introduce \textbf{\method}, a multi-modal diffusion framework that decodes any combination of dense perceptual modalities from a frozen diffusion transformer. Our central finding is that multi-timestep feature fusion is essential: aggregating features across the denoising trajectory yields significantly richer representations than extracting from any single timestep. In our experiments, multi-timestep fusion improves segmentation by +28.7\% mIoU over single-timestep extraction. We further leverage concept-driven attention extraction~\cite{helbling2025conceptattention} for targeted spatial guidance, and show that discriminative features can optionally complement diffusion representations~\cite{siméoni2025dinov3}. By training only lightweight decoder heads from a frozen DiT~\cite{labs2023flux} backbone, we achieve strong performance across semantic segmentation, salient object detection, and depth estimation. As a downstream application, this enables automatic label extraction for synthetic data generation.
We summarize our contributions as follows:
\begin{itemize}
\item We present \textbf{\method}, a framework for extending a frozen DiT to predict any dense perceptual modality using lightweight decoder heads, without modifying the generative backbone.
\item We demonstrate that multi-timestep feature fusion is essential for dense prediction from DiTs, and that single-timestep extraction, as used by all prior works, discards highly valuable information encoded during generation. We further enrich the extracted features using concept-driven attention extraction~\cite{helbling2025conceptattention}, which provides richer spatial guidance than averaged cross-attention.
\item We show, for the first time, that frozen diffusion transformer features achieve comparable performance to discriminative representations such as DINO~\cite{siméoni2025dinov3}, and that the two are complementary, demonstrating an effective combination of generative and discriminative features.
\item \method enables effective synthetic data generation: a single frozen model simultaneously generates images and extracts high-quality annotations, substantially outperforming prior U-Net-based methods~\cite{wu2023datasetdmsynthesizingdataperception, wu2024diffumasksynthesizingimagespixellevel, nguyen2023datasetdiffusion}.
\end{itemize}

\section{Related Work}
\label{sec:related}

\paragraph{Diffusion Transformer Architectures.}

Diffusion models~\cite{ho2020denoising, song2020score, dhariwal2021diffusion} have transitioned from U-Net~\cite{unet} to transformer-based architectures for improved scalability and generation quality. Latent diffusion models~\cite{rombach2022high} enable efficient high-resolution synthesis by operating in a compressed latent space. DiT~\cite{peebles2023scalable} demonstrated that transformers scale more efficiently than U-Nets for image generation, processing all patches with equal capacity through self-attention rather than hierarchical downsampling. FLUX~\cite{labs2023flux} extends this with a dual-stream architecture: 19 double-stream blocks for text-image interaction followed by 38 single-stream blocks for unified processing. Unlike U-Nets, which create semantic hierarchies through spatial downsampling, FLUX maintains 1024 image tokens of dimension 3072 across all layers. Unlike U-Nets, semantic specialization in DiTs arises from attention patterns across layers and timesteps rather than from spatial hierarchies. This fundamentally changes how perceptual information is organized, requiring new strategies to extract it effectively.

\begin{figure*}[t]
  \centering
  \includegraphics[width=\linewidth]{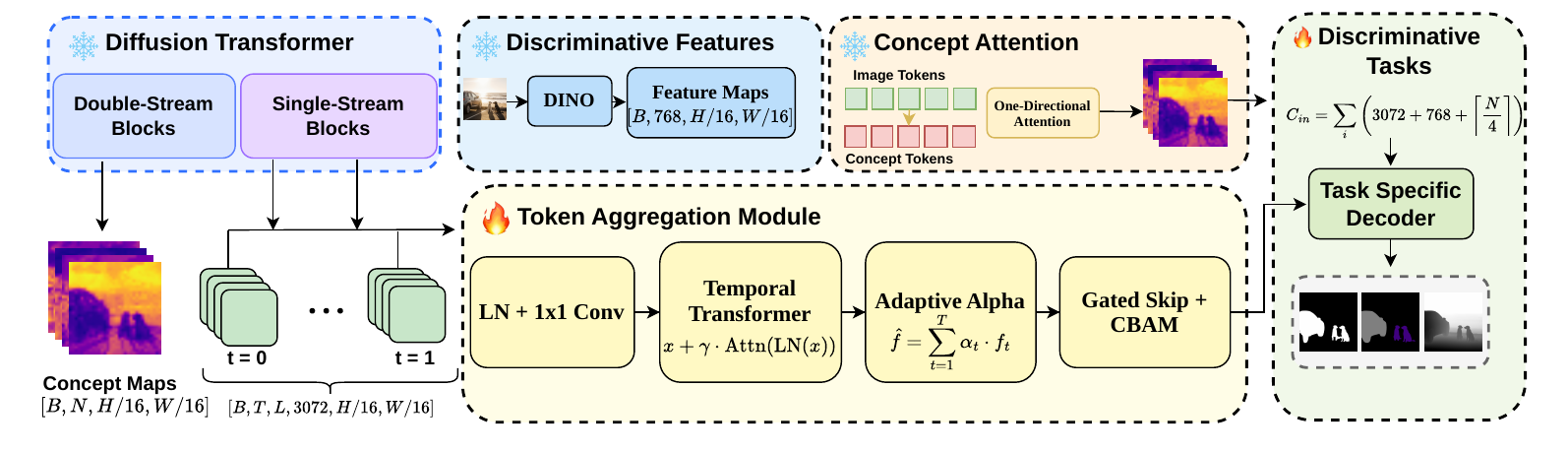}
  
  \caption{ \textbf{\method Pipeline:} Given a text prompt, the diffusion transformer generates an image through its dual-stream and single-stream blocks. Features are extracted, pruned, and combined with concept attention maps and discriminative features, then decoded to produce task-specific predictions for the generated image.}
  \label{fig:pipeline}
\end{figure*}

\paragraph{Multi-Model Diffusion.}

Image and video generators often require additional information about the generated content, such as segmentation or depth, for downstream applications. Recent work extends diffusion models to predict geometry alongside generation, though most approaches train separate decoder networks that operate independently of the generation process. For video, FantasyWorld~\cite{dai2025fantasyworld} trains a separate geometric branch with DPT ~\cite{ranftl2021vision} heads to decode depth, point maps, and camera poses alongside a frozen video foundation model, while JOG3R~\cite{huang2025jog3r} jointly trains a separate DUSt3R-style ~\cite{wang2024dust3r} geometry decoder with the video generator. WVD~\cite{wvd} trains the diffusion transformer itself to output geometry as separate output channels alongside RGB, though geometry is only available after the full generation process completes via the VAE decoder. Wonderland \cite{liang2025wonderland} and Lyra \cite{bahmani2025lyra} train 3DGS decoders from video latent, while FLAT \cite{kupyn2026flat} decodes triangle splats, turning video models into 3D generators. For images, DatasetDM~\cite{wu2023datasetdmsynthesizingdataperception} trains perception decoders on Stable Diffusion features to generate segmentation masks and depth maps, though the coarse spatial resolution of U-Net feature maps limits annotation quality. DiffuMask~\cite{wu2024diffumasksynthesizingimagespixellevel} is similarly constrained to single-concept segmentation by its reliance on cross-attention maps, while Dataset Diffusion~\cite{nguyen2023datasetdiffusion} addresses multi-class scenarios but faces the same U-Net feature limitations. GenPercept~\cite{xu2025what} and Diff-2-in-1~\cite{zheng2025diffin} integrate customized perception decoders with U-Net diffusion features for various dense prediction tasks. However, these U-Net-based approaches encode semantic information hierarchically through spatial downsampling, limiting the richness of extracted features. Our work demonstrates that transformer-based diffusion models distribute semantic information across layers and denoising timesteps without spatial downsampling. By recycling features computed during generation and aggregating them across multiple timesteps with learned spatial weighting, we extract substantially higher-quality annotations than U-Net-based methods using lightweight decoders on frozen transformer features.

\section{Method}
\label{sec:method}
Our goal is to extend a frozen diffusion transformer into a multi-modal generator by extracting and fusing features computed across the denoising process. Given a text prompt, we simultaneously generate an image and extract from the DiT internal representations any combination of dense perceptual modalities, such as semantic segmentation, depth, saliency, and more, using lightweight decoder heads. This transforms a single generative model into a pipeline that jointly generates images with any combination of semantic, geometric, and structural descriptions in a single generation pass.

Modern DiTs, such as FLUX~\cite{labs2023flux}, employ a dual-stream architecture comprising double-stream blocks followed by single-stream transformer blocks. These models maintain a flat architecture in which all tokens are processed at a uniform spatial resolution throughout the network. Unlike U-Nets, which create hierarchical representations through spatial downsampling \cite{zhao2023vpd}, this flat transformer architecture achieves semantic abstraction through attention mechanisms, thereby necessitating different feature extraction strategies. For example, diffusion transformer blocks produce features with very large activation magnitudes \cite{gan2025unleashing}, necessitating careful normalization when aggregating features across multiple layers and timesteps to ensure stable training and effective gradient flow.

Our framework serves two purposes: (1) generating synthetic datasets with automatically extracted annotations during image generation, and (2) extracting features from real images for validation and analysis. For real images, we apply the forward diffusion process~\cite{tang2023emergent} to add Gaussian noise:

\begin{equation}
x_t = \sqrt{\alpha_t}x_0 + \sqrt{1-\alpha_t}\epsilon , 
\end{equation}

where $x_0$ is the clean image, $\epsilon \sim \mathcal{N}(0, I)$ is random noise, and $t$ indexes the diffusion timestep. We then extract features from these noised images at multiple timesteps, enabling the same feature extraction mechanism to work for both synthetic generation and real image analysis.

\subsection{Multi-Timestep Feature Extraction}

We extract features from multiple layers $L_i \in \mathcal{L}$ within the DiT. We choose a variety of early and late layers to provide coverage across the transformer depth while maintaining computational efficiency. 

\textbf{Multi-Timestep Extraction.}
Semantic information in diffusion transformers is distributed across the denoising trajectory \cite{kim2025seg4diff}. Early timesteps 
capture coarse object layouts and scene-level semantics. Late timesteps 
preserve fine details and local structure but lose global semantic context \cite{yue2024exploring}. To access the full range of semantic knowledge, we extract features from multiple timesteps $t_j \in \mathcal{T}$ in the denoising schedule, where $t \in [0, 1]$ with $t=1$ representing pure noise and $t=0$ representing clean data. Combined with the layer selections, this yields $|\mathcal{L}||\mathcal{T}|$ feature maps total per input. We denote the extracted feature at layer $i$ and timestep $t$ as $F_t^{(i)} \in \mathbb{R}^{B \times N \times C}$, where $N$ is the number of spatial tokens and $C$ is the feature dimension.

\begin{figure}[t]
\centering

\begin{subfigure}[b]{\columnwidth}
\centering
\includegraphics[width=0.32\columnwidth]{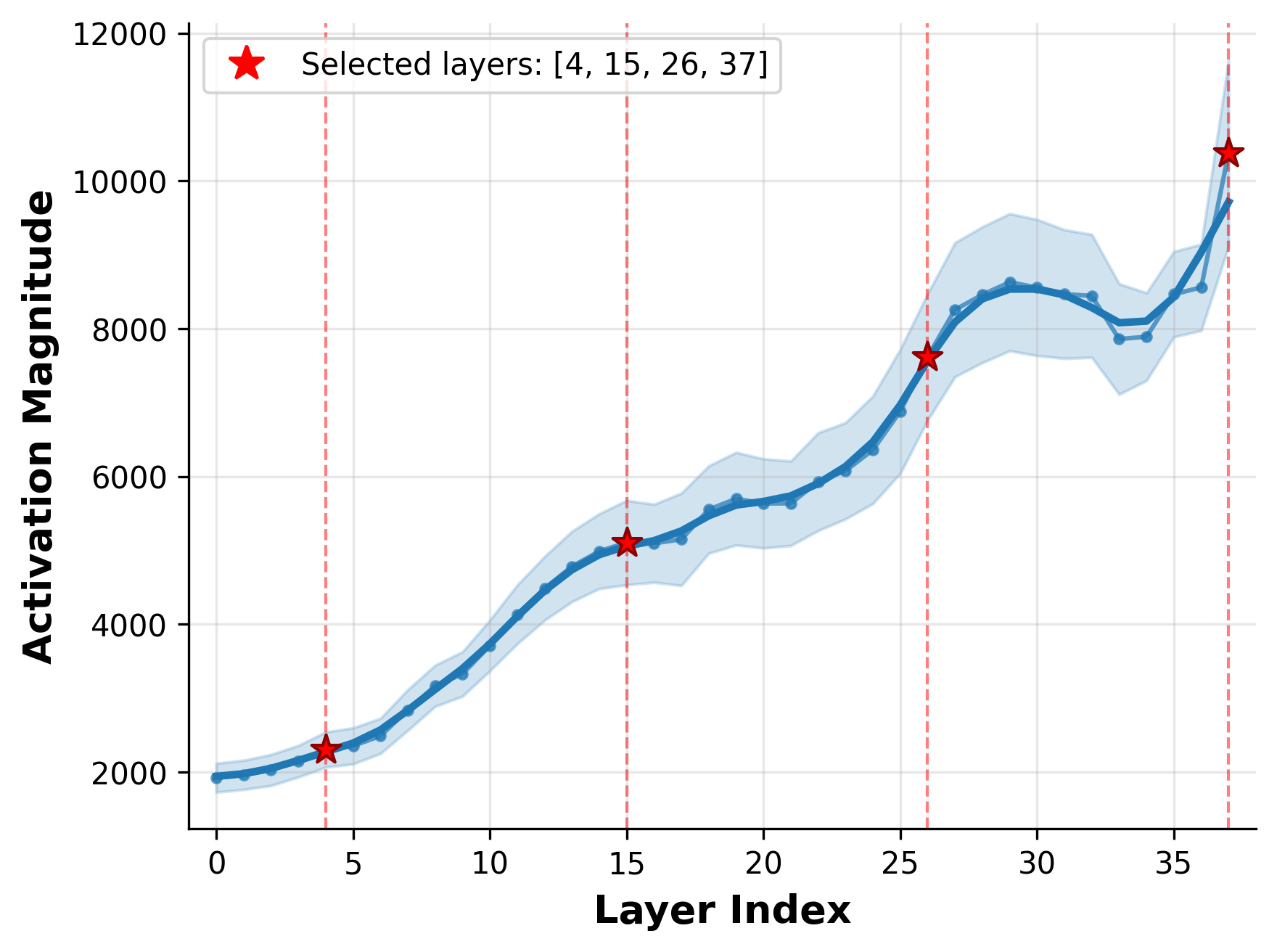}
\hfill
\includegraphics[width=0.32\columnwidth]{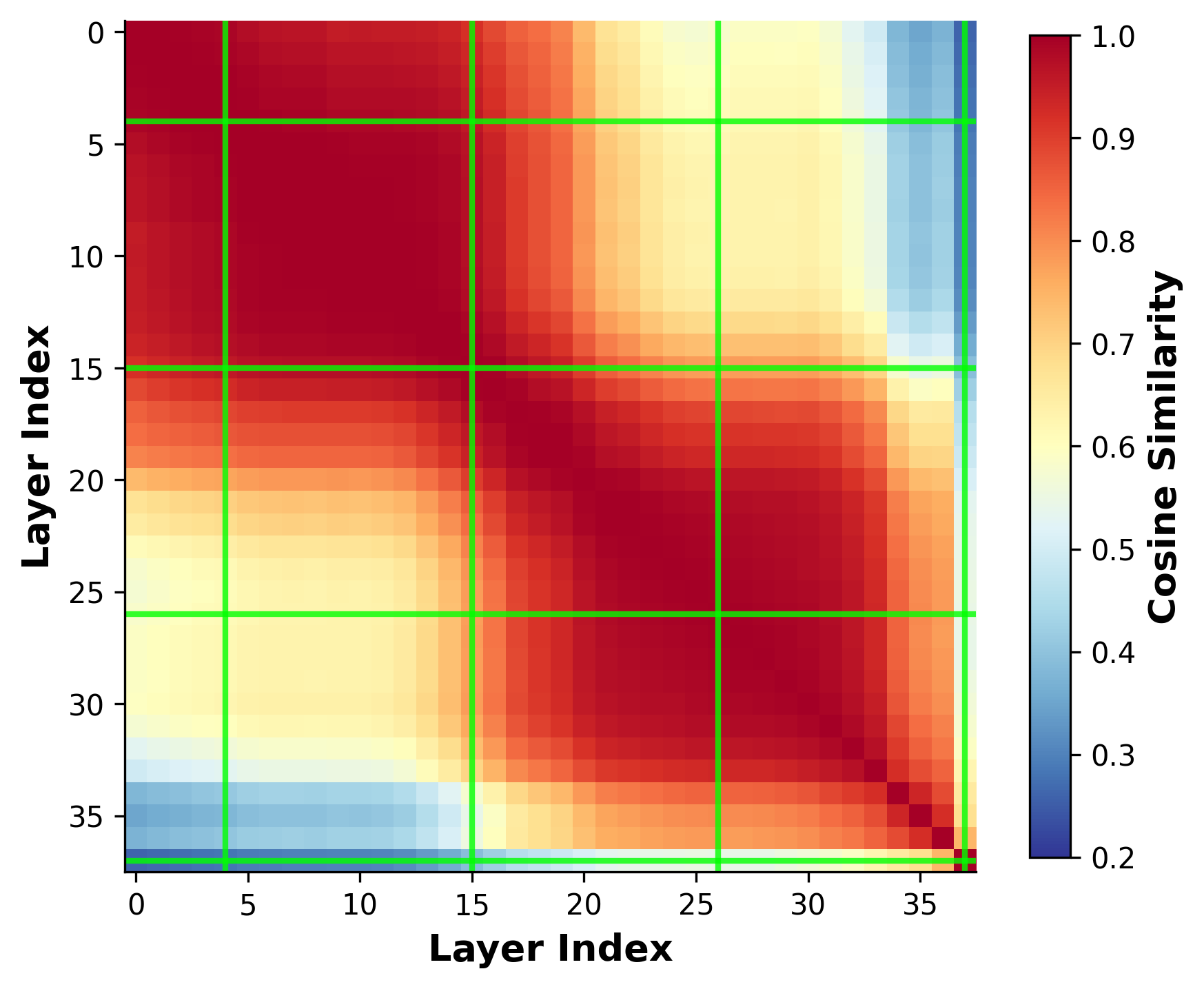}
\hfill
\includegraphics[width=0.32\columnwidth]{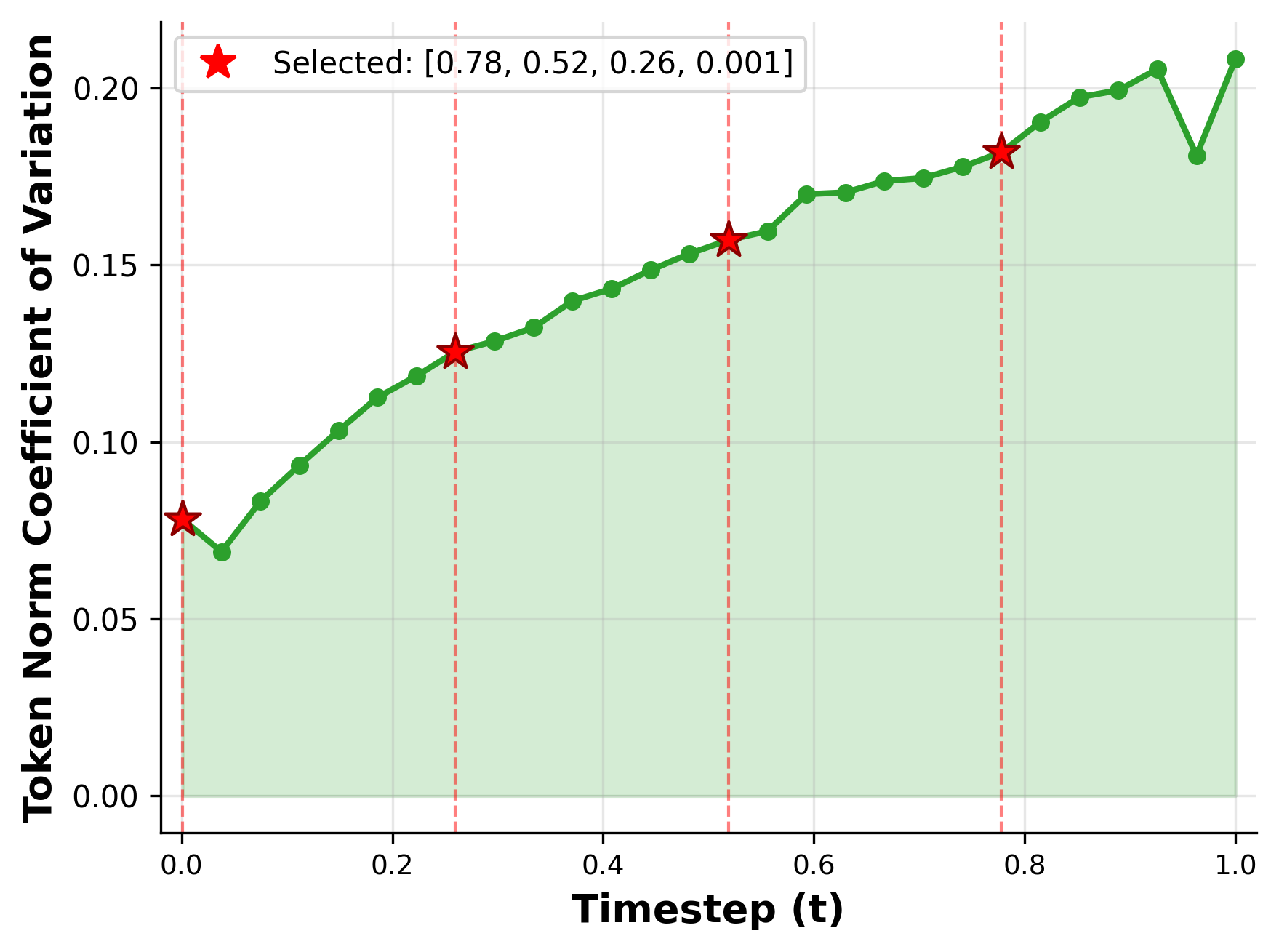}
\caption{Layer and timestep analysis: activation magnitudes, correlations, and token norm variation}
\end{subfigure}

\vspace{0.3em}

\begin{subfigure}[b]{\columnwidth}
\includegraphics[width=\textwidth]{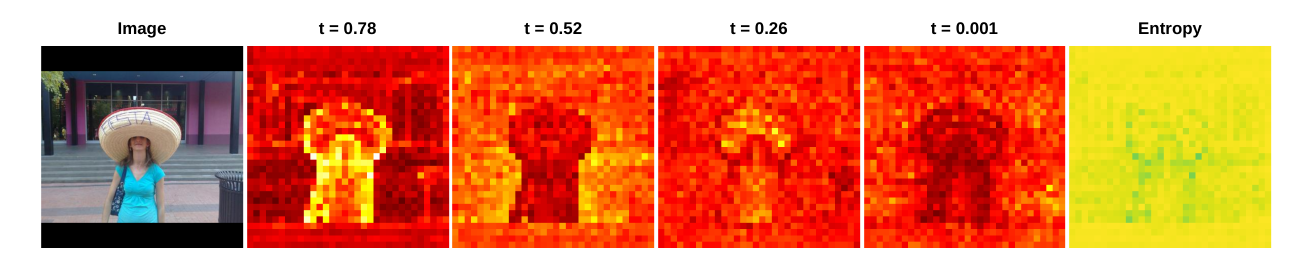}
\caption{Timestep entropy across the denoising trajectory}
\end{subfigure}

\caption{\textbf{Analysis of DiT features.} (a) Activation magnitudes grow monotonically across layers with high inter-layer correlation, motivating sparse layer sampling. (b) Feature entropy varies substantially across timesteps, capturing complementary coarse-to-fine information that justifies learned aggregation.}

\label{fig:rebuttal_analysis}
\end{figure}

\textbf{Token Aggregation.}
Since each feature map typically contains upwards of 1,000 tokens, each with several thousand channels, naive concatenation would yield prohibitively large representations, leading to memory overflow and computational infeasibility. Simple averaging, while memory-efficient, fails to leverage the complementary nature of features across timesteps and layers. Figure~\ref{fig:rebuttal_analysis} analyzes DiT feature characteristics to motivate our aggregation strategy. Figure~\ref{fig:rebuttal_analysis}a shows that activation magnitudes grow monotonically across FLUX's 38 single-stream layers with high correlation between adjacent layers, indicating gradual information propagation. This motivates our sparse layer sampling at $\mathcal{L} = \{4, 15, 26, 37\}$ to maximize coverage while avoiding redundancy. Figure~\ref{fig:rebuttal_analysis}b visualizes feature entropy across timesteps $\mathcal{T} = \{0.78, 0.52, 0.26, 0.001\}$, demonstrating distinct information content at each denoising stage: early timesteps capture coarse layouts with high entropy, middle timesteps refine object boundaries, and late timesteps preserve fine details. The complementary nature of these features necessitates learned spatially-varying aggregation rather than simple averaging.

The module first applies learned linear projections to reduce the feature dimension, producing $F_t^{\text{proj}} \in \mathbb{R}^{B \times N \times C'}$ where $C' < C$, followed by layer normalization. Since diffusion transformer blocks produce features with sometimes very large activation magnitudes, this normalization is critical for stable training. Additionally, different layers exhibit vastly different ranges, which this normalization mitigates.

The core aggregation mechanism is a 3-layer transformer with 8 attention heads and a hidden dimension of 768. Each layer includes pre-normalization, multi-head self-attention with temporal positional embeddings, and feedforward networks with GELU \cite{hendrycks2016gaussian} activation. We use LayerScale initialization~\cite{touvron2021going} ($10^{-4}$) for stability and dropout \cite{srivastava2014dropout} (0.1) for regularization. The transformer predicts per-pixel aggregation weights $\alpha \in \mathbb{R}^{|\mathcal{T}|}$ through a specialized output head, enabling each spatial location to adaptively weight the four timesteps based on local content.

The weighted combination is refined using Convolutional Block Attention (CBAM)~\cite{woo2018cbam}, which applies channel and spatial attention. A residual connection from the cleanest timestep ($t=0$) preserves fine details. Formally, for each spatial location:

\begin{equation}
    F_{\text{fused}} = \mathrm{CBAM}\!\left(\sum_{t \in \mathcal{T}} \alpha_t \cdot F_t^{\text{proj}}\right) + F_0^{\text{proj}} ,
\end{equation}

where the weights $\alpha_t$ sum to 1 at each location. This enables optimal temporal aggregation to vary spatially—smooth regions benefit from early-timestep consistency, whereas boundaries require late-timestep precision.

\subsection{Concept-Driven Attention Extraction}

Cross-attention maps between text and image tokens provide spatial guidance for dense prediction. However, averaging attention across all text tokens, including articles, prepositions, and punctuation, dilutes spatial precision~\cite{kukleva2025refam, erel2025attention}. Moreover, in multi-modal DiT architectures, text prompt embeddings are sequentially updated at each layer, making it difficult to extract attention maps for concepts not present in the original prompt without altering the generated image appearance.

We adopt concept-driven attention extraction~\cite{helbling2025conceptattention}, which addresses these limitations by creating a separate stream of concept embeddings that are not used for image generation. The method injects concept tokens (e.g., ``object'', ``background'') alongside text and image tokens, with initial embeddings produced by a text encoder. These concept embeddings are updated through a one-directional attention mechanism where concepts attend to both image patches and other concepts, but image tokens do not attend to concepts:

\begin{equation}
o_c = \text{softmax}(q_c k_{xc}^T) v_{xc}
\end{equation}

where $q_c$ are concept queries, and $k_{xc}, v_{xc}$ are concatenated image-concept keys and values. This one-directional design ensures that concepts gather information from the image without affecting generation.

Spatial maps are then extracted by computing dot-product similarities between concept output vectors $o_c$ and image patch output vectors $o_x$ in the attention output space:

\begin{equation}
\phi(o_x, o_c) = \text{softmax}(o_x o_c^T)
\end{equation}

This differs from traditional cross-attention maps, which operate in the query-key space. We aggregate information across multiple multi-modal layers by averaging: $\frac{1}{|\mathcal{L}|}\sum_{L=1}^{|\mathcal{L}|} \phi(o_x^L, o_c^L)$ where $|\mathcal{L}|$ denotes the number of multi-modal layers.

This approach provides flexibility to extract task-specific spatial guidance by defining relevant concepts at inference time without modifying the generation process. For semantic segmentation, we use concepts such as ``object'' and ``background'' in scene parsing. For depth estimation, ``near'' and ``far'' encode geometric relationships. For salient object detection, ``salient'' and ``contour'' provide targeted guidance. Importantly, this mechanism eliminates the need to fuse features from external text encoders into diffusion blocks, as done in VPD~\cite{zhao2023vpd}. This mechanism is unique to multi-modal DiT architectures and cannot be replicated by U-Net-based models.

\subsection{Discriminative Feature Complementarity}

While our framework relies solely on diffusion features, we find that discriminative representations can optionally complement them. Diffusion features encode semantic relationships from generative training, while discriminative features capture distinct visual characteristics from contrastive learning, and the two are complementary rather than redundant.
We optionally integrate DINO-v3 features~\cite{siméoni2025dinov3} with our diffusion-based representations. We extract features from DINO-v3-ViT-B/16 and concatenate them with our fused DiT features and concept maps. This creates a comprehensive representation where DiT provides semantic structure, concept maps offer targeted spatial guidance, and DINO-v3 contributes complementary discriminative information. As we demonstrate in our experiments, this integration consistently outperforms either feature type alone, confirming that generative and discriminative representations capture distinct visual information because of their fundamentally different origin.

\subsection{Task-Specific Decoders}

Our framework uses lightweight task-specific decoders that map extracted features to dense predictions. Decoders are trained on a small set of real annotations before being applied to decode any target modality during generation.

For semantic segmentation, we use the DeepLabV3+ decoder head~\cite{chen2018encoder} with ASPP \cite{chen2017deeplab} modules. Depth estimation employs a DPT-style decoder~\cite{ranftl2021vision} with hierarchical reassembly. Salient object detection uses a DPT decoder with a single-channel output. These decoders total approximately 36M trainable parameters, which is much less than a typical DiT (FLUX's 12B parameters remain frozen). We use frozen FLUX.1-dev~\cite{labs2023flux} as our diffusion backbone and extract features from single-stream blocks at layers $\mathcal{L} = \{4, 15, 26, 37\}$ and timesteps $\mathcal{T} = \{0.78, 0.52, 0.26, 0.001\}$. For decoder training, we use AdamW~\cite{loshchilov2017decoupled} with a learning rate of $10^{-4}$ and differential learning rates, where decoders use half the base rate. We apply Exponential Moving Average~\cite{polyak1992acceleration} with $\beta=0.999$ for stability.

\section{Experimental Evaluation}
\label{sec:experiments}

We extensively evaluate \method by training lightweight decoders for three dense prediction tasks: salient object detection, semantic segmentation, and depth estimation. 
U-Net-based approaches use different generators, meaning both task decoders and generated images differ across methods, making direct comparison of annotations against shared ground truth impossible. Since a single protocol cannot fairly evaluate all aspects simultaneously, we structure our evaluation around three complementary protocols, each addressing a distinct aspect of the pipeline: the consistency of generated image-annotation pairs, the effectiveness of the generated data for training discriminative models, and evaluation on real images against ground-truth labels.
Finally, we systematically ablate each component of our framework to quantify its individual contributions to overall performance.

\subsection{Experimental Setup}

We evaluate \method across three dense prediction tasks: semantic segmentation, salient object detection, and depth estimation. We report mean Intersection over Union (mIoU) for segmentation, S-measure ($S_m$)~\cite{smeasure}, max F-score ($F_m$)~\cite{f1max}; MAE~\cite{mae} for saliency; AbsRel, RMSE, and threshold accuracies $\delta_i$ for depth.

We compare against diffusion-based feature extractors (VPD~\cite{zhao2023vpd}, Diffusion Hyperfeatures~\cite{luo2024diffusionhyperfeaturessearchingtime}, DatasetDM~\cite{wu2023datasetdmsynthesizingdataperception}), discriminative encoders (DINO-v3~\cite{siméoni2025dinov3}), and synthetic data generation methods (DiffuMask~\cite{wu2024diffumasksynthesizingimagespixellevel}, Dataset Diffusion~\cite{nguyen2023datasetdiffusion}). To ensure fair comparison across all methods in Table~\ref{tab:feature_extraction_effectiveness}, we retrain all baseline methods under identical conditions. Each baseline uses the same training protocol (20 epochs, the same optimizer and learning rate), the same task-specific decoder architectures, and the same data augmentation strategy. This controlled setup isolates the impact of feature quality from differences in training procedures or decoder architectures.
The complete pipeline, including training decoders on real data, FLUX synthetic generation with label extraction, and downstream model training on synthetic data, requires approximately 24 hours per task on two 48GB GPUs.

\subsection{Multi-Modal Consistency Evaluation}

\begin{table}[t]
\centering
\caption{\textbf{Multi-Modal Consistency.} Each method generates images with their pipeline and extracts other modalities. We evaluate annotation accuracy against manually verified reference labels on each method's generated images.}
\label{tab:annotation_quality}

\resizebox{\columnwidth}{!}{%

\begin{minipage}{\columnwidth}

\scriptsize
\setlength{\tabcolsep}{0.5pt}
\renewcommand{\arraystretch}{1.05}

\begin{minipage}[t]{0.30\linewidth}
\centering
\begin{tabular}{@{}l c@{}}
\toprule
\multicolumn{2}{c}{\textbf{Semantic Segmentation}} \\
\midrule
Method & mIoU $\uparrow$ \\
\midrule
DatasetDM & 62.56 \\
DiffuMask & 57.35 \\
Dataset Diff. & 57.50 \\
\rowcolor{row}
\method (Ours) & \textbf{72.25} \\
\bottomrule
\end{tabular}
\end{minipage}
\hfill
\begin{minipage}[t]{0.68\linewidth}
\centering
\begin{tabular}{@{}l ccc cccc@{}}
\toprule
 & \multicolumn{3}{c}{\textbf{Saliency Detection}} & \multicolumn{4}{c}{\textbf{Depth Estimation}} \\
\cmidrule(lr){2-4} \cmidrule(lr){5-8}
Method
& $S_m \uparrow$ & $F_m \uparrow$ & MAE $\downarrow$
& AbsRel $\downarrow$ & RMSE $\downarrow$ & $\delta_1 \uparrow$ & $\delta_2 \uparrow$ \\
\midrule
DatasetDM
& 0.922 & 0.956 & 0.038
& 2.408 & 0.441 & 0.151 & 0.277 \\
\rowcolor{row}
\method (Ours)
& \textbf{0.944} & \textbf{0.966} & \textbf{0.019}
& \textbf{1.359} & \textbf{0.229} & \textbf{0.247} & \textbf{0.469} \\
\bottomrule
\end{tabular}
\end{minipage}

\end{minipage}
}
\end{table}

A key challenge in evaluating multi-modal diffusion models is that the generated images differ across methods, making a shared ground truth impossible. We instead measure consistency between each method's generated images and their predicted modalities, using SOTA models as pseudo-GT. Additionally, we manually verify the pseudo-GT annotations for correctness. For each task, we generate 200 images using the same text prompts across all methods. Each method generates images using its own generation pipeline and extracts annotations using its own approach. We then create reference annotations for each method's generated images: For salient object detection, we run S3OD~\cite{kupyn2025s3od} on each method's generated images and manually correct any errors. For depth, we use Depth Anything v2~\cite{yang2024depth}. For semantic segmentation, we run EoMT~\cite{kerssies2025eomt} and manually correct any errors in the resulting masks.

Table~\ref{tab:annotation_quality} reports multi-modal consistency against pseudo-GT labels comparing \method to Stable Diffusion-based multi-modal generation approaches \cite{nguyen2023datasetdiffusion, wu2023datasetdmsynthesizingdataperception, wu2024diffumasksynthesizingimagespixellevel}. Our method achieves substantially higher accuracy than baselines across all metrics. This demonstrates that our DiT-based features more accurately capture the visual content of generated images compared to U-Net-based extraction from Stable Diffusion. 

\subsection{Synthetic Data Generation}

We further validate \method by evaluating the pipeline's effectiveness as a synthetic data generator. Unlike the consistency evaluation, which evaluates MMDiff using diffusion features only, we use the full model, combining diffusion and DINO-v3 features, to achieve optimal performance. This also enables a direct comparison with DINO-v3 used as a standalone predictor, which serves as a natural baseline: one could simply generate images with FLUX and extract labels using a DINO-based model. We show that all prior methods fail to outperform this baseline, and the combination of discriminative and generative features we extract further improves performance.

\begin{table}[t]
\small
\centering
\setlength{\tabcolsep}{2.2pt}
\renewcommand{\arraystretch}{1.2}
\caption{\textbf{Semantic segmentation} on PASCAL VOC 2012. Our method outperforms DatasetDM \cite{wu2023datasetdmsynthesizingdataperception}, DiffuMask \cite{wu2024diffumasksynthesizingimagespixellevel}, and Dataset Diffusion \cite{nguyen2023datasetdiffusion} when trained purely on synthetic data, and achieves the best performance when combining synthetic pre-training with real fine-tuning.}
\label{tab:voc_table}
\begin{tabular}{lc|cccccccc|c}
\textbf{Syn.} & \textbf{Real} &
\faPlane & \faBicycle & \faDove & \faCar & \faDog & \faUser &
\faTrain & \faTv & \textbf{mIoU} \\
\toprule
\xmark & \cmark                 & 91.5 & 45.7 & 93.1 & 88.6 & 91.9 & 89.3 & 91.1 & 85.0 & 84.8 \\ \midrule
DatasetDM \cite{wu2023datasetdmsynthesizingdataperception} & \xmark         & 87.6 & 33.8 & 88.5 & 74.4 & \textbf{85.7} & 44.3 & 83.4 & 47.7 & 65.9 \\
DiffuMask \cite{wu2024diffumasksynthesizingimagespixellevel} & \xmark  & 84.7 & 35.8 & 70.3 & 52.5 & 53.1 & 52.8 & 75.8 & 42.9 & 54.7 \\
Dataset Diffusion \cite{nguyen2023datasetdiffusion} & \xmark  & 79.6 & 40.2 & 86.5 & 69.2 & 82.7 & 78.7 & 69.2 & 45.3 & 68.2 \\
DINO-v3 ~\cite{siméoni2025dinov3} & \xmark
& 88.9 & 41.0 & \textbf{90.4} & 78.5 & 82.6 & 86.7 & 84.3 & 63.2 & 75.2 \\
\rowcolor{row}
\method (Ours) & \xmark            & \textbf{93.6} & \textbf{45.1} & 88.8 & \textbf{85.4} & 83.9 & \textbf{92.3} & \textbf{89.8} & \textbf{69.1} & \textbf{78.9} \\
\midrule
DatasetDM \cite{wu2023datasetdmsynthesizingdataperception} & \cmark          & 95.7 & 47.9 & 96.0 & 92.5 & 95.2 & 92.9 & 92.1 & 84.2 & 85.2 \\
DiffuMask \cite{wu2024diffumasksynthesizingimagespixellevel} & \cmark          & 96.4 & 47.1 & 97.0 & \textbf{92.7} & 93.9 & 93.1 & 92.9 & 84.3 & 86.7 \\
Dataset Diffusion \cite{nguyen2023datasetdiffusion} & \cmark  & 95.6 & 47.5 & 95.7 & 91.4 & 94.1 & 92.6 & \textbf{95.8} & 84.0 & 85.7 \\
DINO-v3 ~\cite{siméoni2025dinov3} & \cmark
& 94.2 & 46.1 & 96.3 & 90.7 & 94.0 & 92.0 & 93.7 & 83.0 & 85.3 \\
\rowcolor{row}
\method (Ours) & \cmark          & \textbf{96.4} & \textbf{47.7} & \textbf{97.2} & 89.6 & \textbf{95.7} & \textbf{93.5} & 95.5 & \textbf{87.3} & \textbf{87.8} \\
\bottomrule
\end{tabular}
\end{table}

We use EoMT~\cite{kerssies2025eomt}, a state-of-the-art encoder-only mask transformer, as the downstream model. Pure synthetic training achieves 78.9\% mIoU, outperforming all prior synthetic generation methods: DatasetDM (65.9\%), 
DiffuMask (54.7\%) and Dataset Diffusion (68.2\%) all fall below the simple DINO-v3 standalone baseline (75.2\%). This confirms that simply running a discriminative model on generated images outperforms prior UNet-based methods. In contrast, \method shows that diffusion and discriminative features are complementary, outperforming all baselines and demonstrating the approach's effectiveness for synthetic data generation.

\begin{table*}[t]
\centering
\caption{\textbf{Synthetic data evaluation across tasks.}
We compare DatasetDM, DINO and \method under synthetic-only training, real-only training, and synthetic pre-training followed by real fine-tuning across salient object detection (DUTS) and depth estimation (NYU Depth v2).}
\label{tab:synthetic_eval_duts_nyu}

\resizebox{\textwidth}{!}{
\begin{tabular}{l c ccc ccccc}
\toprule
\textbf{Syn.} & \textbf{Real}
& $S_m \uparrow$ & $F_m \uparrow$ & MAE $\downarrow$
& AbsRel $\downarrow$ & RMSE $\downarrow$ & $\delta_1$ $\uparrow$ & $\delta_2$ $\uparrow$ & $\delta_3$ $\uparrow$ \\
\midrule
\xmark & \cmark
& 0.838 & 0.847 & 0.058
& 0.1291 & 0.416 & 0.868 & 0.983 & 0.996 \\

\midrule
DatasetDM~\cite{wu2023datasetdmsynthesizingdataperception} & \xmark
& 0.781 & 0.788 & 0.084
& 0.3918 & 1.044 & 0.620 & 0.879 & 0.957 \\
DINO-v3 ~\cite{siméoni2025dinov3} & \xmark
& 0.764 & 0.787 & 0.109
& 0.2389 & 0.626 &  0.647 & 0.887 & 0.972 \\
\rowcolor{row}
\method (Ours) & \xmark
& \textbf{0.784} & \textbf{0.798} & \textbf{0.070}
& \textbf{0.1880} & \textbf{0.521} & \textbf{0.714} & \textbf{0.914} & \textbf{0.982} \\

\midrule
DatasetDM~\cite{wu2023datasetdmsynthesizingdataperception} & \cmark
& 0.813 & 0.840 & 0.076
& 0.1648 & 0.490 & 0.804 & 0.982 & 0.996 \\
DINO-v3 ~\cite{siméoni2025dinov3} & \cmark
& 0.849 & 0.853 & 0.053
& 0.1415 & 0.452 & 0.850 & 0.987 & 0.997 \\
\rowcolor{row}
\method (Ours) & \cmark
& \best{0.863} & \best{0.876} & \best{0.049}
& \best{0.1185} & \best{0.392} & \best{0.885} & \best{0.989} & \best{0.998} \\

\bottomrule
\end{tabular}
}
\end{table*}

Table~\ref{tab:synthetic_eval_duts_nyu} reports results for salient object detection on DUTS  and depth estimation on NYU Depth v2. Consistent with segmentation, all prior methods fall below the DINO-v3 standalone baseline in the synthetic-only setting, while \method outperforms it across both tasks. Combining synthetic pretraining with fine-tuning on real data yields the best results across all metrics, confirming that labels extracted by our pipeline complement real data.

\subsection{Evaluation on Real Images}

\begin{table*}[t]
\centering
\caption{\textbf{Feature Quality Validation:} \method combines features from multiple sources, outperforming both existing diffusion-based extractors and discriminative encoders such as DINO-v3 \cite{siméoni2025dinov3} across all tasks. All methods use frozen encoders except where noted.}
\label{tab:feature_extraction_effectiveness}
\resizebox{\textwidth}{!}{
\begin{tabular}{l c c ccc ccccc}
\toprule
 & \multicolumn{1}{c}{\textbf{Trainable}}
 & \multicolumn{1}{c}{\textbf{PASCAL VOC 2012}}
 & \multicolumn{3}{c}{\textbf{DUTS}}
 & \multicolumn{5}{c}{\textbf{NYU Depth v2}} \\
\cmidrule(lr){2-2} \cmidrule(lr){3-3} \cmidrule(lr){4-6} \cmidrule(lr){7-11}
\textbf{Method}
& \textbf{Encoder}
& mIoU $\uparrow$ 
& $S_m \uparrow$ & $F_m \uparrow$ & MAE $\downarrow$
& AbsRel $\downarrow$ & RMSE $\downarrow$ & $\delta_1$ $\uparrow$ & $\delta_2$ $\uparrow$ & $\delta_3$ $\uparrow$ \\
\midrule
DatasetDM \cite{wu2023datasetdmsynthesizingdataperception}
& \xmark
& 41.19 
& 0.845 & 0.909 & 0.077 
& 0.1536 & 0.612 & 0.772 & 0.971 & \textbf{0.983} \\
Diffusion Hyperfeatures \cite{luo2024diffusionhyperfeaturessearchingtime}
& \xmark
& 67.57
& 0.892 & 0.910 & \underline{0.021} 
& 0.1348 & 0.503 & 0.844 & 0.971 & 0.978 \\
VPD \cite{zhao2023vpd}
& \cmark
& \textbf{82.36}
& \underline{0.912} & \underline{0.916} & 0.024 
& \underline{0.1244} & \underline{0.402} & \underline{0.867} & \underline{0.973} & \underline{0.981}  \\
\rowcolor{row}
\method (Ours)
& \xmark
& \underline{78.90}
& \textbf{0.918} & \textbf{0.929} & \textbf{0.020}
& \textbf{0.1175} & \textbf{0.370} & \textbf{0.951} & \textbf{0.976} & \textbf{0.983} \\
\midrule
\midrule
DINO-v3 \cite{siméoni2025dinov3}
& \cmark
& 83.09
& \underline{0.920} & \underline{0.922} & 0.021 
& 0.1288 & 0.389 & 0.945 & \underline{0.975} & \textbf{0.983} \\

VPD + DINO-v3 \cite{zhao2023vpd,siméoni2025dinov3}
& \cmark
& \underline{83.77}
& 0.918 & 0.914 & \underline{0.020}
& \underline{0.1223} & \underline{0.387} & \underline{0.946} & \underline{0.975} & \textbf{0.983} \\

\rowcolor{row}
\method + DINO-v3
& \xmark
& \textbf{84.95} 
& \textbf{0.934} & \textbf{0.947} & \textbf{0.018}
& \textbf{0.1164} & \textbf{0.365} & \textbf{0.952} & \textbf{0.976} & \textbf{0.983} \\
\bottomrule
\end{tabular}
}
\end{table*}

Finally, we fix the image and evaluate the predicted modalities against the ground-truth labels, isolating prediction quality from generation. For real images, we follow the noising procedure described in Section~\ref{sec:method} to extract features from the frozen backbone. We generate conditioning prompts by captioning each image with Gemma-3-4B~\cite{gemma_2025}. This evaluation addresses two questions: how do frozen DiT features compare with discriminative encoders and methods that fine-tune diffusion backbones for discriminative tasks, and what is the impact of multi-timestep aggregation versus optional DINO integration?
Table~\ref{tab:feature_extraction_effectiveness} evaluates the quality of extracted features by comparing performance across three dense prediction tasks. This controlled evaluation demonstrates that our multi-timestep aggregation, concept-driven attention, and discriminative feature integration produce strong representations for dense prediction.

\setlength{\cellw}{0.132\textwidth}
\setlength{\rowh}{2.0cm} 

\begin{figure*}[t]
  \centering
  \setlength{\tabcolsep}{1pt}
  \renewcommand{\arraystretch}{0}
  \begin{tabular}{@{}*{7}{m{\cellw}}@{}}

    \img{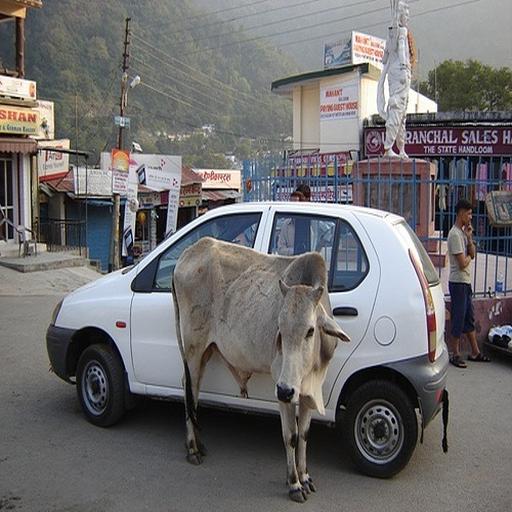} &
    \img{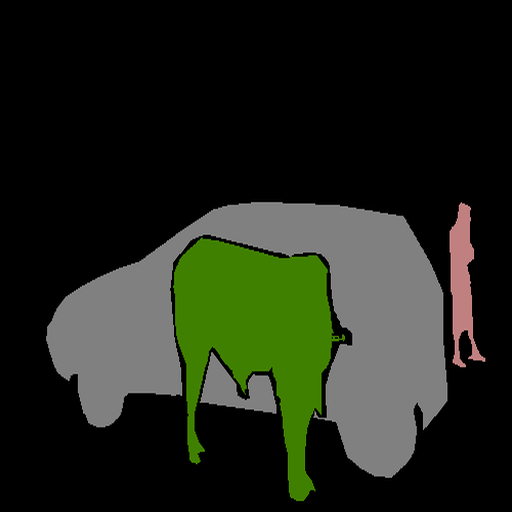} &
    \img{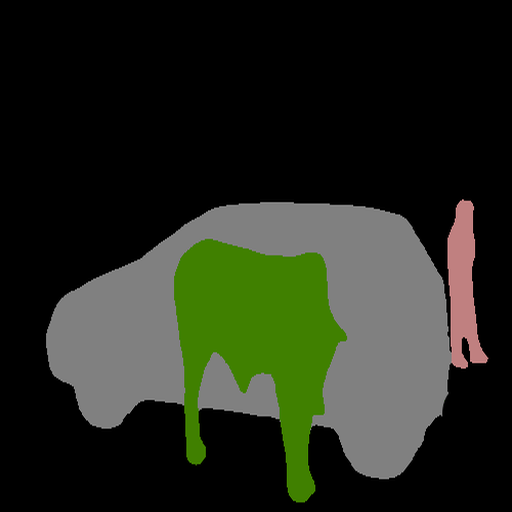} &
    \img{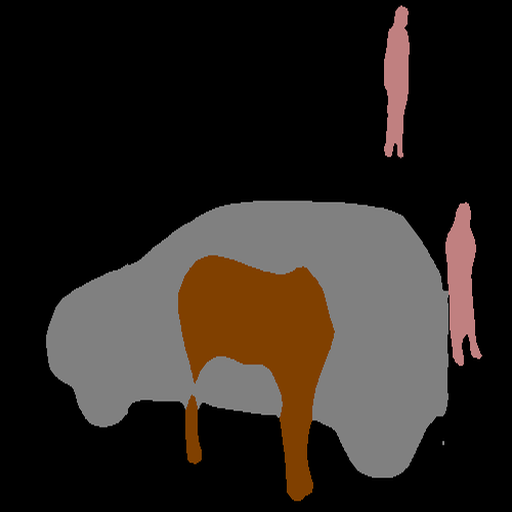} &
    \img{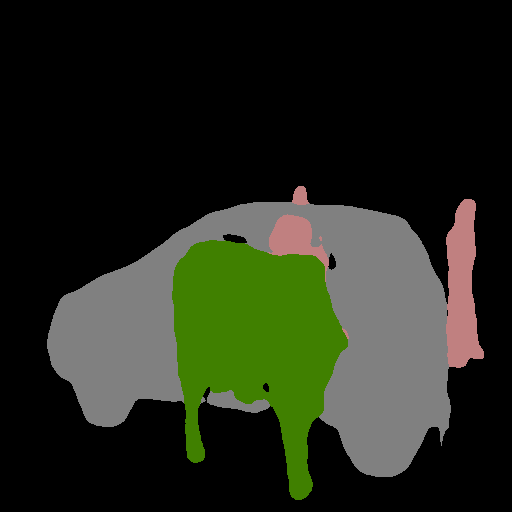} &
    \img{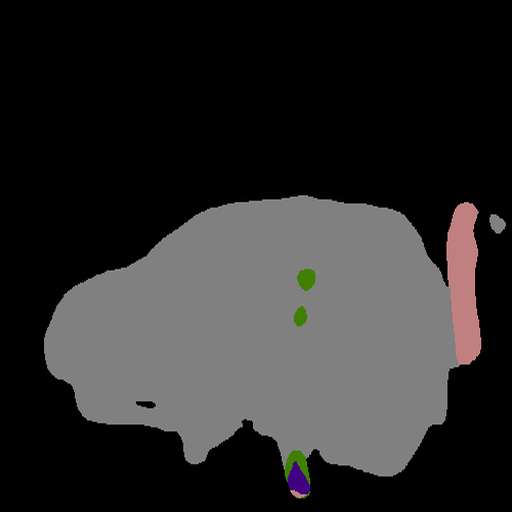} &
    \img{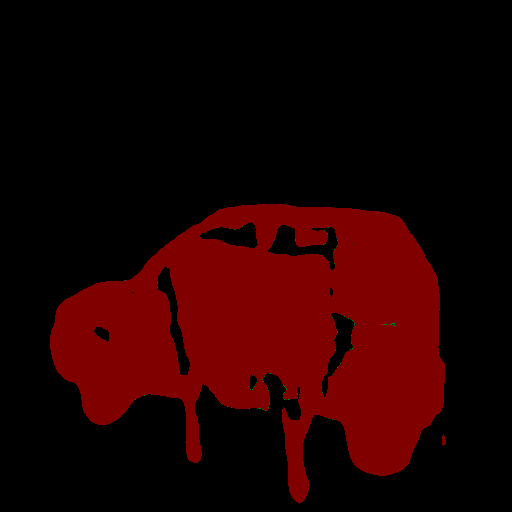} \\[-2pt]

    \img{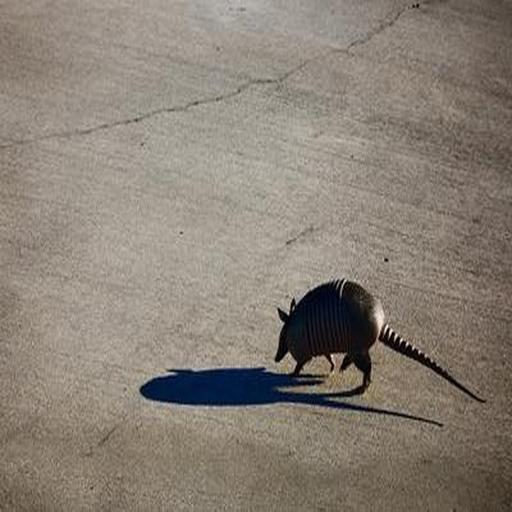} &
    \img{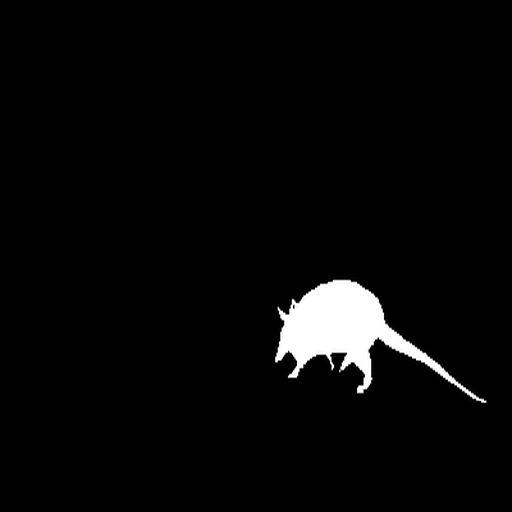} &
    \img{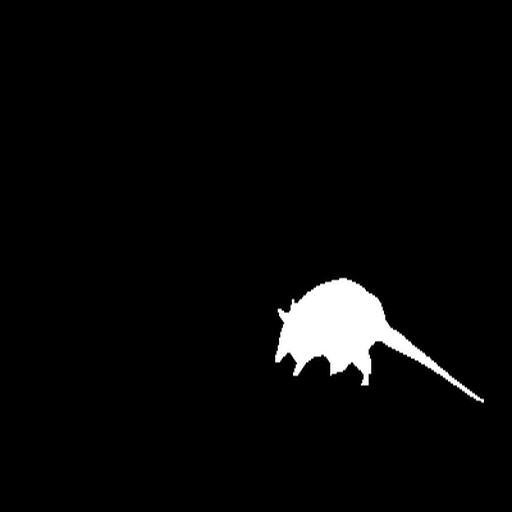} &
    \img{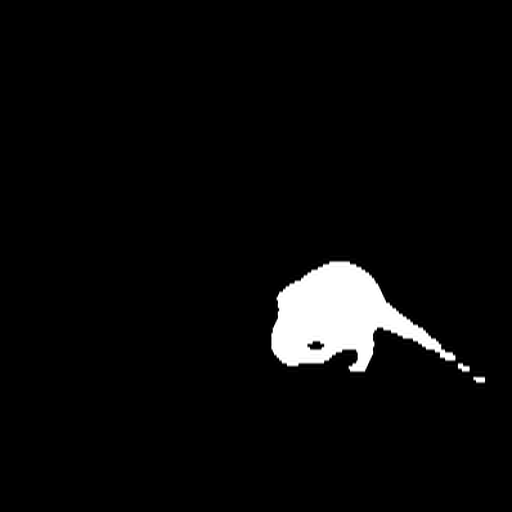} &
    \img{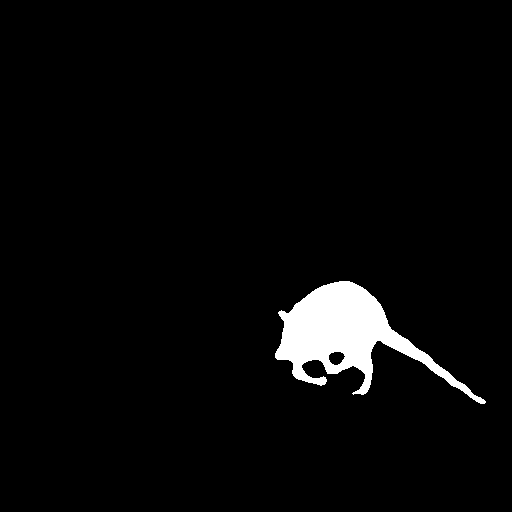} &
    \img{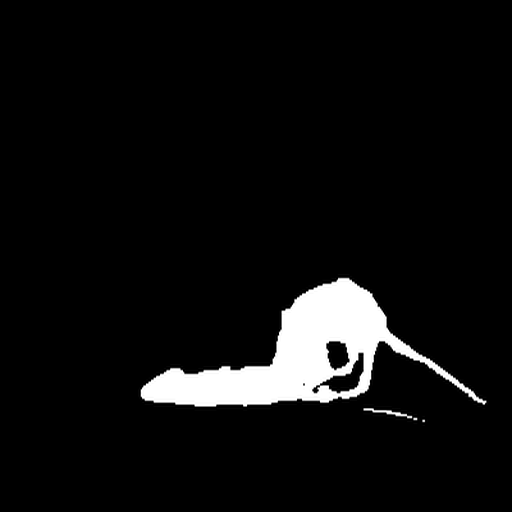} &
    \img{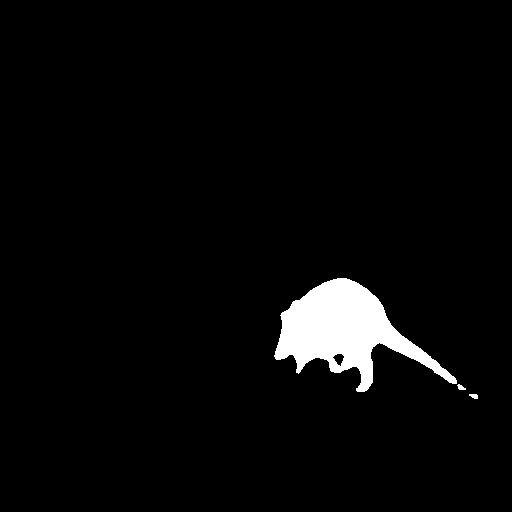} \\[-2pt]

    \img{} &
    \img{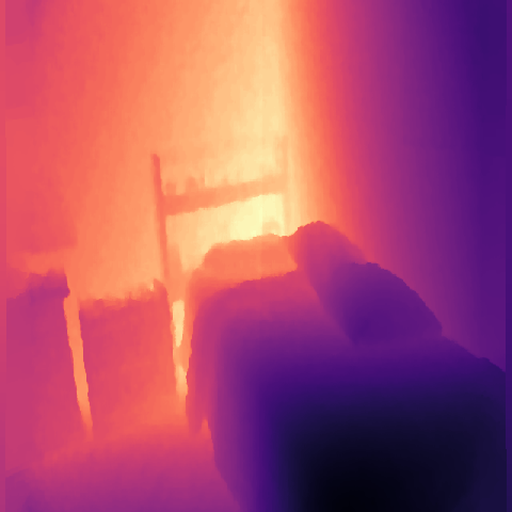} &
    \img{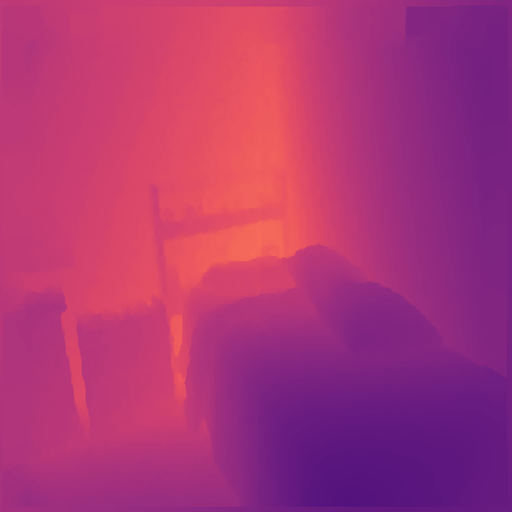} &
    \img{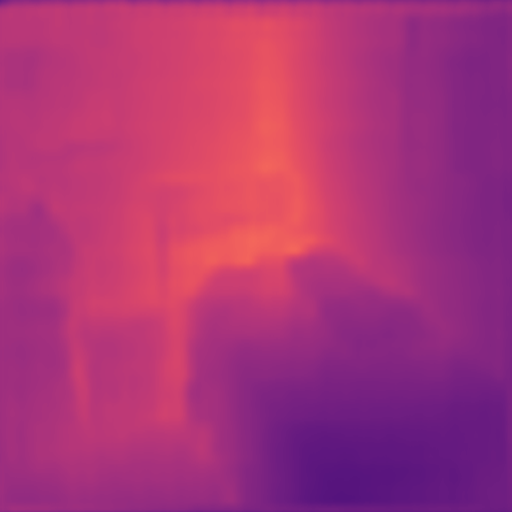} &
    \img{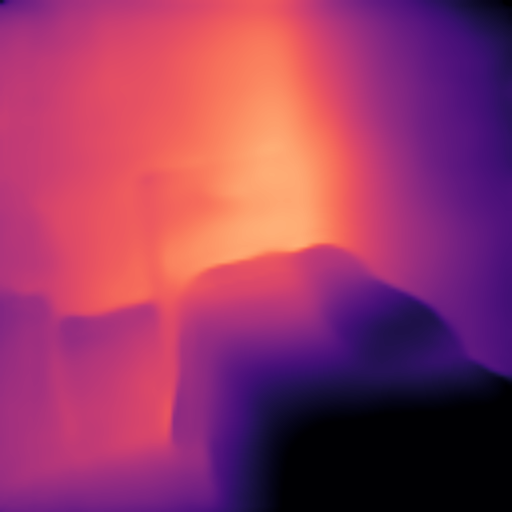} &
    \img{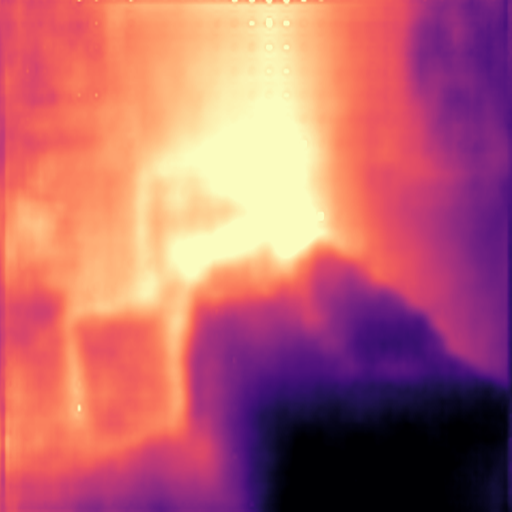} &
    \img{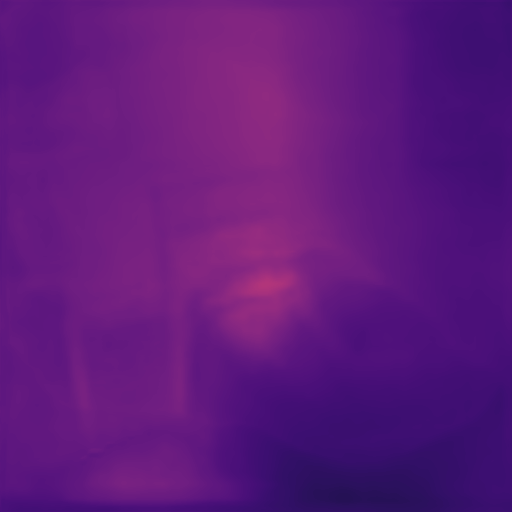} \\[-1pt]

    \parbox[c]{\cellw}{\centering\tiny\textbf{\mbox{Image}}} &
    \parbox[c]{\cellw}{\centering\tiny\textbf{\mbox{GT}}} &
    \parbox[c]{\cellw}{\centering\tiny\textbf{\mbox{\method\ (ours)}}} &
    \parbox[c]{\cellw}{\centering\tiny\textbf{\mbox{VPD~\cite{zhao2023vpd}}}} &
    \parbox[c]{\cellw}{\centering\tiny\textbf{\mbox{DINO-v3~\cite{siméoni2025dinov3}}}} &
    \parbox[c]{\cellw}{\centering\tiny\textbf{\mbox{DHF~\cite{luo2024diffusionhyperfeaturessearchingtime}}}} &
    \parbox[c]{\cellw}{\centering\tiny\textbf{\mbox{DatasetDM~\cite{wu2023datasetdmsynthesizingdataperception}}}} \\
  \end{tabular}

  \caption{
    \textbf{Qualitative comparisons on PASCAL (top), DUTS (middle), and NYU (bottom).}
    Our method achieves sharper boundaries and more consistent depth ordering than prior works.
  }
  \label{fig:qualitative_grid}
\end{figure*}

\noindent\textbf{Comparison with U-Net-based Diffusion Methods.}
\method substantially outperforms all U-Net-based diffusion methods while keeping the encoder fully frozen. Compared with VPD~\cite{zhao2023vpd}, which requires fine-tuning the denoising UNet, we achieve a +2.08\% mIoU on segmentation and a +0.011 S-measure on saliency with a frozen encoder. Diffusion Hyperfeatures~\cite{luo2024diffusionhyperfeaturessearchingtime}, another U-Net extractor, achieves only 67.57\% mIoU compared to our 84.44\%. The comparison with Diffusion Hyperfeatures is particularly informative, as both methods extract multi-timestep features, yet we achieve a 16.87\% higher mIoU. This validates that our adaptive aggregation mechanism, which utilizes spatially varying weights, is crucial for effectively combining temporal information from transformer architectures.
Notably, our FLUX-only features (78.90\% mIoU) outperform Diffusion Hyperfeatures (67.57\%) by 11.33\%, demonstrating the superiority of transformer-based diffusion architectures over U-Nets for feature extraction.

\noindent\textbf{Frozen vs. Trainable Encoders.}
An important distinction is that our method achieves 84.44\% mIoU with a completely frozen 12B-parameter FLUX encoder, whereas VPD (82.36\%) and DINO-v3 (83.09\%) require encoder fine-tuning. This demonstrates that our features are sufficient to support dense prediction without requiring expensive encoder adaptation. For synthetic data generation, this frozen approach is essential: we can generate annotations using the same frozen features across all tasks without task-specific fine-tuning.

\noindent\textbf{Feature Complementarity.}
While our frozen diffusion features alone are competitive with finetuned discriminative encoders, integrating DINO-v3 further improves performance from 78.90\% to 84.44\% mIoU. Crucially, this combination outperforms both FLUX-only (78.90\%) and DINO-only (83.09\%) features, as well as VPD+DINO-v3 (83.77\%), thereby confirming that our multi-timestep diffusion features capture complementary information to that of discriminative encoders. Diffusion features encode rich semantic and compositional knowledge from their generative training objective, while DINO captures distinct visual characteristics from contrastive learning.
Together, these results establish three findings: for the first time, we show that frozen DiT features are competitive with fine-tuned discriminative encoders; multi-timestep aggregation is the primary driver of this performance; and combining generative and discriminative features yields the best results because of their fundamentally different training objectives.

\noindent\textbf{Qualitative Results.}
Figure~\ref{fig:qualitative_grid} demonstrates the superior quality of our extracted features across semantic segmentation, salient object detection, and depth estimation. Our method produces coherent multi-object segmentations, clean saliency maps with precise boundaries, and smooth depth predictions with preserved structural detail. In contrast, U-Net-based methods exhibit semantic confusion, fragmented predictions, and noisy outputs. 

\subsection{Ablation Study}

\begin{table*}[t]
\centering
\caption{\textbf{Ablation study.} Single-stream features outperform double-stream features even at single timestep. Multi-timestep aggregation provides the largest improvement. Concept-driven attention and DINO integration provide complementary benefits, with the full model achieving best performance across all tasks.}
\label{tab:ablation}

\footnotesize
\setlength{\tabcolsep}{3pt}
\renewcommand{\arraystretch}{1.05}

\begin{adjustbox}{max width=\textwidth}
\begin{tabular}{l c ccc ccccc}
\toprule
 & \multicolumn{1}{c}{\textbf{VOC'12}}
 & \multicolumn{3}{c}{\textbf{DUTS}}
 & \multicolumn{5}{c}{\textbf{NYU Depth v2}} \\
\cmidrule(lr){2-2} \cmidrule(lr){3-5} \cmidrule(lr){6-10}
\textbf{Component}
& mIoU $\uparrow$
& $S_m \uparrow$ & $F_m \uparrow$ & MAE $\downarrow$
& AbsRel $\downarrow$ & RMSE $\downarrow$ & $\delta_1$ $\uparrow$ & $\delta_2$ $\uparrow$ & $\delta_3$ $\uparrow$ \\
\midrule
double-stream block (t=0) & 40.52 & 0.884 & 0.916 & 0.033 & 0.622 & 1.213 & 0.740 & 0.930 & 0.975 \\
single-stream block (t=0) & 50.20 & 0.890 & 0.914 & 0.032 & 0.478 & 0.738 & 0.780 & 0.945 & 0.978 \\
\quad + Cross-attention maps & 58.31 & 0.869 & 0.900 & 0.043 & 0.455 & 0.723 & 0.800 & 0.953 & 0.980 \\
\quad + Concept attention & 63.77 & 0.903 & 0.921 & 0.024 & 0.431 & 0.700 & 0.820 & 0.960 & 0.982 \\
Multi-timestep + Concept attention & 78.90 & 0.918 & 0.929 & 0.020 & 0.118 & 0.370 & 0.951 & 0.976 & 0.983 \\
Full model (w/o aggregation) & 78.21 & 0.918 & 0.927 & 0.017 & 0.130 & 0.571 & 0.814 & 0.972 & 0.979 \\
\midrule
\textbf{Full model (4T + Concepts + DINO)} & \textbf{84.95} & \textbf{0.934} & \textbf{0.947} & \textbf{0.018} & \textbf{0.116} & \textbf{0.365} & \textbf{0.952} & \textbf{0.976} & \textbf{0.983} \\
\bottomrule
\end{tabular}
\end{adjustbox}
\end{table*}

Table~\ref{tab:ablation} evaluates the contribution of each component in our framework. We systematically ablate components to quantify their impact on the pipeline.

\noindent\textbf{Multi-Timestep Token Aggregation:}
Aggregating features over four timesteps yields the largest improvement, increasing mIoU from 50.20\% to 78.90\% (+28.7\%). This substantial gain validates our core hypothesis: semantic information in diffusion transformers is temporally distributed across the denoising trajectory. Early timesteps capture global structure and scene layouts, while later timesteps refine local details. Single-timestep extraction at $t=0$ captures only the final refined features, missing the coarse semantic information present at earlier steps.

\noindent\textbf{Concept-Driven Attention:}
Concept attention provides substantial improvements with single-timestep features (+13.57\% mIoU), demonstrating its value for spatial guidance. While temporal aggregation captures much of the semantic structure across the denoising trajectory, concept maps contribute task-specific spatial cues that enhance prediction quality. Concept attention also improves upon averaged cross-attention maps, achieving 63.77\% mIoU compared to 58.31\% with single-timestep features. This demonstrates that querying specific semantic concepts produces sharper, more task-relevant spatial guidance than averaging across all text tokens.

\noindent\textbf{Discriminative Feature Integration:}
Adding DINO-v3 features to our multi-timestep FLUX features improves the mIoU from 78.90\% to 84.44\% (+5.54\%). This demonstrates that combining features from generative and discriminative pre-training objectives yields complementary information, with the final model outperforming both FLUX-only (78.90\%) and DINO-only (83.09\%) models.

These ablations explain why our synthetic data outperforms prior methods. Multi-timestep aggregation captures the full semantic trajectory, producing more consistent annotations across diverse scenes. Concept-driven attention provides task-specific spatial guidance, yielding sharper boundaries in the extracted masks. Together, these components enable high-quality label extraction, making our synthetic data practical for training discriminative models.

\section{Extension to Video Diffusion Models}
\label{sec:video}
To test whether feature aggregation transfers beyond image-space dense prediction, we apply \method to a video-to-3D setting. We adopt the latent-to-3DGS~\cite{kerbl2023gaussian} pipeline of Lyra~\cite{bahmani2025lyra} and Wonderland~\cite{liang2025wonderland}, which reconstructs 3D Gaussians from a denoised video latent of world-consistent video model ~\cite{wvd, kupyn2025epipolar, du2026videogpa, an2026vggrpo}, and replace that input with our multi-timestep aggregated features. We train the Lyra decoder on 1,000 RealEstate10K~\cite{zhou2018stereo} videos and evaluate on 200 held-out scenes.

\begin{table}[h]
\centering
\small
\setlength{\tabcolsep}{8pt}
\renewcommand{\arraystretch}{1.15}
\caption{\textbf{Video $\rightarrow$ 3DGS transfer.} Replacing the denoised latent
with aggregated \method features yields consistent gains in a low-data setting.}
\label{tab:video_transfer}
\begin{tabular}{l|ccc}
\toprule
\textbf{Input} & \textbf{PSNR $\uparrow$} & \textbf{SSIM $\uparrow$} & \textbf{LPIPS $\downarrow$} \\
\midrule
Denoised Latent Only & 21.37 & 0.705 & 0.258 \\
\rowcolor{row}
\method Aggregated Features & \textbf{21.61} & \textbf{0.747} & \textbf{0.241} \\
\bottomrule
\end{tabular}
\end{table}

\begin{figure}[h]
\centering
\includegraphics[width=\linewidth]{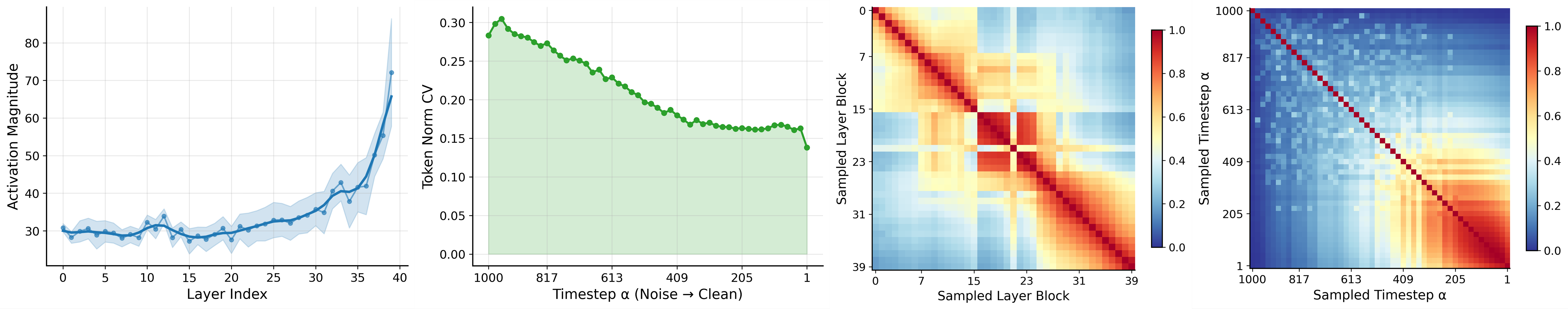}
\caption{\textbf{Wan-2.1 layer/timestep analysis.} Feature behavior across video
DiT layers and timesteps mirrors the image-model trends.}
\label{fig:wan_analysis}
\end{figure}

Aggregated features improve over the denoised latent across all three metrics ($+0.24$ PSNR, $+0.042$ SSIM, $-0.017$ LPIPS), indicating that the benefit of multi-timestep aggregation is not specific to image-space dense prediction but extends to video and 3D. Fig.~\ref{fig:wan_analysis} shows a layer/timestep analysis on Wan-2.1~\cite{wan2025}, which follows the same trends observed for image DiTs: early timesteps and shallow layers carry fine detail, later timesteps and deeper layers carry semantic structure.

\section{Conclusion}
\label{sec:conclusion}

We presented \method, a framework that transforms a frozen diffusion transformer into a multi-modal generative system that extracts dense perceptual annotations during image generation. By recycling the rich features computed across the denoising trajectory rather than discarding them after rendering, our approach addresses a key inefficiency in current generative pipelines: the need to re-extract semantic, geometric, and structural information from generated pixels using separate task-specific models.
Our work demonstrates that transformer-based diffusion models encode substantially richer perceptual representations than U-Net architectures. This architectural advantage stems from the temporal distribution of semantic information across the denoising trajectory, which we capture through adaptive multi-timestep aggregation with spatially varying weights. Combined with concept-driven attention extraction for targeted spatial guidance and optional discriminative feature integration, this approach yields high-quality annotations that substantially outperform prior U-Net-based methods on synthetic data generation, direct annotation quality, and real-image prediction tasks.
By enabling multi-modal annotation extraction from a single frozen generative model, \method provides a practical framework for cost-effective, scalable synthetic dataset creation. This opens new possibilities for leveraging generative models in vision tasks where manual annotation is expensive or prohibitive, and establishes that diffusion transformers can serve as effective feature extractors beyond their generative capabilities.

\section*{Acknowledgments}
This research was funded by the European Union (ERC, Volute, 101222037). O.K. is supported by a Google unrestricted gift. Views and opinions expressed are however those of the author(s) only and do not necessarily reflect those of the European Union or the European Research Council. Neither the European Union nor the granting authority can be held responsible for them.

\bibliographystyle{splncs04}
\bibliography{main}

\clearpage
\appendix
\maketitlesupplementary

\section{Learning from Limited Data}

\method is particularly valuable when labeled data is limited. By combining complementary features from multiple timesteps, concept-driven attention, and discriminative representations, our framework extracts diverse information from each sample, reducing overfitting and providing a richer training signal than any single feature source alone. This diversity is especially helpful in low-data regimes, where a narrow representation otherwise overfits quickly. Tables~\ref{tab:voc_limited_supp}, \ref{tab:duts_limited_supp}, and \ref{tab:nyu_limited_supp} evaluate whether generating synthetic samples with \method improves performance in limited data scenarios. We investigate three training scenarios: (1) baseline models trained only on limited real data, (2) models trained purely on synthetic samples, and (3) models pre-trained on synthetic samples then fine-tuned on limited real data.

\subsection{Semantic Segmentation}

\begin{table}[h]
\small
\centering
\setlength{\tabcolsep}{2.2pt}
\renewcommand{\arraystretch}{1.2}
\caption{\textbf{Semantic segmentation with limited supervision} on PASCAL VOC 2012. Synthetic data generated from limited real supervision (1k, 5k, 10.5k) provides substantial improvements, with the best performance achieved by combining synthetic pre-training and real fine-tuning.}
\label{tab:voc_limited_supp}
\begin{tabular}{lc|cccccccc|c}
\textbf{Syn.} & \textbf{Real} &
\faPlane & \faBicycle & \faDove & \faCar & \faDog & \faUser &
\faTrain & \faTv & \textbf{mIoU} \\
\toprule
\multicolumn{11}{l}{\textit{Decoder trained with 1k real images}} \\
\xmark & \cmark & 94.3 & 45.9 & 94.7 & 87.1 & 58.4 & \textbf{93.8} & 90.8 & 81.8 & 79.3 \\
\method & \xmark & 91.7 & \textbf{46.1} & 92.4 & 83.4 & 71.8 & 90.3 & 89.4 & 70.9 & 72.7 \\
\rowcolor{row}
\method & \cmark & \textbf{95.2} & 45.9 & \textbf{97.0} & \textbf{90.9} & \textbf{92.1} & 92.8 & \textbf{95.6} & \textbf{85.5} & \textbf{85.2} \\
\midrule
\multicolumn{11}{l}{\textit{Decoder trained with 5k real images}} \\
\xmark & \cmark & 92.6 & 45.8 & 91.6 & 90.0 & 91.2 & 89.3 & 88.8 & 82.9 & 84.5 \\
\method & \xmark & 94.8 & 47.7 & 87.2 & 88.9 & 78.6 & 91.4 & 86.2 & 68.0 & 77.4 \\
\rowcolor{row}
\method & \cmark & \textbf{96.0} & \textbf{50.0} & \textbf{96.7} & \textbf{93.8} & \textbf{95.3} & \textbf{93.0} & \textbf{93.2} & \textbf{86.3} & \textbf{87.7} \\
\midrule
\multicolumn{11}{l}{\textit{Decoder trained with 10.5k real images}} \\
\xmark & \cmark & 91.5 & 45.7 & 93.1 & 88.6 & 91.9 & 89.3 & 91.1 & 85.0 & 84.8 \\
\method & \xmark & 93.6 & 45.1 & 88.8 & 85.4 & 83.9 & 92.3 & 89.8 & 69.1 & 78.9 \\
\rowcolor{row}
\method & \cmark & \textbf{96.4} & \textbf{47.7} & \textbf{97.2} & \textbf{89.6} & \textbf{95.7} & \textbf{93.5} & 95.5 & \textbf{87.3} & \textbf{87.8} \\
\bottomrule
\end{tabular}
\end{table}

We use EoMT~\cite{kerssies2025eomt}, a state-of-the-art encoder-only mask transformer, as the downstream model. With only 1k real samples, training purely on 10k synthetic samples achieves 72.7\% mIoU, while training on the 1k real data yields 79.3\% mIoU. This shows our extracted labels are accurate enough to supervise training, though limited real data still provides an advantage. Combining synthetic pre-training with real fine-tuning yields the best results at 85.2\% mIoU, a +5.9\% improvement over real-only training.

With 5k real samples, the pattern holds: synthetic-only achieves 77.4\% mIoU, real-only 84.5\%, and combined 87.7\% mIoU. The consistent gains from combining synthetic pre-training with real fine-tuning across all data regimes demonstrate that synthetic data provides diverse coverage of the visual domain, while real data captures distribution-specific nuances. Notably, the improvement from synthetic pre-training remains substantial even with full data (+3.0\% mIoU), confirming the value of our approach beyond low-data scenarios.

\subsection{Salient Object Detection}

\begin{table}[h]
\centering
\caption{\textbf{Salient object detection with limited supervision} on DUTS. Training on synthetic images achieves strong performance even with limited real supervision (1k, 5k, 10.5k), with the best results from combining synthetic pre-training and real fine-tuning.}
\label{tab:duts_limited_supp}
\begin{tabular}{lc|ccc}
\textbf{Syn.} & \textbf{Real} & \textbf{$S_m \uparrow$} & \textbf{$F_m \uparrow$} & \textbf{MAE $\downarrow$} \\
\midrule
\multicolumn{5}{l}{\textit{Decoder trained with 1k real images}} \\[2pt]
\xmark & \cmark & 0.777 & 0.741 & 0.083 \\
\method & \xmark & 0.799 & 0.827 & 0.075 \\
\rowcolor{row}
\method & \cmark & \textbf{0.840} & \textbf{0.838} & \textbf{0.057} \\
\midrule
\multicolumn{5}{l}{\textit{Decoder trained with 5k real images}} \\[2pt]
\xmark & \cmark & 0.826 & 0.819 & 0.062 \\
\method & \xmark & 0.782 & 0.803 & 0.073 \\
\rowcolor{row}
\method & \cmark & \textbf{0.854} & \textbf{0.863} & \textbf{0.054} \\
\midrule
\multicolumn{5}{l}{\textit{Decoder trained with 10.5k real images}} \\[2pt]
\xmark & \cmark & 0.838 & 0.847 & 0.058 \\
\method & \xmark & 0.784 & 0.798 & 0.070 \\
\rowcolor{row}
\method & \cmark & \textbf{0.863} & \textbf{0.876} & \textbf{0.049} \\
\bottomrule
\end{tabular}
\end{table}

Employing IS-Net~\cite{qin2022dis}, training on synthetic images generated from only 1k real samples achieves 0.799 S-measure, approaching the performance of 10× more real data (0.838 S-measure with 10.5k real samples). The combination of synthetic pre-training with 1k real fine-tuning yields 0.840 S-measure, outperforming the real-only baseline. 

With 5k real samples, the combined approach achieves 0.854 S-measure and 0.054 MAE, demonstrating consistent benefits. With the full 10.5k training set, synthetic-only achieves 0.784 S-measure, real-only 0.838 S-measure, and combined 0.863 S-measure with 0.049 MAE. The consistent improvements from synthetic pre-training across all data regimes confirm that synthetic data provides complementary coverage that enhances performance beyond low-data scenarios.

\subsection{Depth Estimation}

\begin{table}[h]
\centering
\caption{\textbf{Depth estimation with limited supervision} on NYU Depth v2. Synthetic data generated from limited real samples (2.4k, 12k, 24k) substantially improves performance when combined with real fine-tuning.}
\label{tab:nyu_limited_supp}
\small
\setlength{\tabcolsep}{4pt}
\renewcommand{\arraystretch}{1.05}
\begin{tabular}{lc|ccccc}
\textbf{Syn.} & \textbf{Real} & \textbf{AbsRel $\downarrow$} & \textbf{RMSE $\downarrow$} &
\textbf{$\delta_1$ $\uparrow$} & \textbf{$\delta_2$ $\uparrow$} & \textbf{$\delta_3$ $\uparrow$} \\
\midrule
\multicolumn{7}{l}{\textit{Decoder trained with 2.4k real images}} \\
\xmark & \cmark  & 0.1386 & 0.447 & 0.808 & 0.915 & 0.927 \\
\method & \xmark & 0.1938 & 0.537 & 0.693 & 0.887 & 0.953 \\
\rowcolor{row}
\method & \cmark & \textbf{0.1219} & \textbf{0.403} & \textbf{0.860} & \textbf{0.961} & \textbf{0.970} \\
\midrule
\multicolumn{7}{l}{\textit{Decoder trained with 12k real images}} \\
\xmark & \cmark  & 0.1303 & 0.420 & 0.860 & 0.974 & 0.987 \\
\method & \xmark & 0.1900 & 0.527 & 0.706 & 0.904 & 0.971 \\
\rowcolor{row}
\method & \cmark & \textbf{0.1192} & \textbf{0.394} & \textbf{0.880} & \textbf{0.983} & \textbf{0.992} \\
\midrule
\multicolumn{7}{l}{\textit{Decoder trained with 24k real images}} \\
\xmark & \cmark  & 0.1291 & 0.416 & 0.868 & 0.983 & 0.996 \\
\method & \xmark & 0.1880 & 0.521 & 0.714 & 0.914 & 0.982 \\
\rowcolor{row}
\method & \cmark & \textbf{0.1185} & \textbf{0.392} & \textbf{0.885} & \textbf{0.989} & \textbf{0.998} \\
\bottomrule
\end{tabular}
\end{table}

Experiments with DepthFormer~\cite{li2023depthformer} demonstrate the effectiveness of synthetic data across different data regimes. With 2.4k real samples used to train the feature extractor, real-only training achieves 0.1386 AbsRel and 0.808 $\delta_1$, while synthetic-only training yields 0.1938 AbsRel and 0.693 $\delta_1$—showing that real data still provides better direct supervision than synthetic data alone at this scale. However, combining synthetic pre-training with real fine-tuning achieves the best performance at 0.1219 AbsRel and 0.860 $\delta_1$, a substantial improvement over real-only training.

With 12k real samples, the pattern holds: real-only achieves 0.1303 AbsRel, synthetic-only 0.1900 AbsRel, while the combined approach yields the best results at 0.1192 AbsRel and 0.880 $\delta_1$. With the full 24k training set, real-only achieves 0.1291 AbsRel, synthetic-only 0.1880 AbsRel, and combined 0.1185 AbsRel with 0.885 $\delta_1$. The consistent gains from combining synthetic pre-training with real fine-tuning across all data regimes demonstrate that synthetic data provides complementary signal that enhances generalization even when substantial real data is available. Notably, the improvement remains meaningful even with full data (0.1185 vs 0.1291 AbsRel, an 8.2\% reduction in relative error). 

\section{Unified Multi-Task Decoder}

The experiments above train a separate decoder per task. Here we test whether a single decoder can serve multiple tasks from the same frozen \method features. We train unified decoders on VOC+DUTS, VOC+NYU, and all three tasks jointly, with per-task loss weights set from the relative loss magnitudes early in training.

\begin{table}[h]
\centering
\caption{\textbf{Single- vs.\ multi-task decoders.} Each unified decoder is
evaluated only on its trained tasks (``--'' = not trained in that config).
Semantic tasks transfer positively; depth degrades under joint training with
semantic tasks.}
\label{tab:multitask}
\small
\resizebox{\textwidth}{!}{
\begin{tabular}{l c ccc ccccc}
\toprule
 & \multicolumn{1}{c}{\textbf{VOC}}
 & \multicolumn{3}{c}{\textbf{DUTS}}
 & \multicolumn{5}{c}{\textbf{NYU Depth v2}} \\
\cmidrule(lr){2-2} \cmidrule(lr){3-5} \cmidrule(lr){6-10}
\textbf{Decoder}
& mIoU $\uparrow$
& $S_m \uparrow$ & $F_m \uparrow$ & MAE $\downarrow$
& AbsRel $\downarrow$ & RMSE $\downarrow$ & $\delta_1 \uparrow$ & $\delta_2 \uparrow$ & $\delta_3 \uparrow$ \\
\midrule
Single-task
& 84.95
& \textbf{0.934} & \textbf{0.947} & \textbf{0.018}
& \textbf{0.116} & \textbf{0.365} & \textbf{0.952} & \textbf{0.976} & \textbf{0.983} \\
VOC + DUTS
& 84.92
& 0.925 & 0.946 & 0.022
& -- & -- & -- & -- & -- \\
VOC + NYU
& \textbf{85.42}
& -- & -- & --
& 0.190 & 0.505 & 0.861 & 0.950 & 0.978 \\
VOC + DUTS + NYU
& 84.96
& 0.929 & 0.947 & 0.020
& 0.220 & 0.579 & 0.736 & 0.920 & 0.960 \\
\bottomrule
\end{tabular}
}
\end{table}

A single decoder on VOC+DUTS matches the per-task baselines (VOC $84.92$ vs.\ $84.95$; DUTS $F_m$ $0.946$ vs.\ $0.947$), and VOC improves under joint training with NYU ($85.42$) and all three tasks ($84.96$), indicating positive transfer between semantic tasks. Depth, however, degrades sharply when combined with semantic supervision (NYU AbsRel $0.22$ for all-three vs.\ $0.116$ single-task), reflecting a geometry/semantics gap rather than a limitation of the features. \method thus supports both single- and multi-task decoders, and the per-task choice is governed by this trade-off.

\section{Feature and Concept Visualizations}

\begin{figure}[h!]
  \centering
  \includegraphics[width=\linewidth]{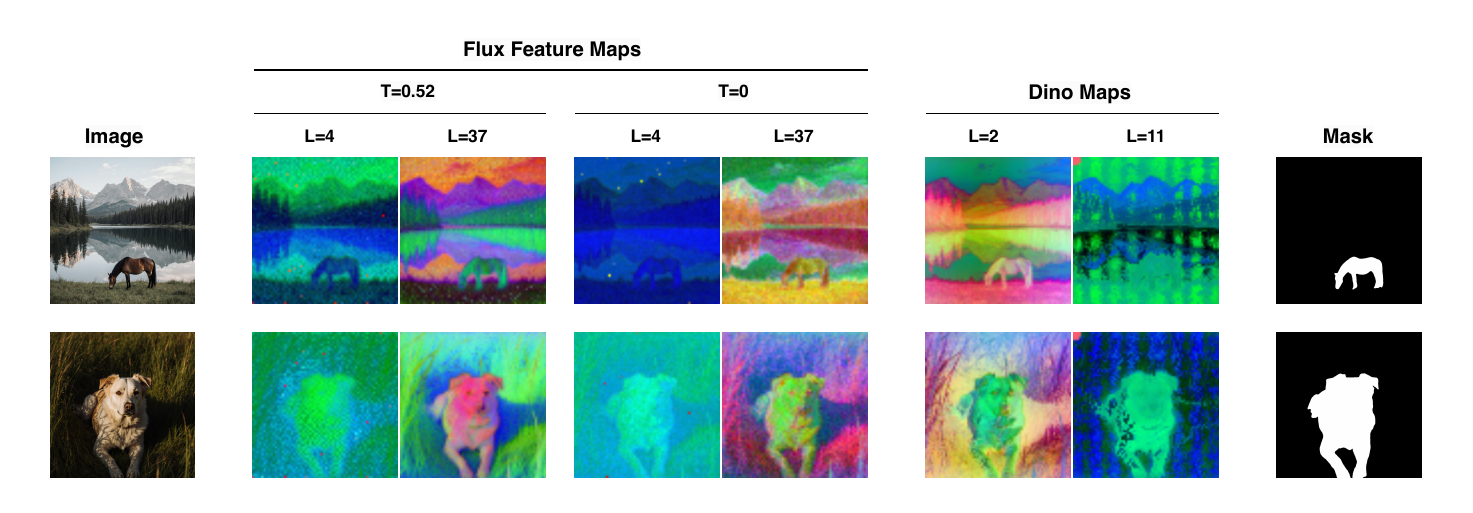}
  
  \caption{\textbf{Multi-modal feature extraction for salient object detection.} FLUX features extracted at multiple timesteps ($t=0.52$, $t=0$) and layers (L=4, L=37), combined with Dino-v3 features (L=2, L=11), enable accurate segmentation on synthetic images. Different timesteps and layers capture complementary information: early timesteps and shallow layers preserve fine-grained details, while late timesteps and deep layers encode semantic object-level features. The diversity across temporal and spatial dimensions provides rich multi-scale representations for precise segmentation.}
  \label{fig:duts_feature}
\end{figure}

\begin{figure}[h!]
  \centering
  \includegraphics[width=\linewidth]{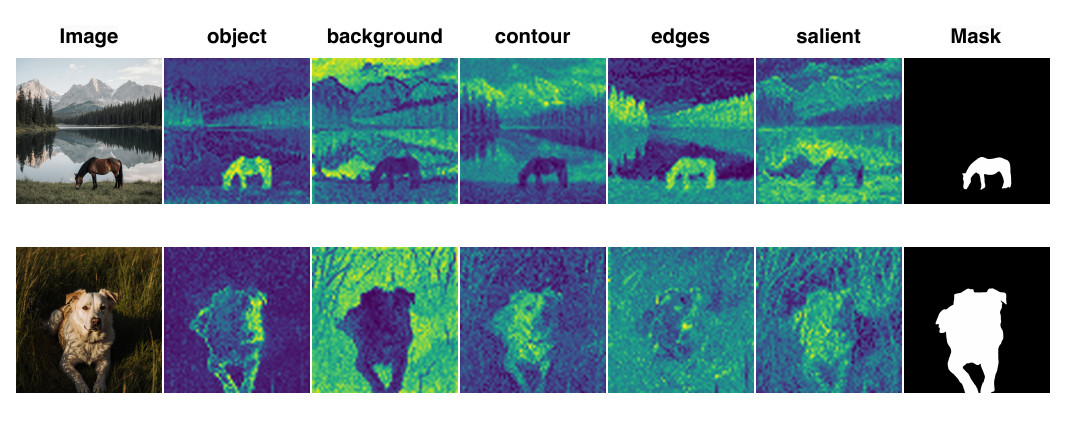}
  
  \caption{\textbf{Concept attention visualization.} Visualization of learned concept-specific attention maps for salient object detection. Each concept attends to distinct semantic aspects: \textit{object} (salient regions), \textit{background} (surrounding context), \textit{contour} (object boundaries), \textit{edges} (fine-grained edge information), and \textit{salient} (integrated saliency map). The clear separation between concepts demonstrates that our attention mechanism learns interpretable, complementary representations}
  \label{fig:duts_concept}
\end{figure}

To understand how \method leverages multi-modal representations, we visualize both the extracted features and learned concept attention maps in Fig.~\ref{fig:duts_feature} and Fig.~\ref{fig:duts_concept}.

\textbf{Multi-timestep feature extraction.} Fig.~\ref{fig:duts_feature} demonstrates the complementary information captured across different timesteps and layers of FLUX. Features from early layers (L=4) capture fine-grained textural details, while deeper layers (L=37) encode more semantic, object-level information with stronger color contrast corresponding to salient regions. Notably, the FLUX features at different timesteps and layers exhibit distinct activation patterns where some emphasize object boundaries, others highlight uniform regions within objects. This diversity across the temporal and spatial dimensions of the transformer provides rich, multi-scale representations. When combined with Dino-v3 features, which offer strong semantic discrimination and object localization capabilities learned through self-supervised learning, the multi-modal feature set enables precise segmentation.

\textbf{Concept attention patterns.} Fig.~\ref{fig:duts_concept} visualizes the learned concept-specific attention maps that guide the segmentation decoder. These visualizations reveal how our concept attention mechanism decomposes the segmentation task into interpretable sub-tasks, with each concept map learning to attend to specific visual properties. The clear separation between concepts (e.g., sharp object-background distinction, precise contour localization) demonstrates that the concept attention mechanism successfully learns meaningful decompositions rather than redundant representations. This interpretable, concept-driven approach contributes to \method's robust performance on diverse salient object detection scenarios.

\section{Effect of Concept Sets}

\begin{table}[t]
\centering
\footnotesize
\setlength{\tabcolsep}{4pt}
\caption{Pascal VOC performance of single-stream ($t=0$) models with different concept sets.}
\label{tab:voc_concepts}

\begin{tabular}{c p{7.0cm}}
\toprule
\textbf{mIoU $\uparrow$} & \textbf{Concept Set} \\
\midrule
50.2 & -- \\

62.1 & living, vehicle, furniture, object, background \\

63.5 & object, background, living, vehicle, furniture, detail, edges \\

63.6 & object, background, detail, edges, airplane, bicycle, bird, boat, bottle, bus, car, cat, chair, cow, table, dog, horse, motorbike, person, plant, sheep, sofa, train, television \\

63.3 & living, vehicle, furniture, object, background, detail, edges, airplane, bicycle, bird, boat, bottle, bus, car, cat, chair, cow, table, dog, horse, motorbike, person, plant, sheep, sofa, train, television \\

\textbf{63.8} & object, background, detail, edges \\

\bottomrule
\end{tabular}

\end{table}

\begin{table}[t]
\centering
\footnotesize
\setlength{\tabcolsep}{4pt}
\caption{DUTS performance of single-stream ($t=0$) models with different concept sets.}
\label{tab:duts_single_concepts}

\begin{tabular}{c c c p{4.4cm}}
\toprule
\textbf{MAE $\downarrow$} & \textbf{F$_m$ $\uparrow$} & \textbf{S$_m$ $\uparrow$} & \textbf{Concept Set} \\
\midrule

0.032 & 0.914 & 0.890 & -- \\

\textbf{0.025} & 0.915 & 0.899 & object, foreground, background, scene \\

0.026 & 0.918 & 0.894 & object, background, detail, edges \\

\textbf{0.025} & \textbf{0.921} & \textbf{0.903} & background, object, edges, salient, contour \\

\bottomrule
\end{tabular}

\end{table}

\begin{table}[t]
\centering
\footnotesize
\setlength{\tabcolsep}{4pt}
\caption{NYU Depth V2 performance of single-stream ($t=0$) models with different concept sets.}
\label{tab:nyu_concepts}

\begin{tabular}{c c c c c p{3.0cm}}
\toprule

\textbf{AbsRel $\downarrow$} &
\textbf{RMSE $\downarrow$} &
\textbf{$\delta_1 \uparrow$} &
\textbf{$\delta_2 \uparrow$} &
\textbf{$\delta_3 \uparrow$} &
\textbf{Concept Set} \\

\midrule

0.478 & 0.738 & 0.780 & 0.945 & 0.988 & -- \\

0.462 & 0.785 & 0.792 & 0.945 & 0.987 &
depth, surface, object, background \\

0.478 & 0.810 & 0.780 & 0.940 & 0.985 &
wall, floor, furniture, object, depth \\

0.470 & 0.795 & 0.788 & 0.943 & 0.986 &
near, far, depth, distance, surface \\

\textbf{0.431} & \textbf{0.700} & \textbf{0.820} & \textbf{0.960} & \textbf{0.992} &
depth, surface, object, near, far \\

\bottomrule
\end{tabular}

\end{table}

To determine the optimal spatial guidance for each task, we evaluated various concept sets using the single-stream (latest timestep) model. We compare these against a baseline model that utilizes extracted FLUX features without concept attention guidance. For computational efficiency, NYU Depth experiments were conducted on a subset of 1,159 training images and 290 test images, while DUTS and Pascal VOC experiments utilize the full datasets.

\textbf{Impact of Concept Guidance.}
Incorporating concept attention yields significant improvements across all tasks. On Pascal VOC, adding concept guidance yields a massive performance jump from 50.2\% to 63.8\% mIoU, demonstrating that while frozen diffusion features contain rich semantics, they lack the precise spatial localization required for dense prediction. Similarly, on NYU Depth, the concept module reduces relative error (AbsRel) from 0.478 to 0.431, and on DUTS, it improves F$_m$ from 0.914 to 0.921.

\textbf{Structural vs. Categorical Concepts.}
For Pascal VOC (Table~\ref{tab:voc_concepts}), we investigated whether explicit class enumeration is necessary for semantic disambiguation. We compared a concise, class-agnostic set (``object, background, detail, edges'') against a set containing all 20 Pascal VOC class names. We observe that the generic set (63.8\%) performs comparably to, and slightly better than, the exhaustive class list (63.3\%). 

We attribute this to the distinct roles of feature extraction and attention. The primary role of the concept attention module in our pipeline is spatial localization, guiding the decoder to the precise boundaries of foreground entities. The broad concept ``object'' generates a dense, high-confidence attention map covering all foreground elements, providing a robust spatial prior. In contrast, specific class tokens generate sparse attention maps that are often empty if the class is absent, diluting the spatial signal with noise without adding discriminative value that isn't already present in the feature embeddings. Thus, we define our final concept sets based on structural utility rather than categorical exhaustiveness.

\textbf{Geometric Priors for Depth and Saliency.}
This observation extends to geometric tasks, where concepts describing spatial relationships prove most effective. On NYU Depth (Table~\ref{tab:nyu_concepts}), adding geometric descriptors like ``near'' and ``far'' to the standard ``depth'' token provided the best performance (AbsRel 0.431), outperforming semantic scene descriptions like ``furniture'' or ``floor''. Similarly, for salient object detection (Table~\ref{tab:duts_single_concepts}), explicit boundary-related terms like ``edges'', ``salient'', and ``contour'' (MaxF 0.921) outperformed generic scene descriptors. This confirms that concept maps function best as structural priors that align with the physical definitions of the task rather than semantic labels.

\section{Token Aggregation Module Architecture}

\begin{figure}[h!]
  \centering
  \includegraphics[width=\linewidth]{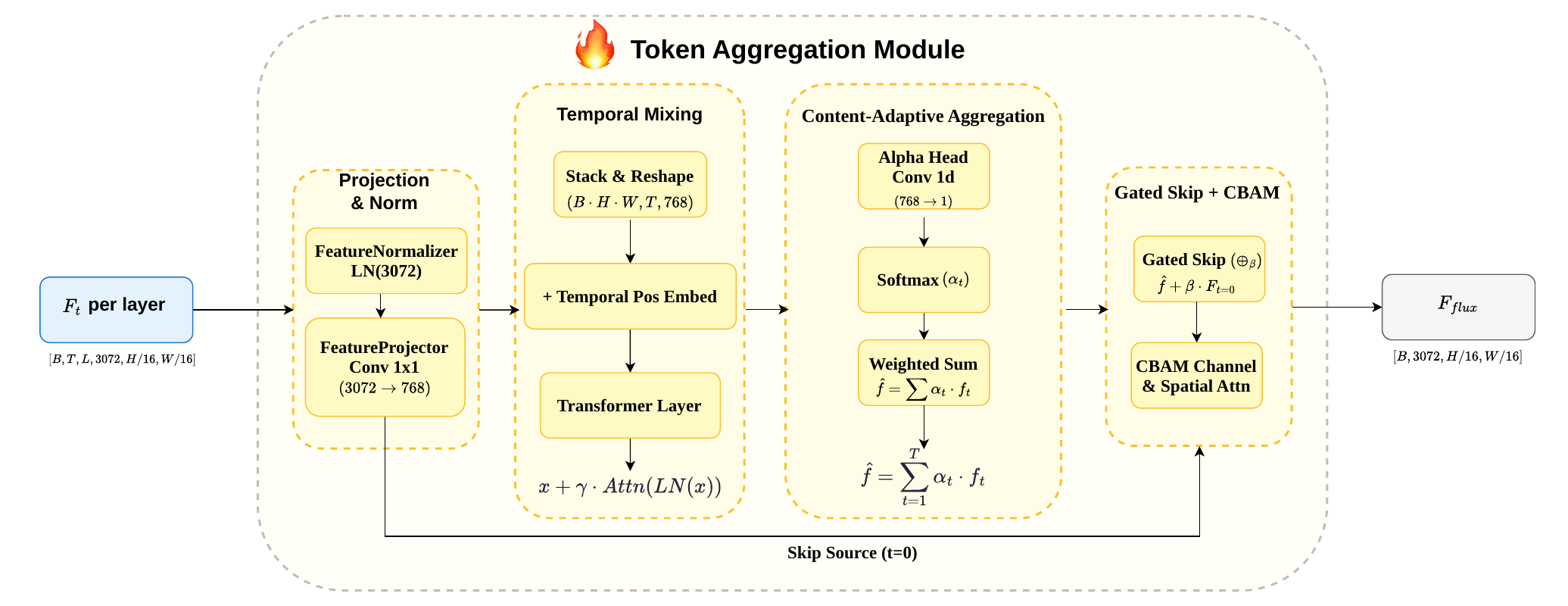}
  
  \caption{\textbf{Token Aggregation Module.} Projected multi-timestep features are aggregated using a temporal transformer, followed by skip-projection and CBAM refinement to produce the final fused token representation.}
  \label{fig:agg_module}
\end{figure}

We provide the architectural details of the token aggregation module, which aggregates information across the temporal dimension while maintaining spatial specificity.

\textbf{Feature Projection and Sequence Formation.} The module is applied independently to each extracted DiT layer. For a given layer, the module accepts feature maps $\{F_t\}_{t \in \mathcal{T}}$ from $|\mathcal{T}|=4$ distinct timesteps. To reduce the channel dimensionality from $C=3072$ to the hidden dimension $D=768$, we employ a two stage preparation. First, a FeatureNormalizer applies learnable, timestep specific Layer Normalization to address the high variance of diffusion activations. Second, a FeatureProjector uses independent $1\times1$ convolutions for each timestep to map features into the hidden space. These features are stacked and reshaped into temporal sequences $S \in \mathbb{R}^{(B \cdot H \cdot W) \times |\mathcal{T}| \times D}$, treating each spatial location as an independent sequence of length 4.

\textbf{Temporal Transformer Encoder.} We employ a 3 layer Transformer encoder to model non linear temporal relationships. We add learnable temporal positional embeddings $\in \mathbb{R}^{1 \times |\mathcal{T}| \times D}$. To ensure optimization stability, we employ LayerScale in the residual branches. Scaling factors are initialized to small values, specifically $10^{-6}$, so that layers initially act as identity mappings. The transformer uses 8 attention heads to process the temporal dimension.

\textbf{Content Adaptive Aggregation.} The transformer output is passed to a prediction head, which is a $1\times1$ convolution mapping $D \to 1$, that generates per pixel mixing logits. A softmax operation produces weights $\alpha \in \mathbb{R}^{|\mathcal{T}|}$ such that $\sum_{t=1}^{T} \alpha_t = 1$ at every spatial location. We compute the weighted sum of the features using $\alpha$, allowing the network to dynamically prioritize early or late timesteps based on local content. Finally, we apply a gated skip connection from the projected features of the cleanest timestep ($t=0$), controlled by a learnable scalar initialized to $0.1$. The aggregated features are refined with a Convolutional Block Attention Module (CBAM) before downstream decoding.

\section{Extension to Other DiT Backbones}

To demonstrate generalizability, we apply MMDiff to Stable Diffusion 3.5 without any architectural or training modifications. Table~\ref{tab:sd35_results} shows strong performance across all three tasks, confirming that our multi-timestep aggregation and concept-driven attention transfer effectively to other DiT backbones.

\begin{table}[h]
\centering
\caption{\textbf{MMDiff with SD 3.5 backbone.} Performance across tasks demonstrates our method transfers to other DiT architectures without architectural or training changes. All encoders are frozen.}
\label{tab:sd35_results}
\small
\resizebox{\textwidth}{!}{
\begin{tabular}{l c ccc ccccc}
\toprule
 & \multicolumn{1}{c}{\textbf{PASCAL VOC 2012}}
 & \multicolumn{3}{c}{\textbf{DUTS}}
 & \multicolumn{5}{c}{\textbf{NYU Depth v2}} \\
\cmidrule(lr){2-2} \cmidrule(lr){3-5} \cmidrule(lr){6-10}
\textbf{Features}
& mIoU $\uparrow$ 
& $S_m \uparrow$ & $F_m \uparrow$ & MAE $\downarrow$
& AbsRel $\downarrow$ & RMSE $\downarrow$ & $\delta_1$ $\uparrow$ & $\delta_2$ $\uparrow$ & $\delta_3$ $\uparrow$ \\
\midrule
SD 3.5 only
& 70.14 
& 0.898 & 0.912 & 0.026
& 0.1305 & 0.401 & 0.940 & 0.973 & 0.982 \\
SD 3.5 + DINO-v3
& 84.09
& 0.915 & 0.915 & 0.019
& 0.1296 & 0.394 & 0.948 & 0.975 & 0.982 \\
\bottomrule
\end{tabular}
}
\end{table}

SD 3.5 features alone achieve 70.14\% mIoU on PASCAL VOC and competitive saliency and depth results; combined with DINO-v3, performance improves substantially across all tasks, again confirming the complementarity of generative and discriminative features.
Compared to FLUX, SD 3.5 performs worse in the diffusion-only setting. Since our encoder is frozen, the quality of extracted features is directly bounded by the generator's internal representations -- stronger generators produce richer features, which in turn yield better dense predictions. This gap can therefore be interpreted as an indirect measure of relative generator quality, and suggests that as diffusion transformers continue to improve, \method performance will improve correspondingly without any changes to the framework.

\section{Qualitative Results}

\begin{figure}[t]
  \centering
  \includegraphics[width=\linewidth]{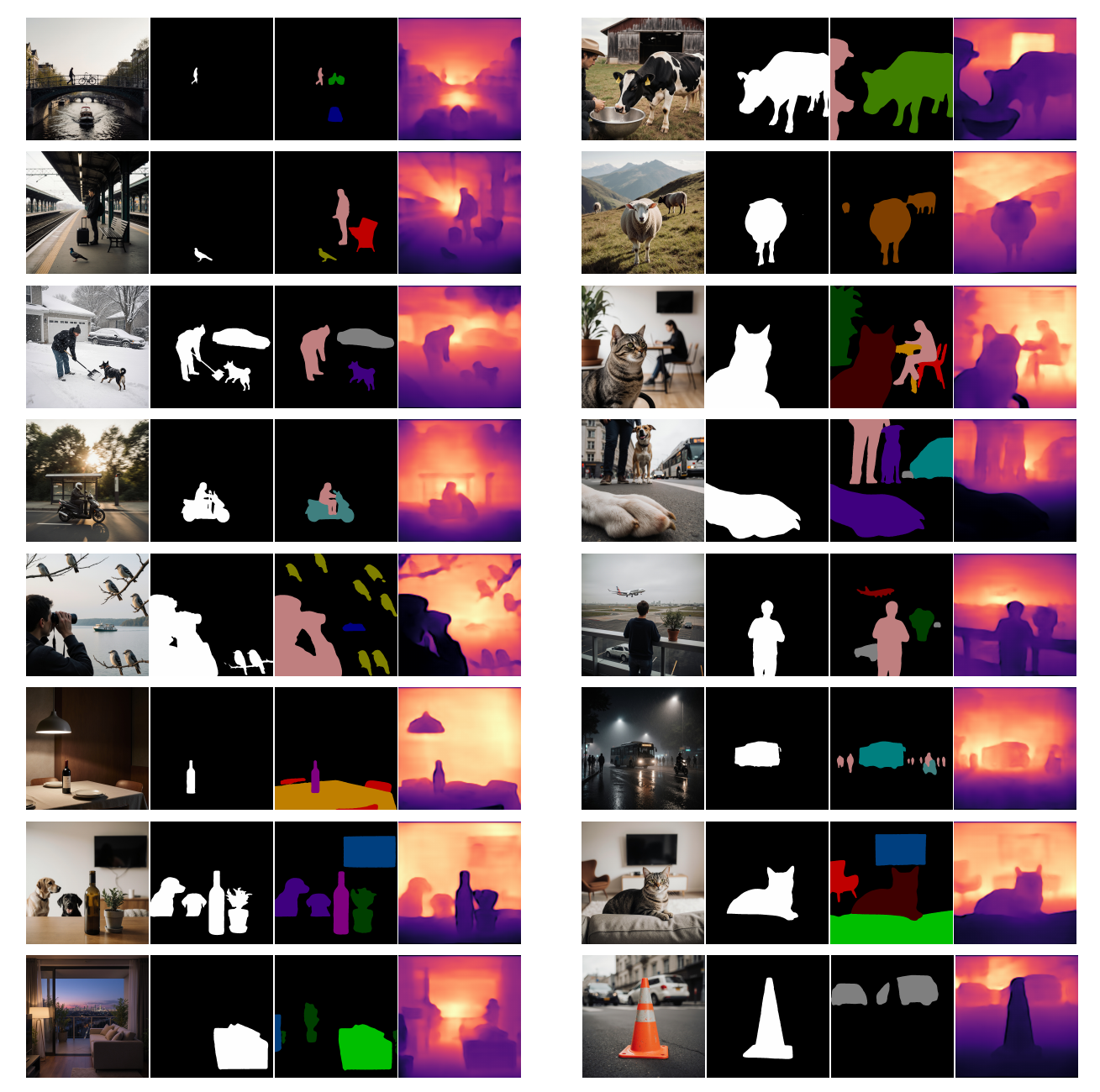}
  
  \caption{Qualitative results from \method across diverse scenes. Each group shows the synthetic input image, predicted binary mask, predicted semantic segmentation, and predicted depth map.}
  \label{fig:synt_qual-1}
\end{figure}

We show extended qualitative predictions on synthetic data generated across a wide variety of scenes (Fig.~\ref{fig:synt_qual-1}).
The examples highlight the model’s ability to simultaneously infer binary masks, semantic segmentation, and depth from complex visual inputs.

We also present qualitative comparisons across all three tasks in Fig.~\ref{fig:nyu_qualitative_grid}, Fig.~\ref{fig:duts_qualitative_grid}, and Fig.~\ref{fig:pascal_qualitative_grid}. For depth estimation, \method produces smooth, spatially coherent depth maps with sharp object boundaries, while VPD \cite{zhao2023vpd} exhibits over-smoothing and DHF \cite{luo2024diffusionhyperfeaturessearchingtime} generates noisy predictions. On salient object detection, our method enables accurate segmentation of challenging cases including multi-object scenes and low-contrast subjects, where baselines produce incomplete or fragmented masks. For semantic segmentation, \method accurately segments diverse categories with clean boundaries, successfully handling small objects and partial occlusions where other methods fail.

\setlength{\cellw}{0.128\columnwidth}
\setlength{\rowh}{1.7cm}

\begin{figure}[t]
  \centering
  \setlength{\tabcolsep}{1pt}
  \renewcommand{\arraystretch}{0}
  \begin{tabular}{@{}*{7}{m{\cellw}}@{}}

    \img{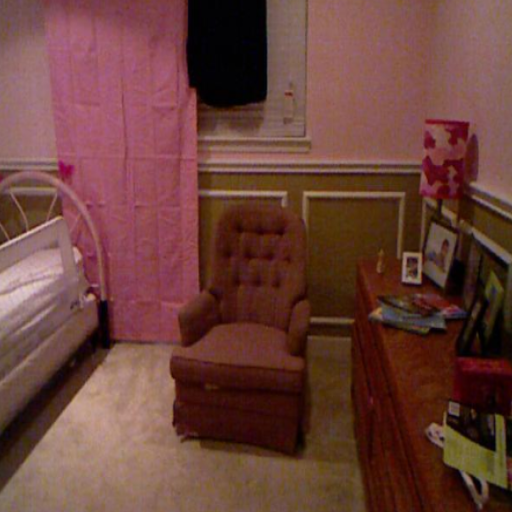} &
    \img{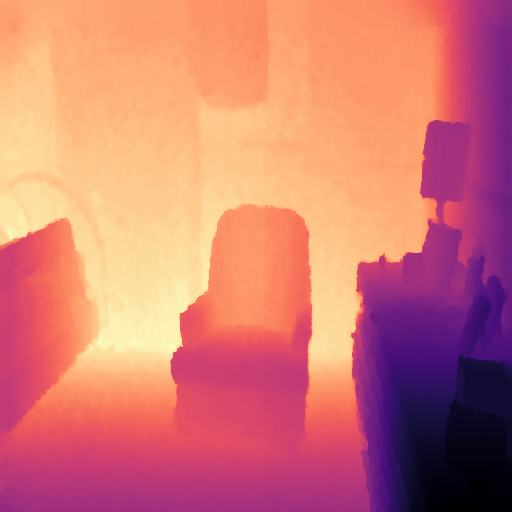} &
    \img{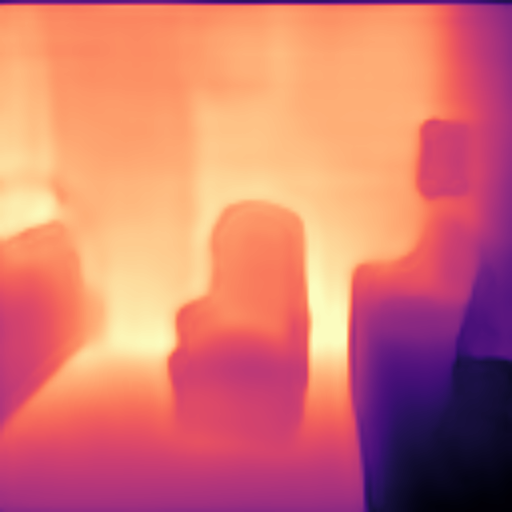} &
    \img{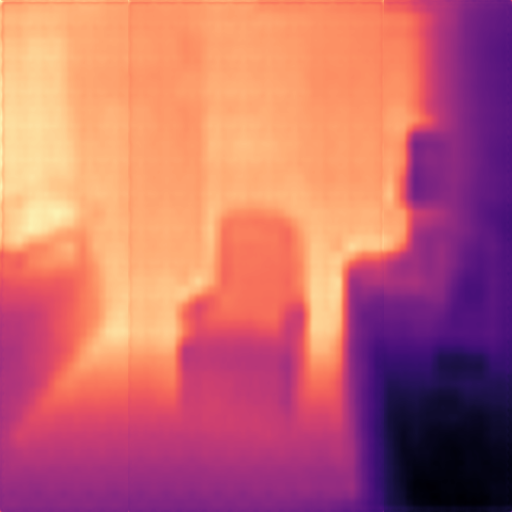} &
    \img{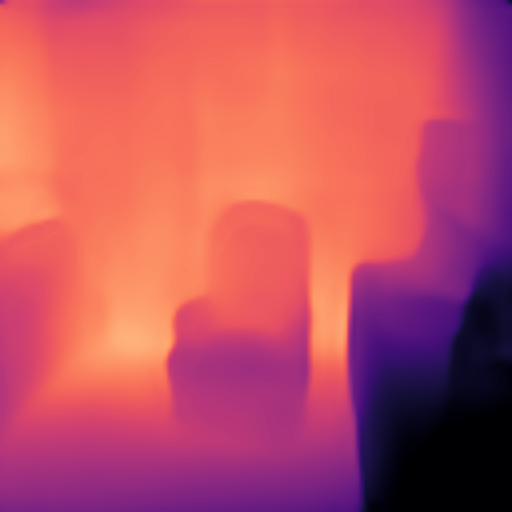} &
    \img{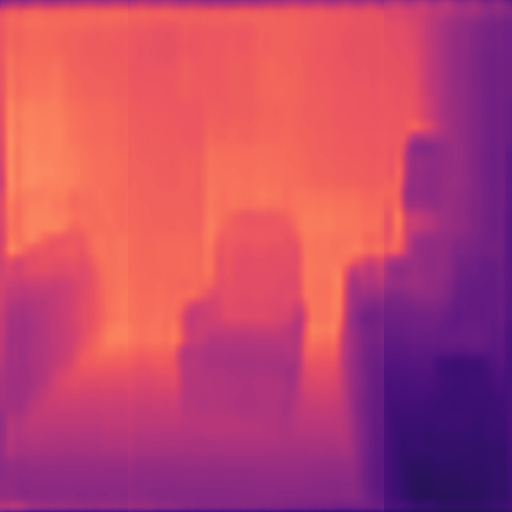} &
    \img{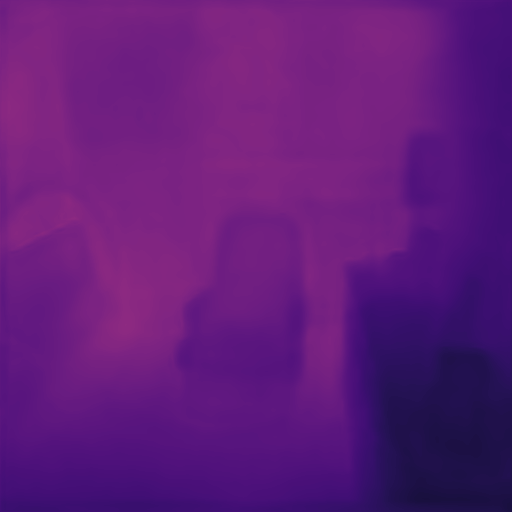} \\[-2pt]

    \img{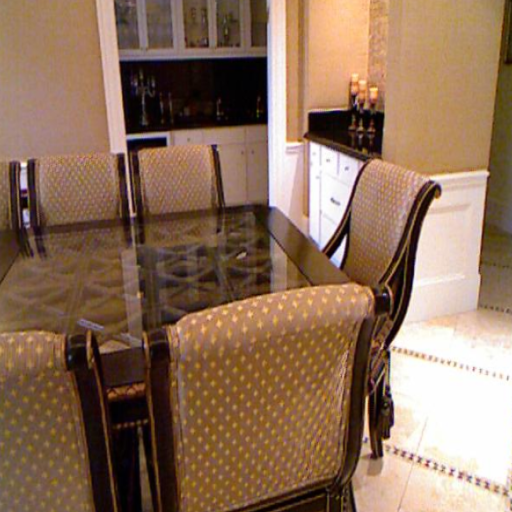} &
    \img{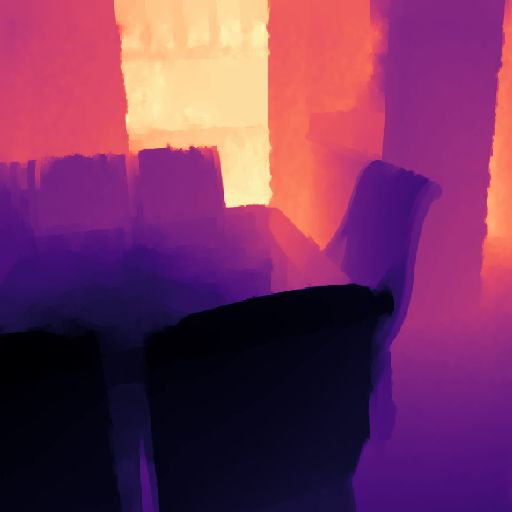} &
    \img{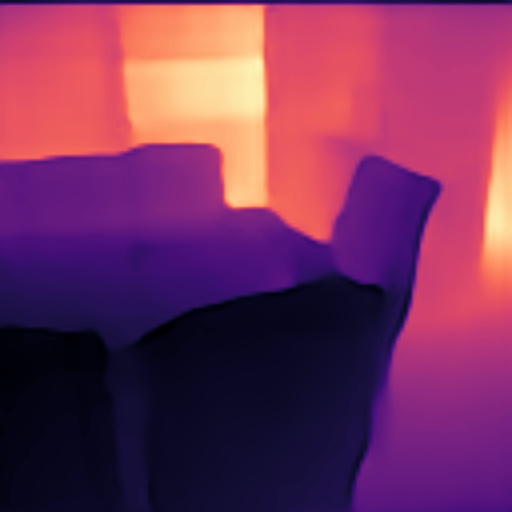} &
    \img{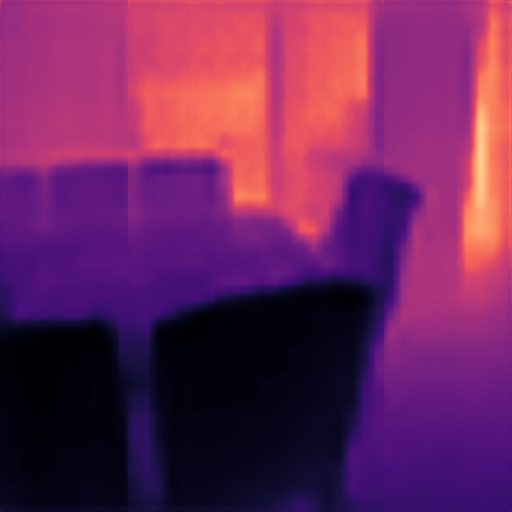} &
    \img{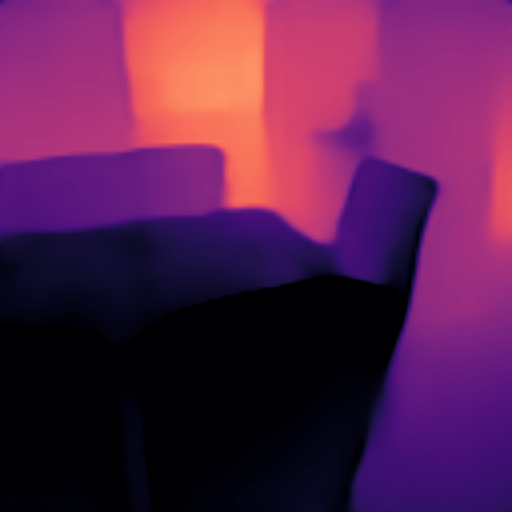} &
    \img{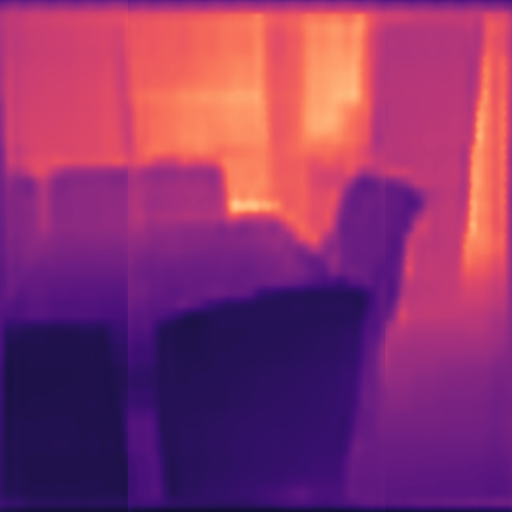} &
    \img{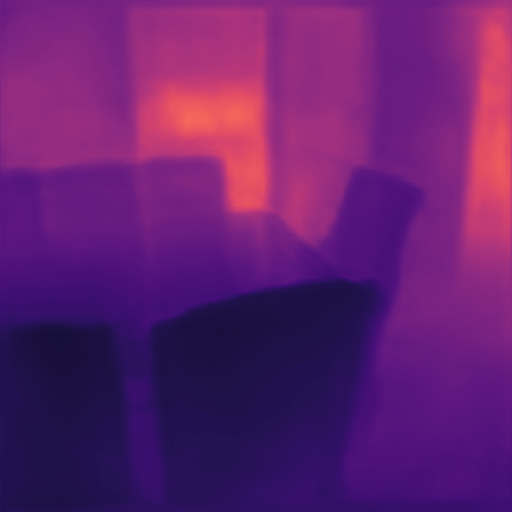} \\[-2pt]

    \img{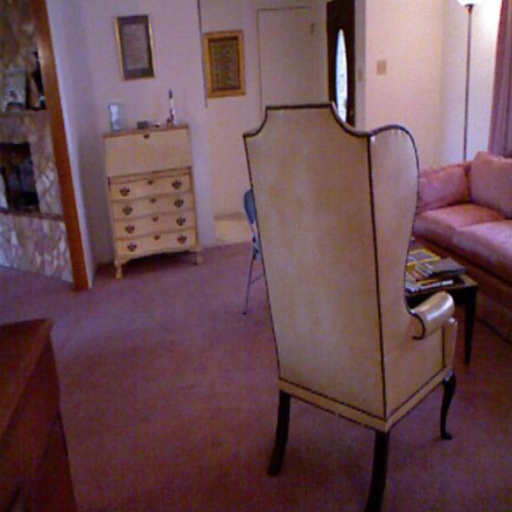} &
    \img{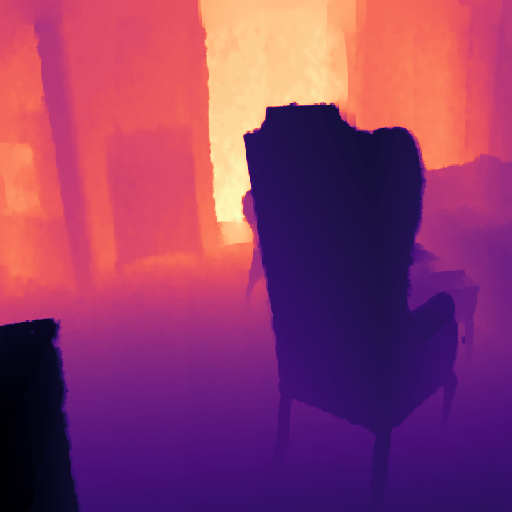} &
    \img{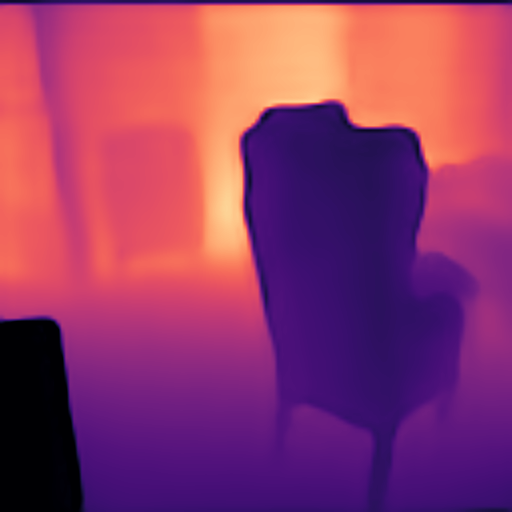} &
    \img{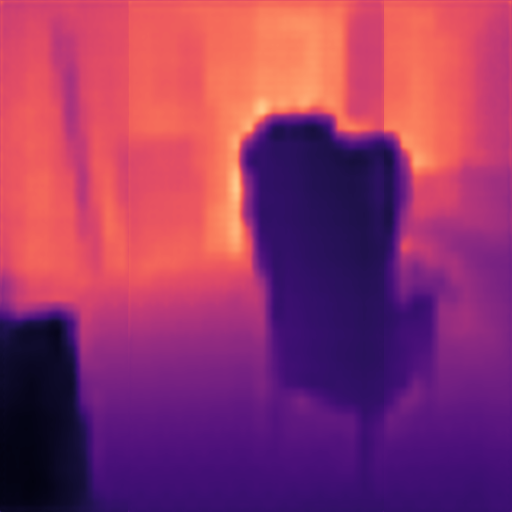} &
    \img{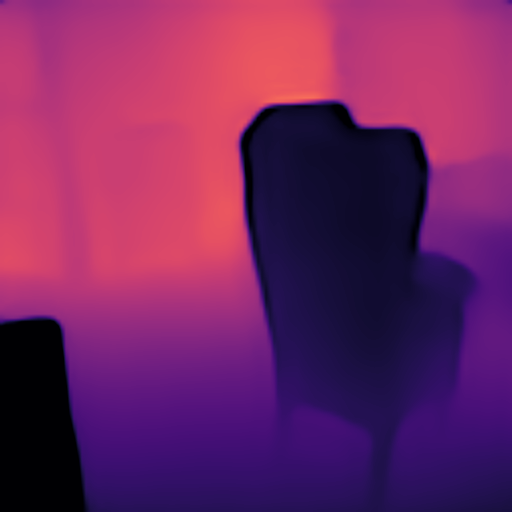} &
    \img{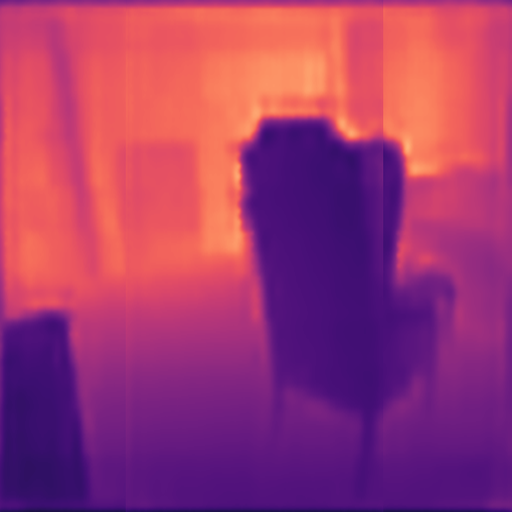} &
    \img{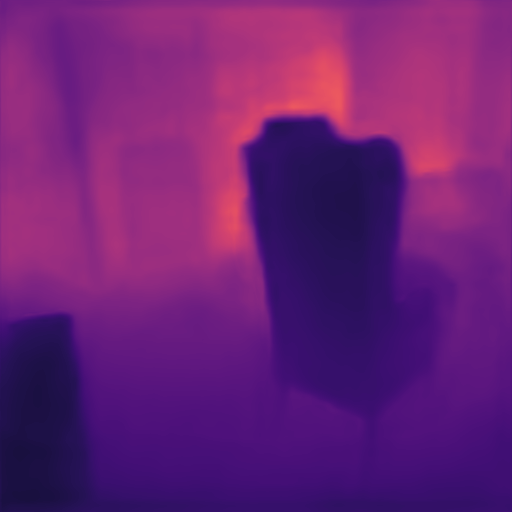} \\[2pt]

    \img{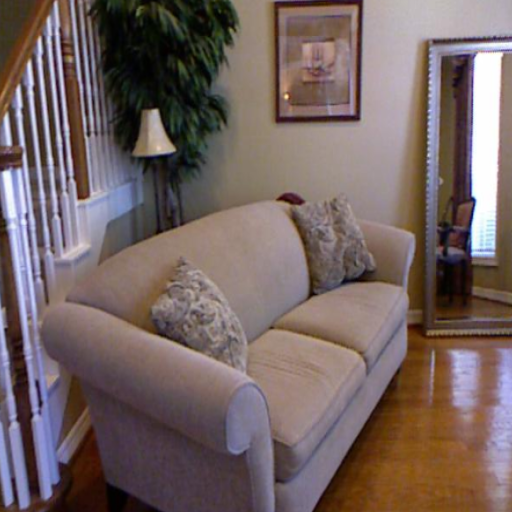} &
    \img{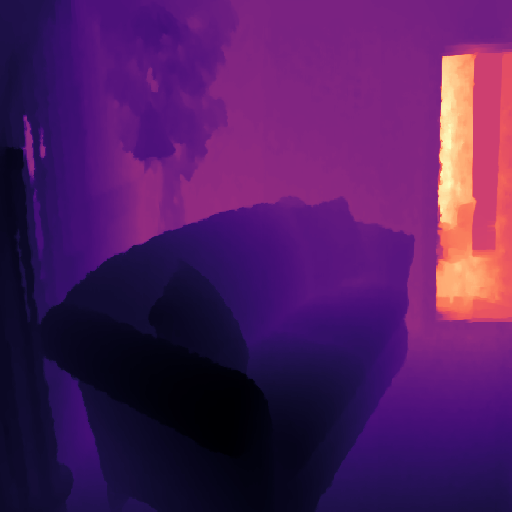} &
    \img{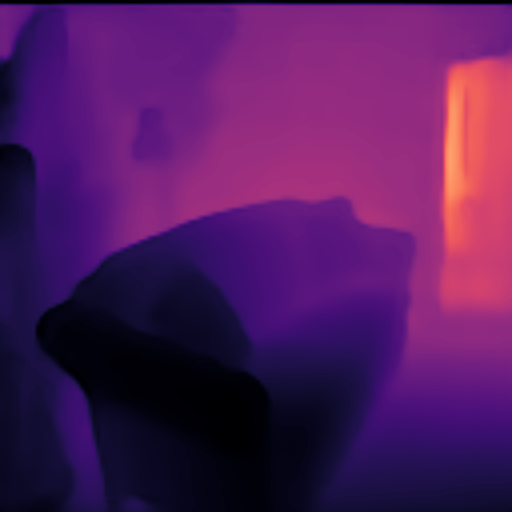} &
    \img{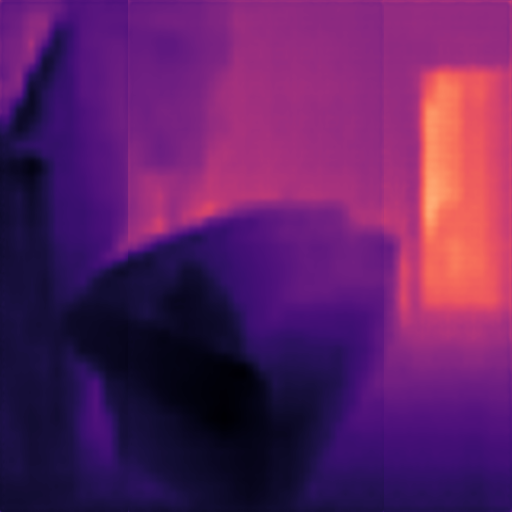} &
    \img{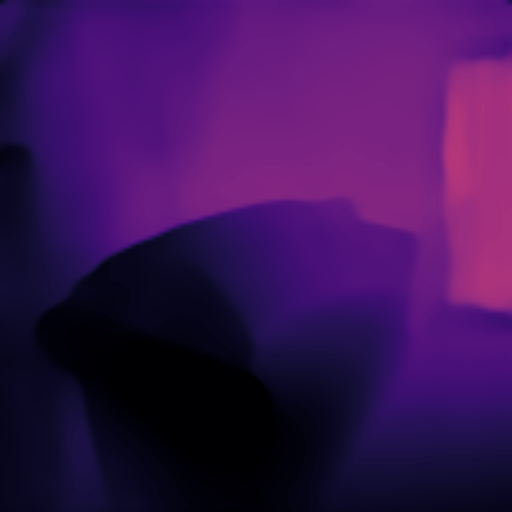} &
    \img{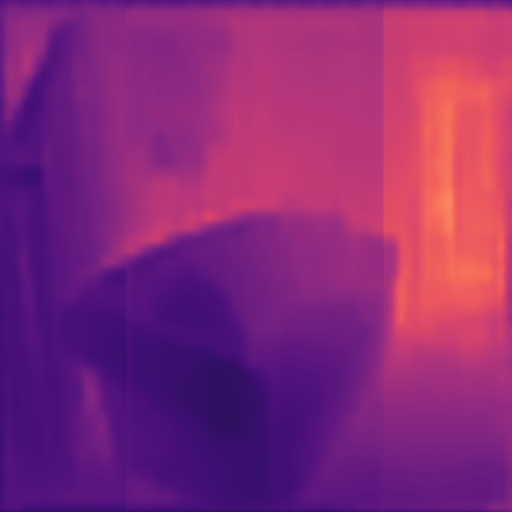} &
    \img{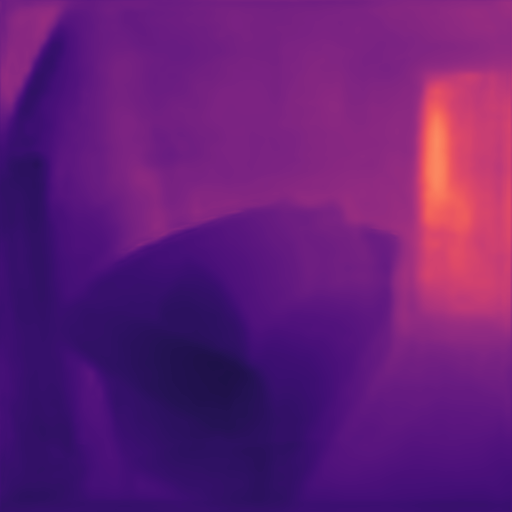} \\[2pt]

    \img{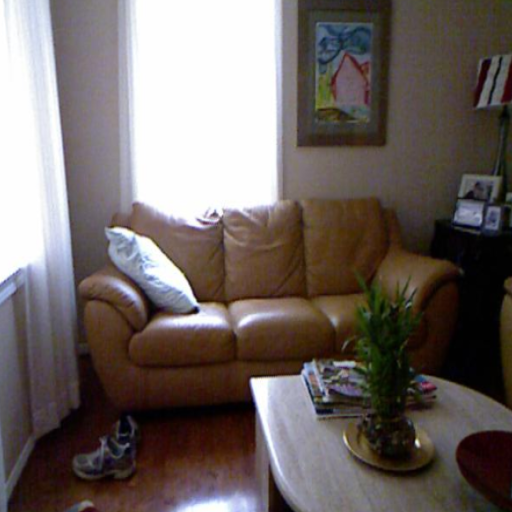} &
    \img{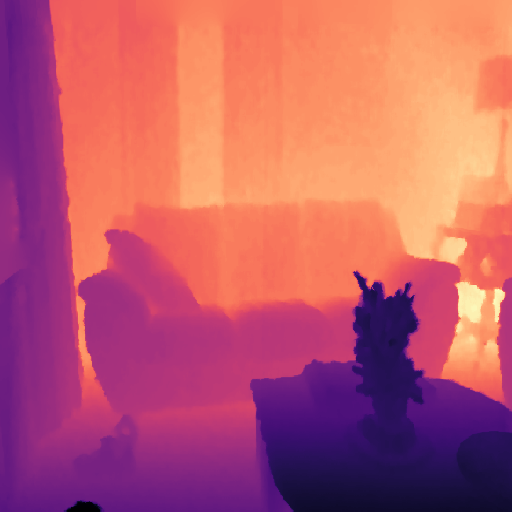} &
    \img{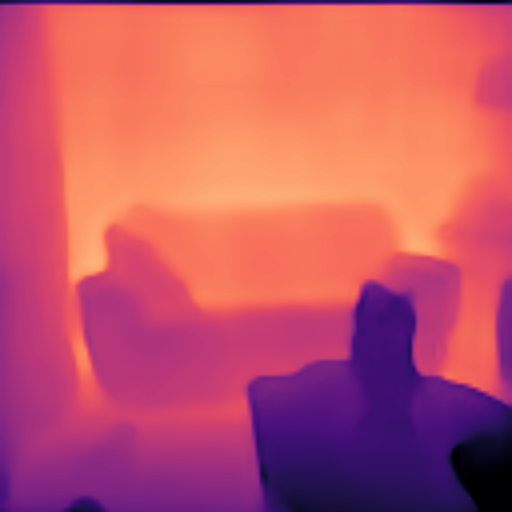} &
    \img{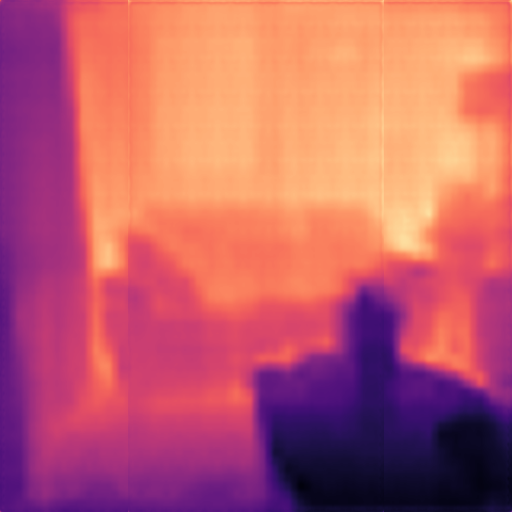} &
    \img{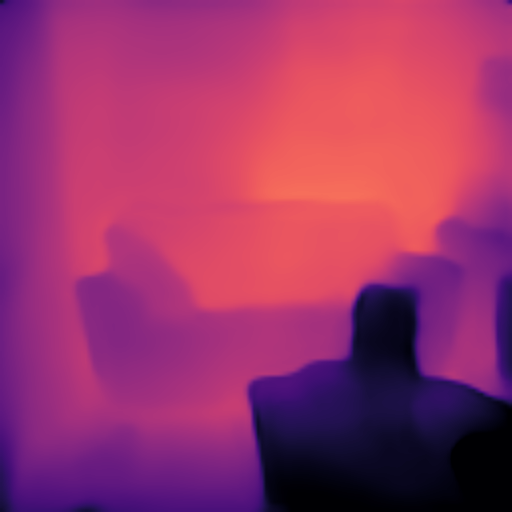} &
    \img{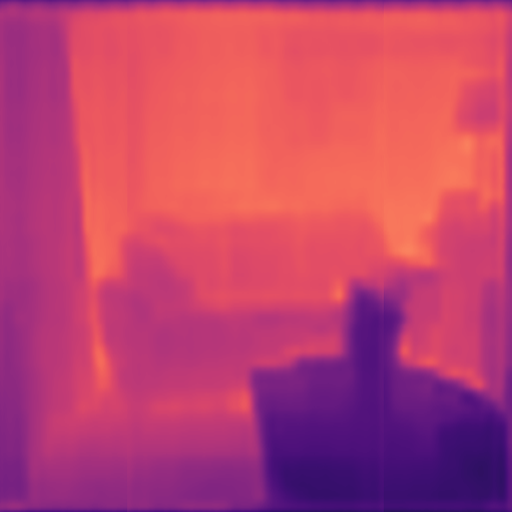} &
    \img{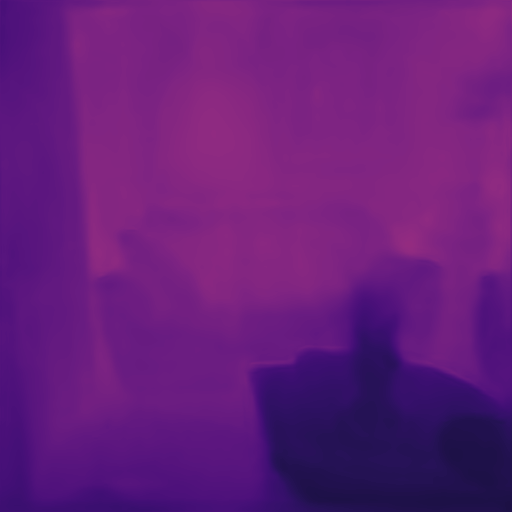} \\[2pt]

    \img{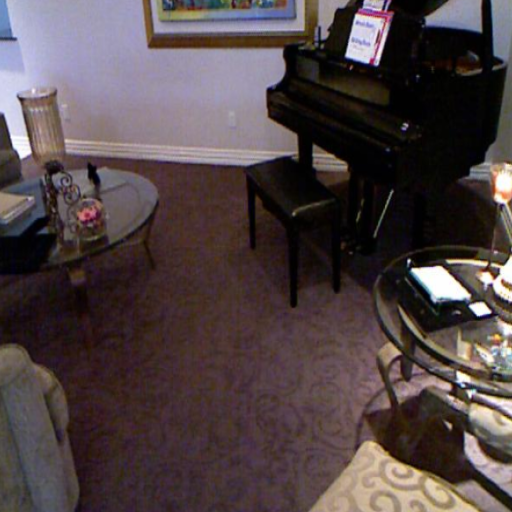} &
    \img{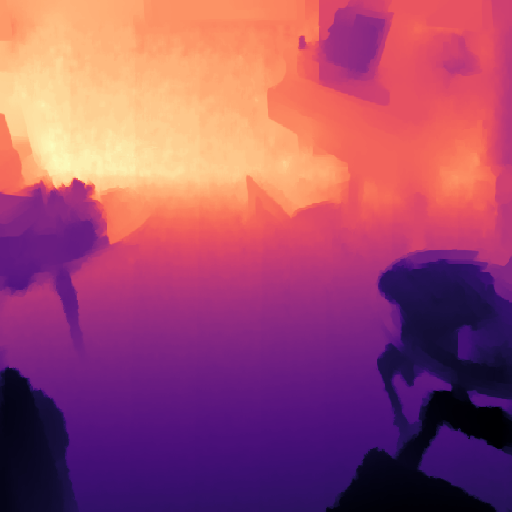} &
    \img{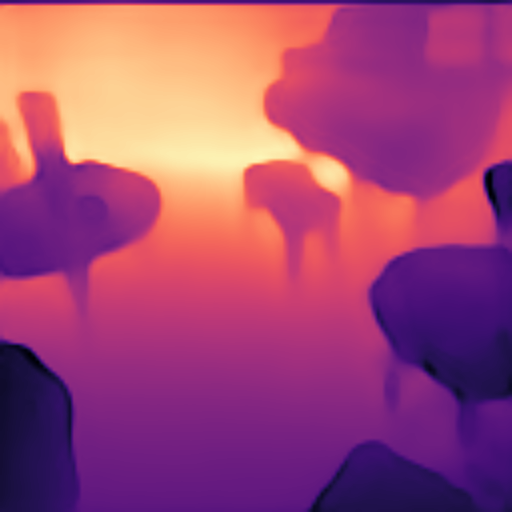} &
    \img{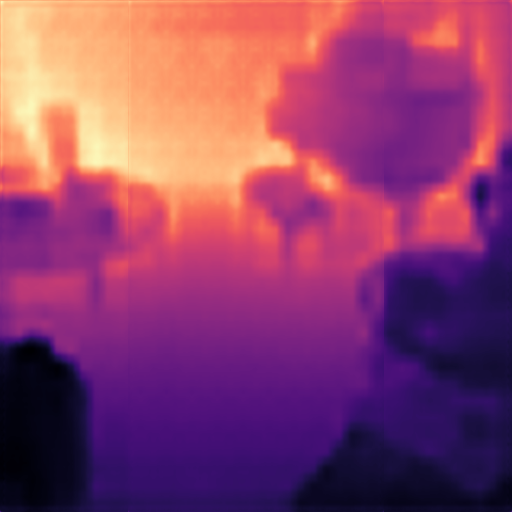} &
    \img{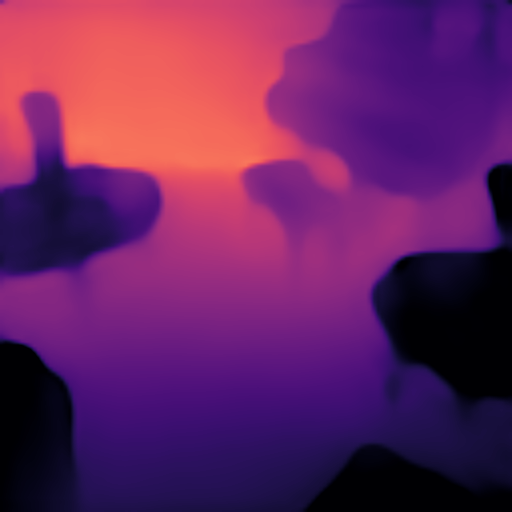} &
    \img{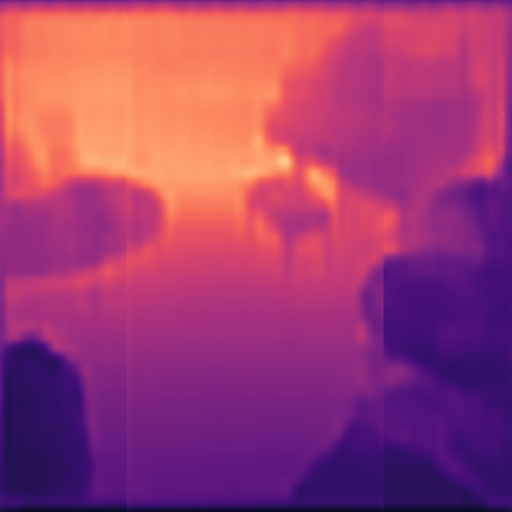} &
    \img{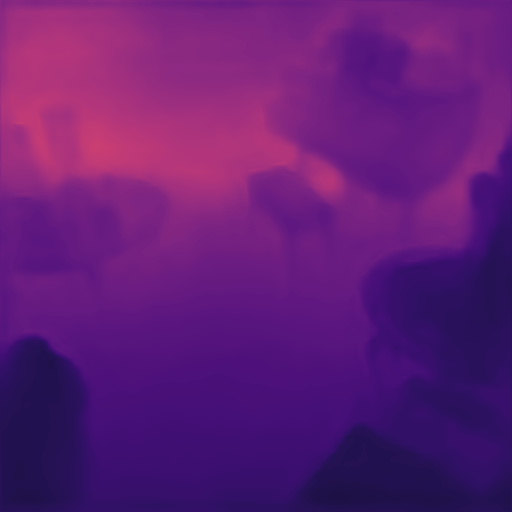} \\[4pt]

    \parbox[c]{\cellw}{\centering\fontsize{6}{6}\selectfont\textbf{\mbox{Image}}} &
    \parbox[c]{\cellw}{\centering\fontsize{6}{6}\selectfont\textbf{\mbox{GT}}} &
    \parbox[c]{\cellw}{\centering\fontsize{6}{6}\selectfont\textbf{\mbox{\method}}} &
    \parbox[c]{\cellw}{\centering\fontsize{6}{6}\selectfont\textbf{\mbox{VPD~\cite{zhao2023vpd}}}} &
    \parbox[c]{\cellw}{\centering\fontsize{6}{6}\selectfont\textbf{\mbox{DINO-v3~\cite{siméoni2025dinov3}}}} &
    \parbox[c]{\cellw}{\centering\fontsize{6}{6}\selectfont\textbf{\mbox{DHF~\cite{luo2024diffusionhyperfeaturessearchingtime}}}} &
    \parbox[c]{\cellw}{\centering\fontsize{6}{6}\selectfont\textbf{\mbox{DatasetDM~\cite{wu2023datasetdmsynthesizingdataperception}}}} \\

  \end{tabular}
  \caption{
    \textbf{Qualitative comparisons on NYU Depth V2 examples.}
  }
  \label{fig:nyu_qualitative_grid}
\end{figure}

\begin{figure}[t]
  \centering
  \setlength{\tabcolsep}{1pt}
  \renewcommand{\arraystretch}{0}
  \begin{tabular}{@{}*{7}{m{\cellw}}@{}}

    \img{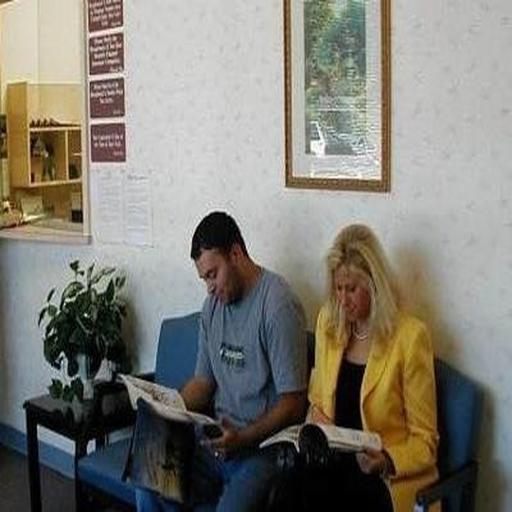} &
    \img{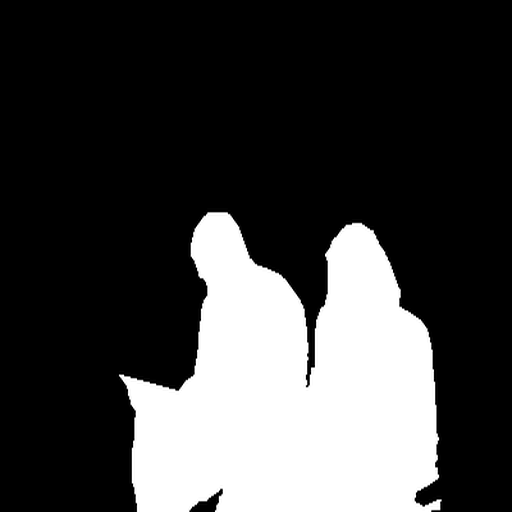} &
    \img{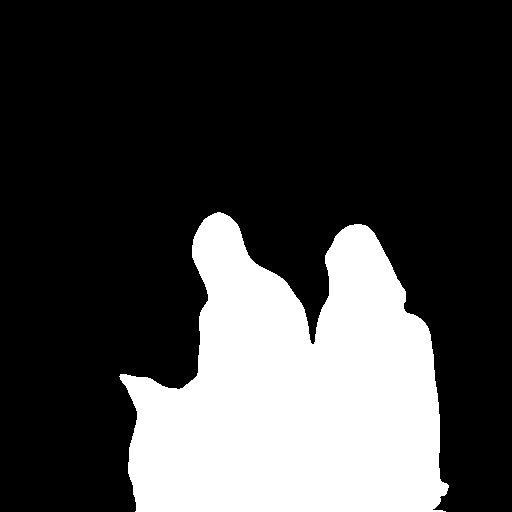} &
    \img{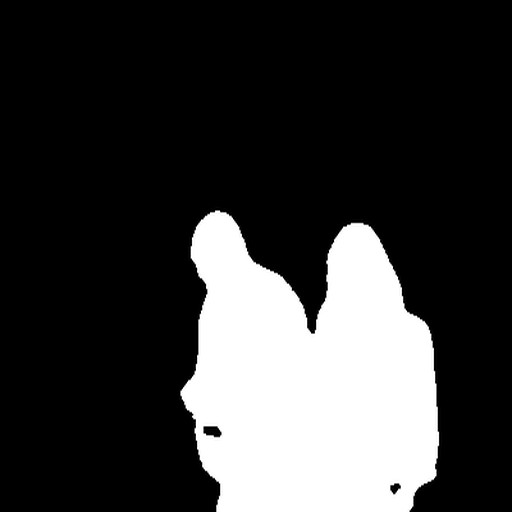} &
    \img{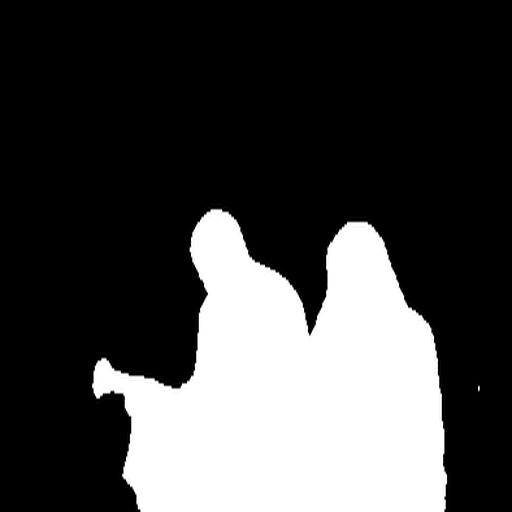} &
    \img{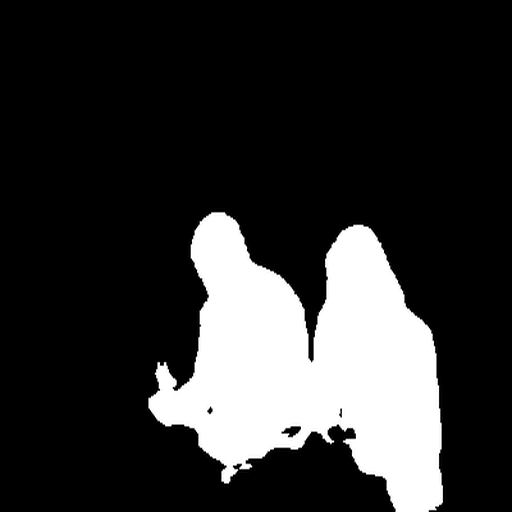} &
    \img{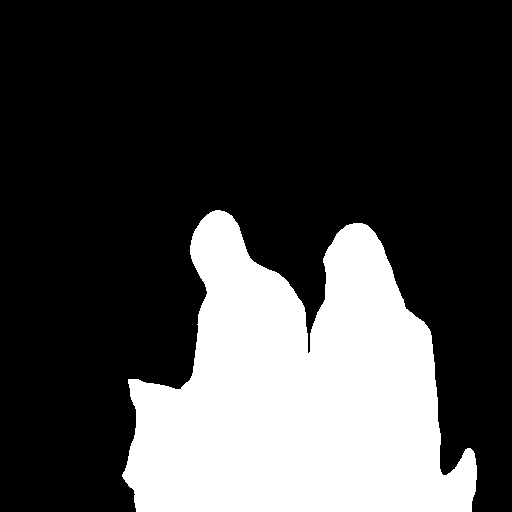} \\[-2pt]

    \img{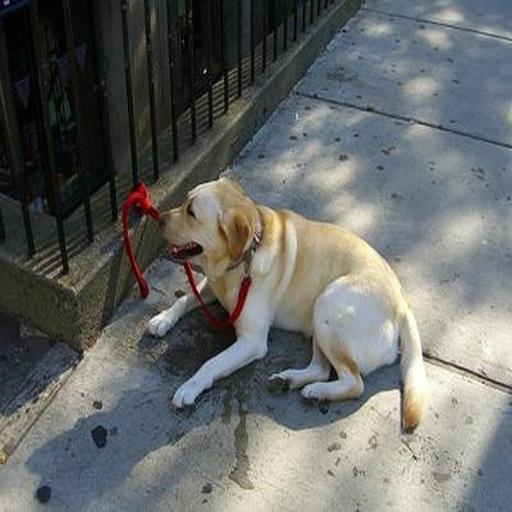} &
    \img{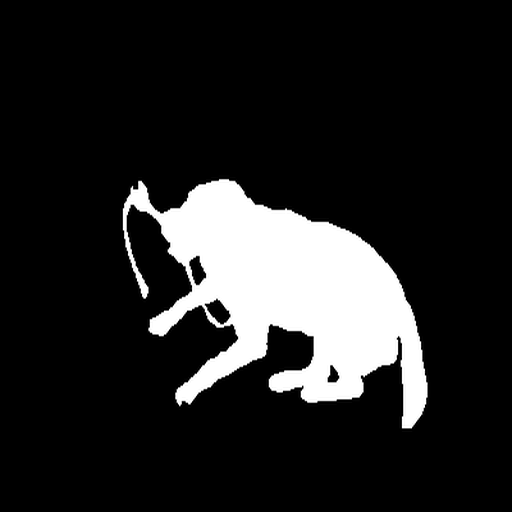} &
    \img{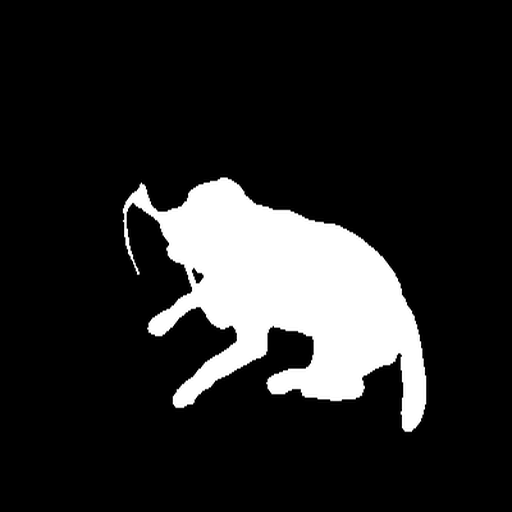} &
    \img{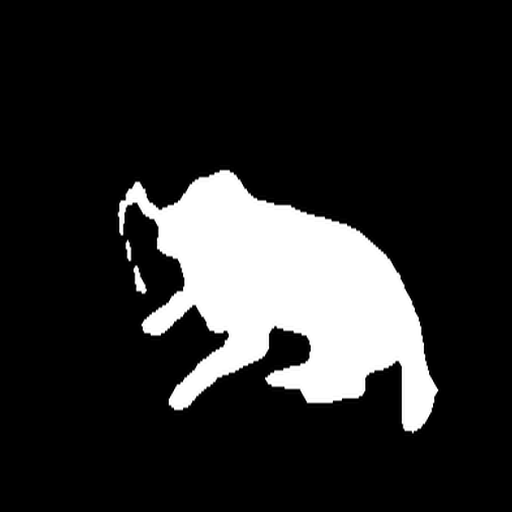} &
    \img{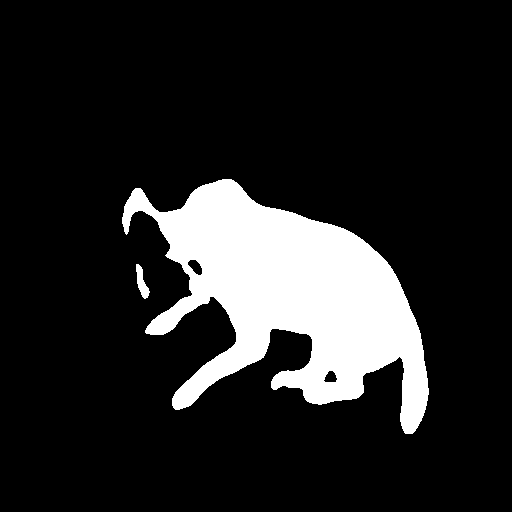} &
    \img{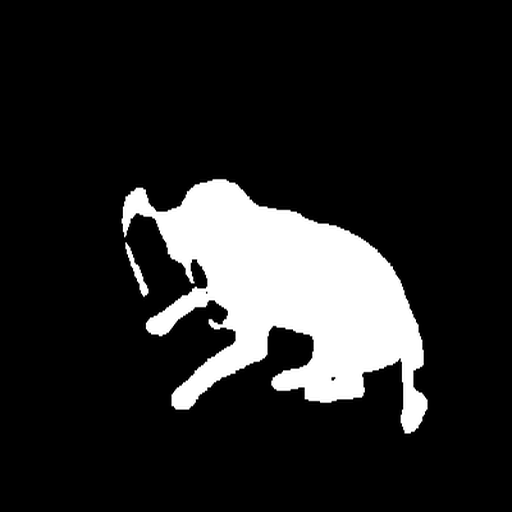} &
    \img{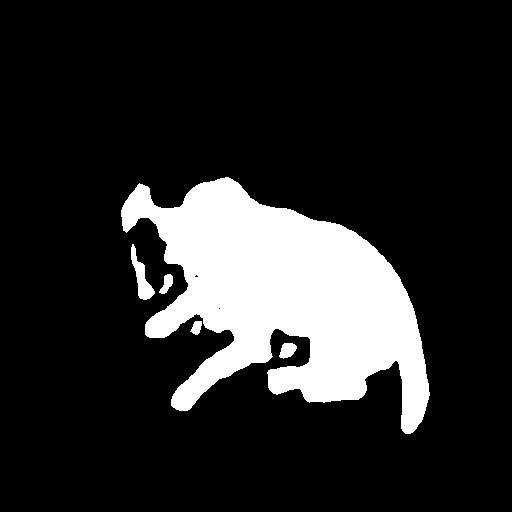} \\[-2pt]

    \img{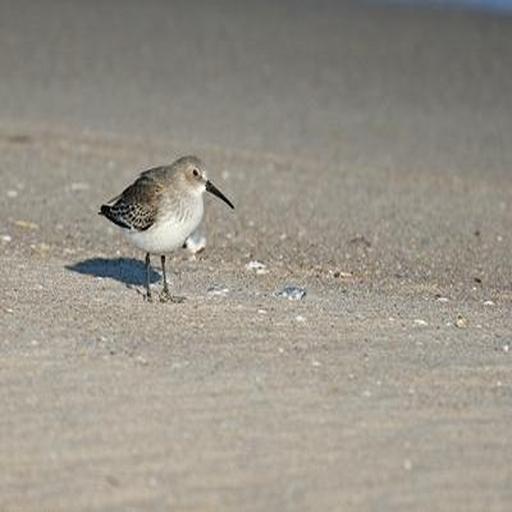} &
    \img{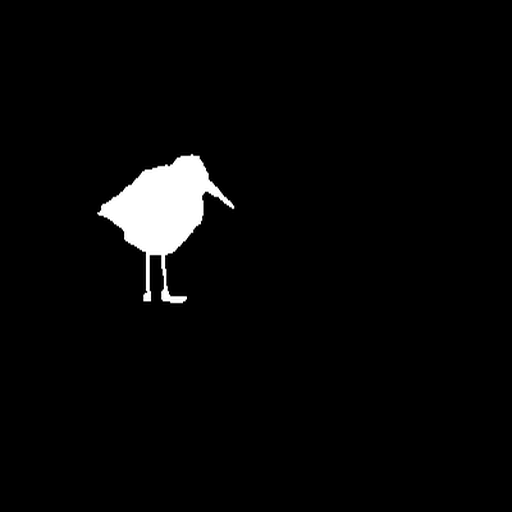} &
    \img{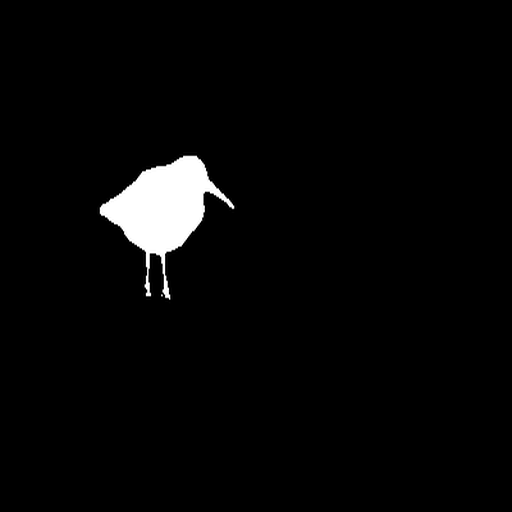} &
    \img{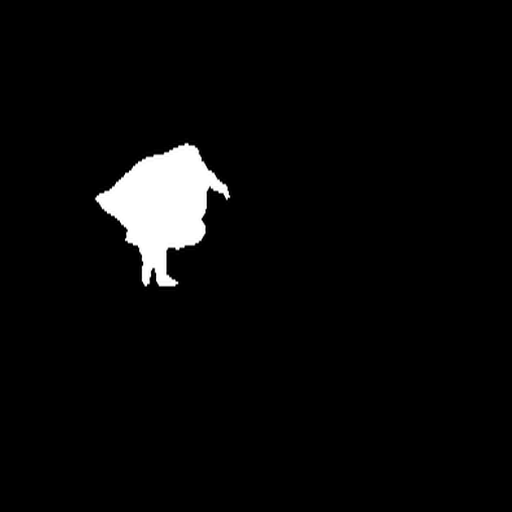} &
    \img{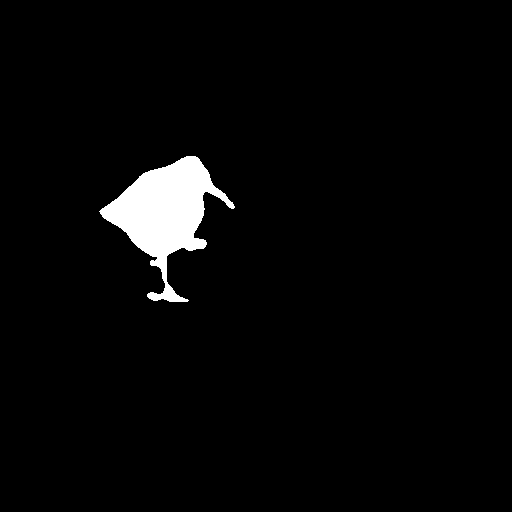} &
    \img{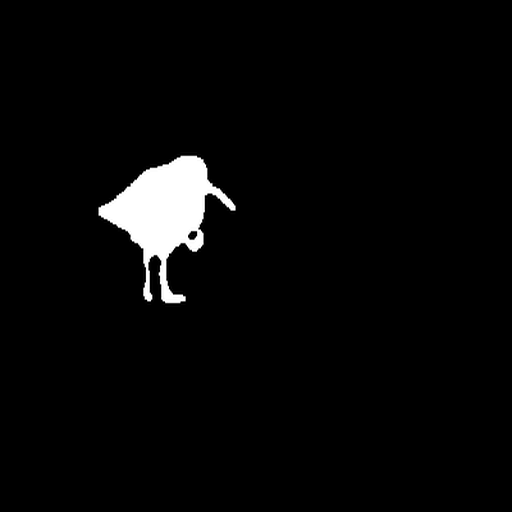} &
    \img{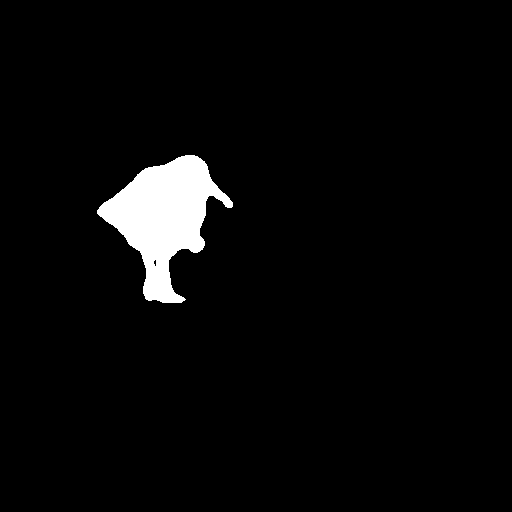} \\[2pt]

    \img{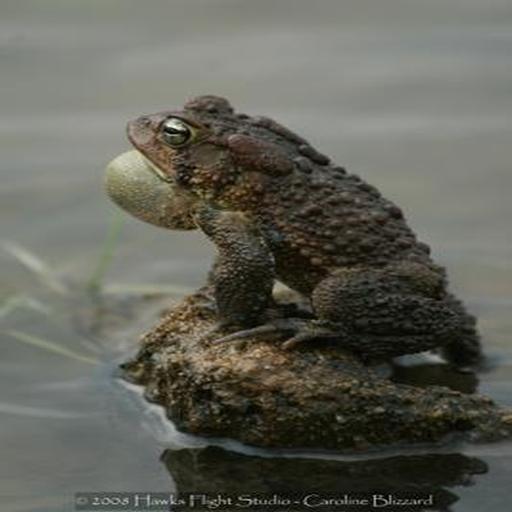} &
    \img{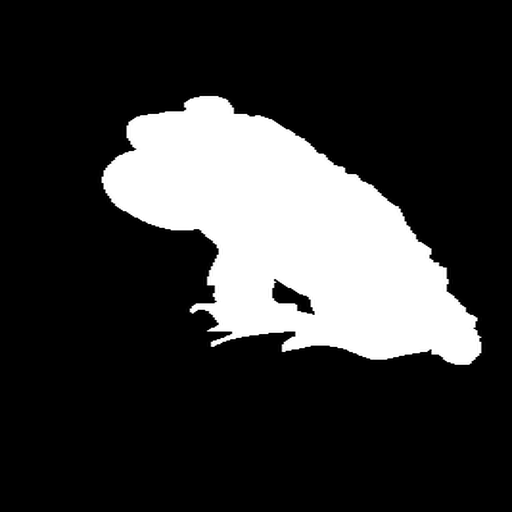} &
    \img{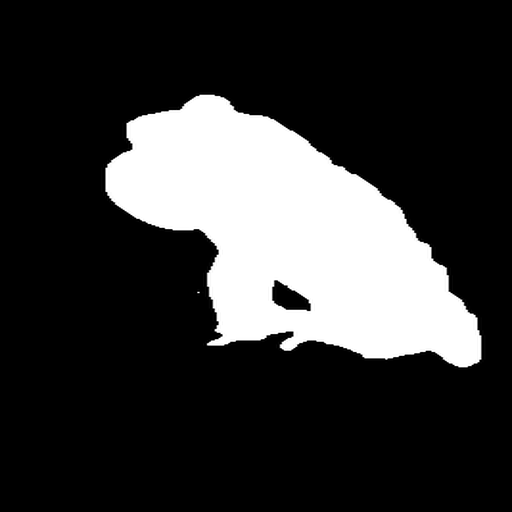} &
    \img{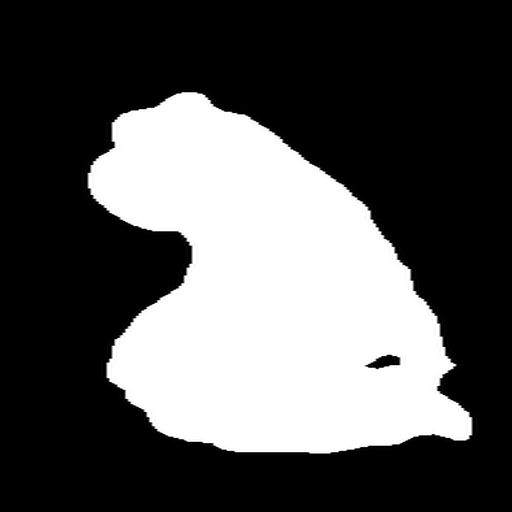} &
    \img{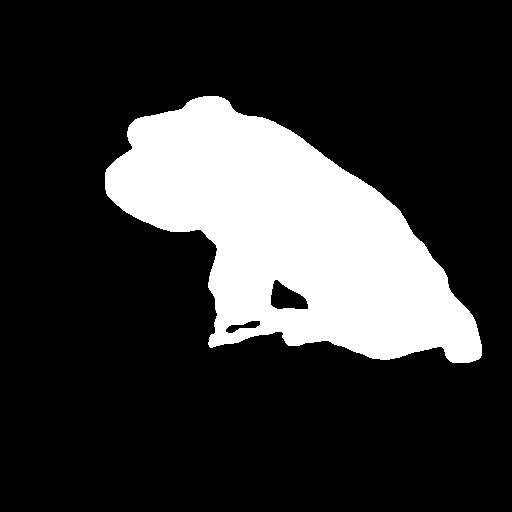} &
    \img{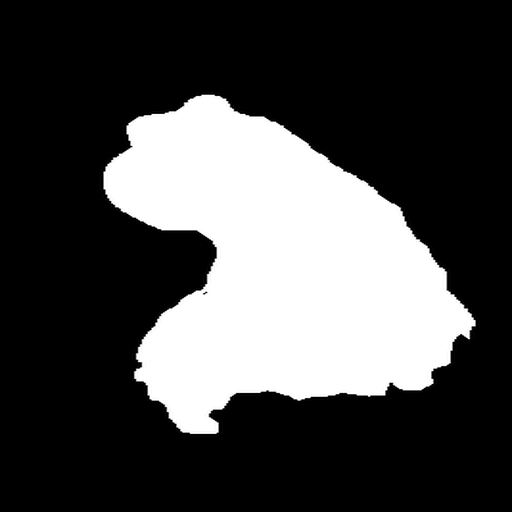} &
    \img{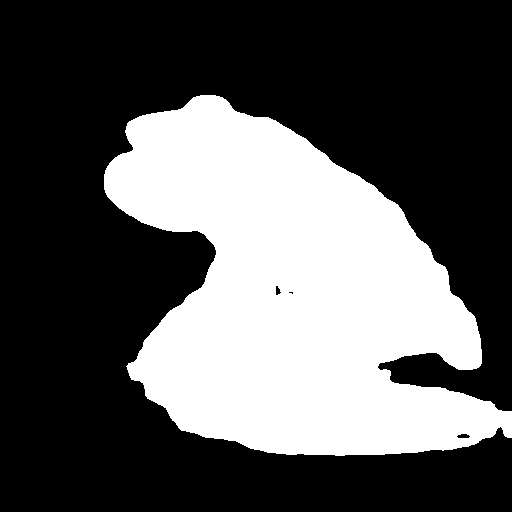} \\[2pt]

    \img{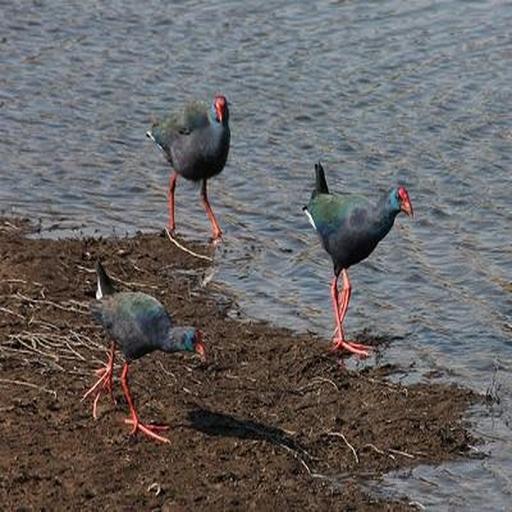} &
    \img{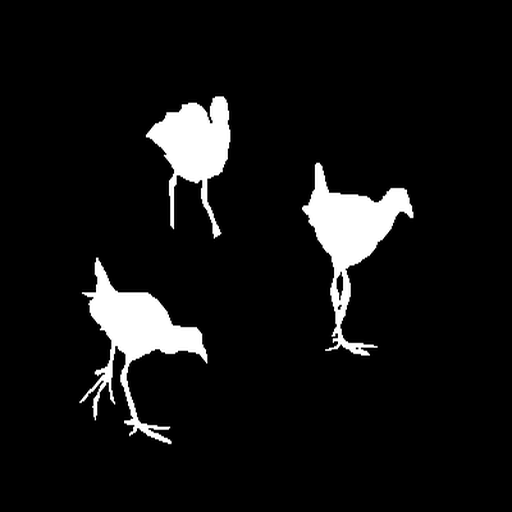} &
    \img{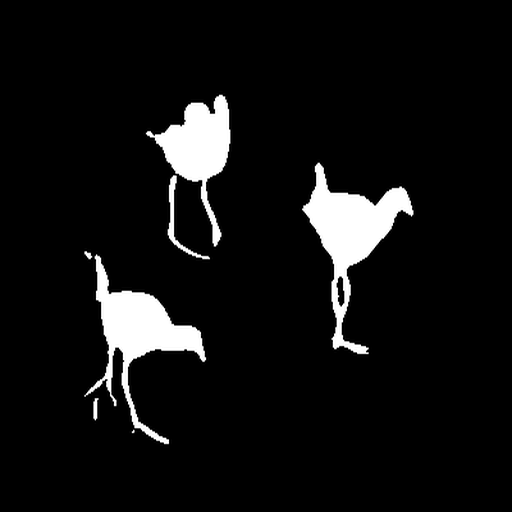} &
    \img{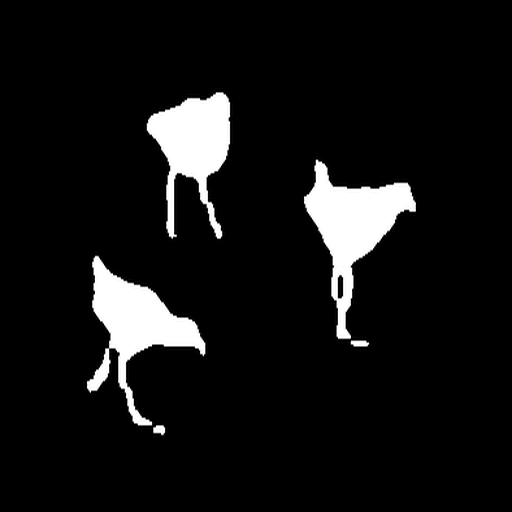} &
    \img{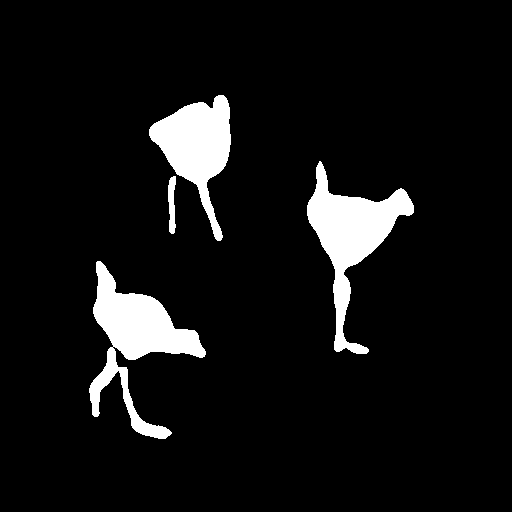} &
    \img{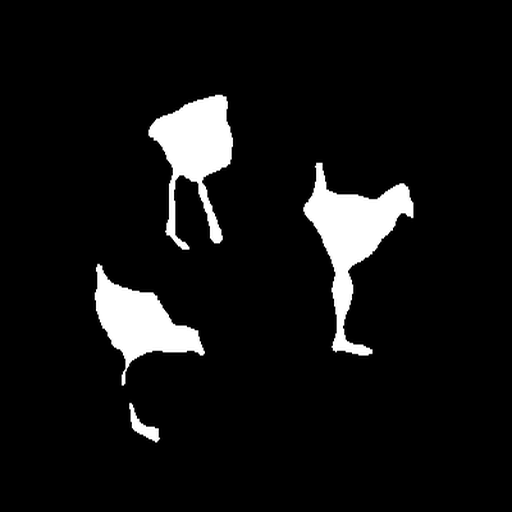} &
    \img{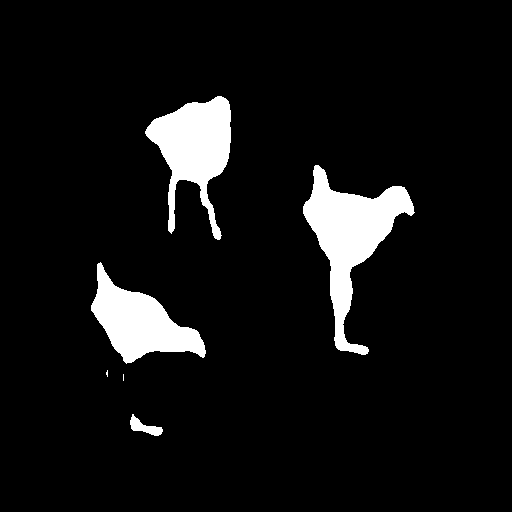} \\[2pt]

    \img{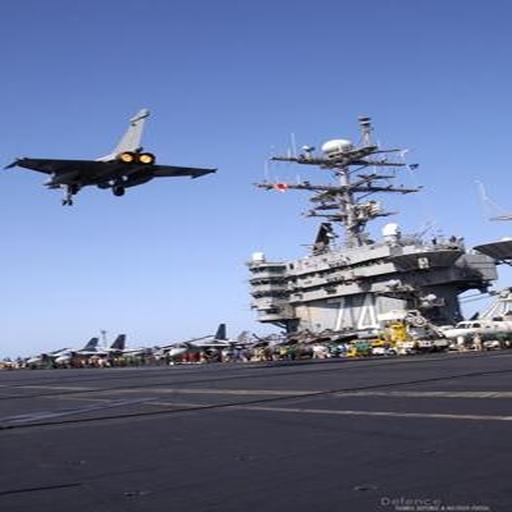} &
    \img{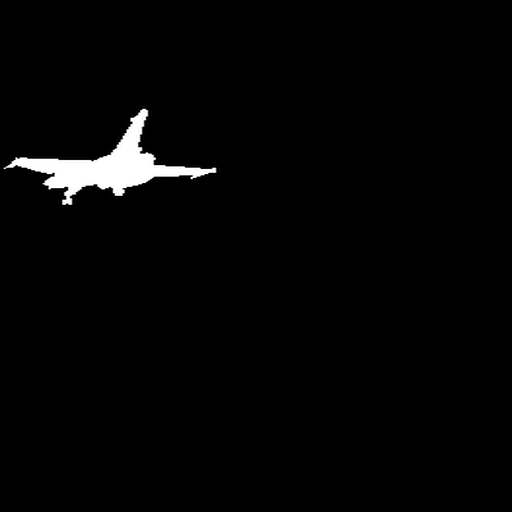} &
    \img{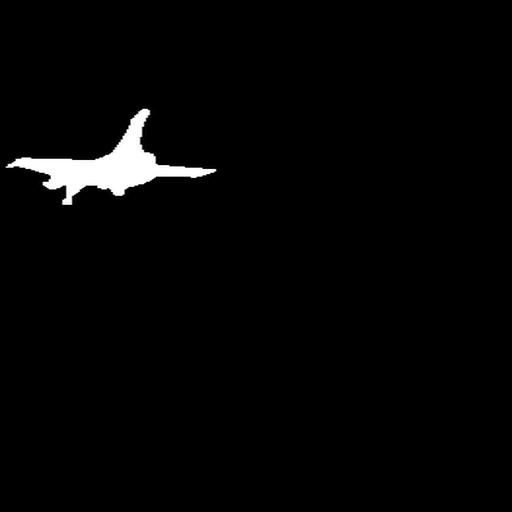} &
    \img{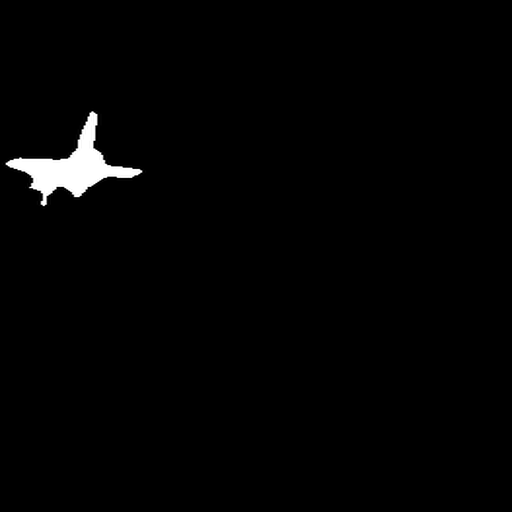} &
    \img{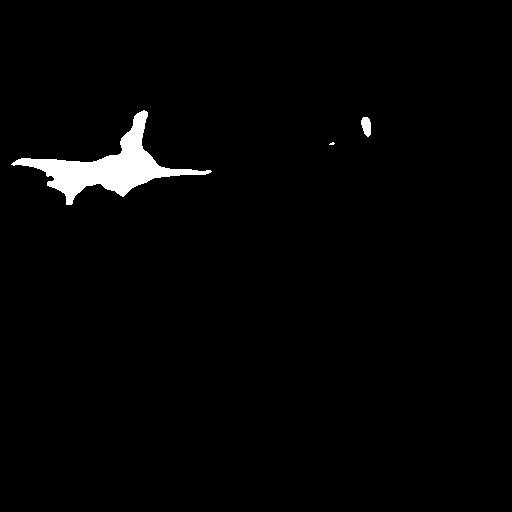} &
    \img{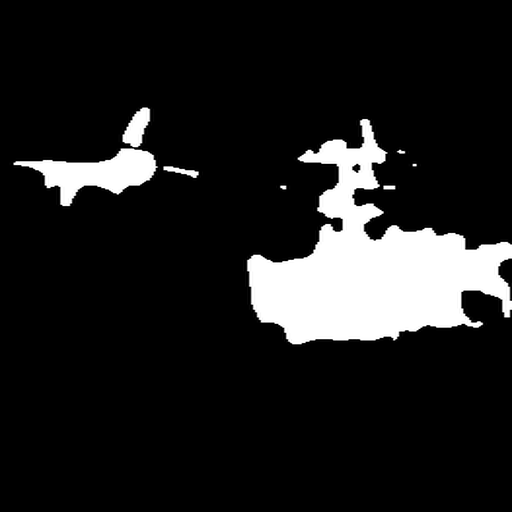} &
    \img{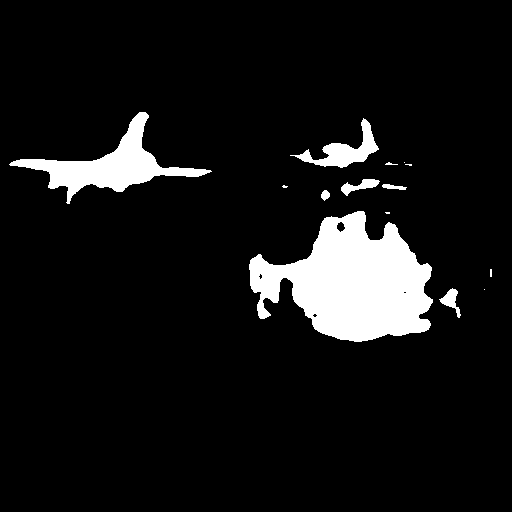} \\[4pt]

    \parbox[c]{\cellw}{\centering\fontsize{6}{6}\selectfont\textbf{\mbox{Image}}} &
    \parbox[c]{\cellw}{\centering\fontsize{6}{6}\selectfont\textbf{\mbox{GT}}} &
    \parbox[c]{\cellw}{\centering\fontsize{6}{6}\selectfont\textbf{\mbox{\method}}} &
    \parbox[c]{\cellw}{\centering\fontsize{6}{6}\selectfont\textbf{\mbox{VPD~\cite{zhao2023vpd}}}} &
    \parbox[c]{\cellw}{\centering\fontsize{6}{6}\selectfont\textbf{\mbox{DINO-v3~\cite{siméoni2025dinov3}}}} &
    \parbox[c]{\cellw}{\centering\fontsize{6}{6}\selectfont\textbf{\mbox{DHF~\cite{luo2024diffusionhyperfeaturessearchingtime}}}} &
    \parbox[c]{\cellw}{\centering\fontsize{6}{6}\selectfont\textbf{\mbox{DatasetDM~\cite{wu2023datasetdmsynthesizingdataperception}}}} \\

  \end{tabular}
  \caption{
    \textbf{Qualitative comparisons on DUTS examples.}
  }
  \label{fig:duts_qualitative_grid}
\end{figure}

\begin{figure}[t]
  \centering
  \setlength{\tabcolsep}{1pt}
  \renewcommand{\arraystretch}{0}
  \begin{tabular}{@{}*{7}{m{\cellw}}@{}}

    \img{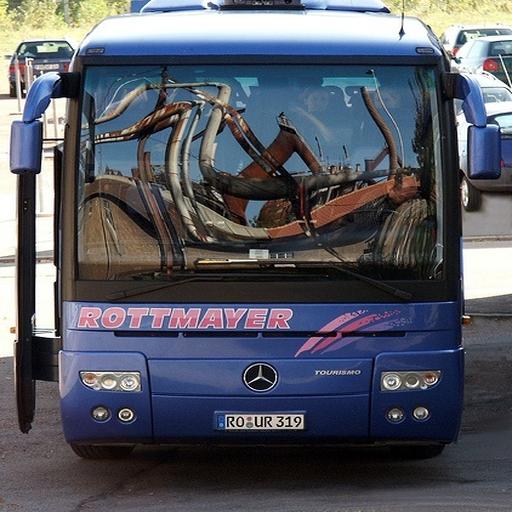} &
    \img{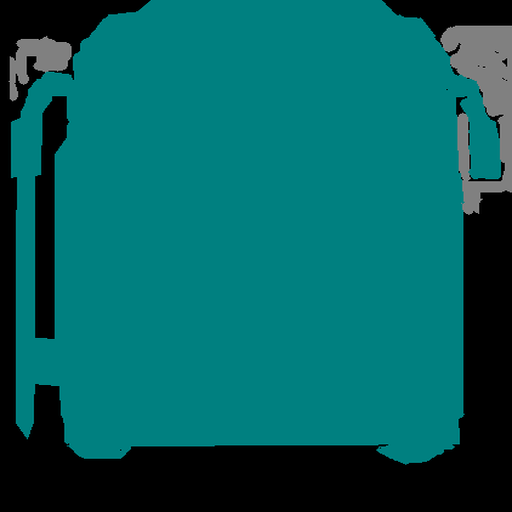} &
    \img{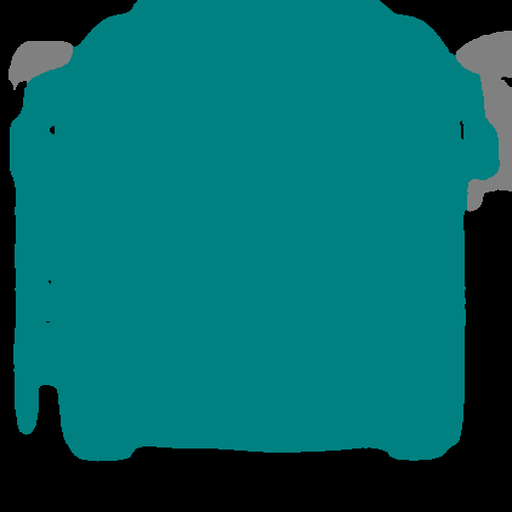} &
    \img{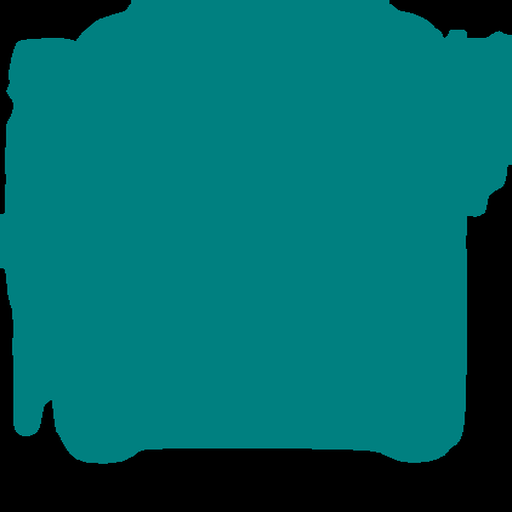} &
    \img{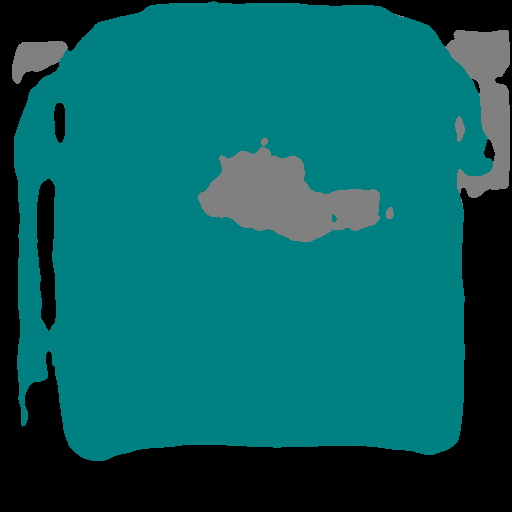} &
    \img{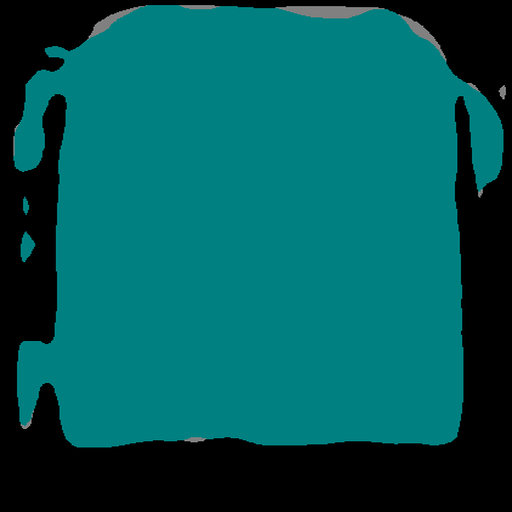} &
    \img{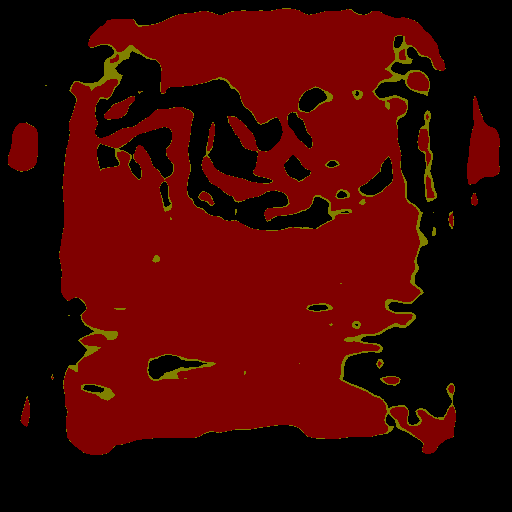} \\[-2pt]

    \img{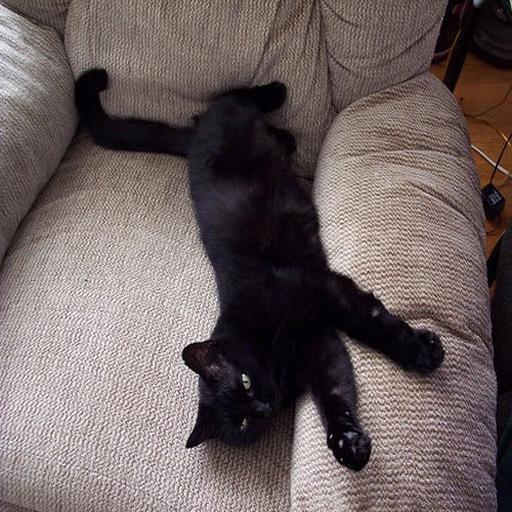} &
    \img{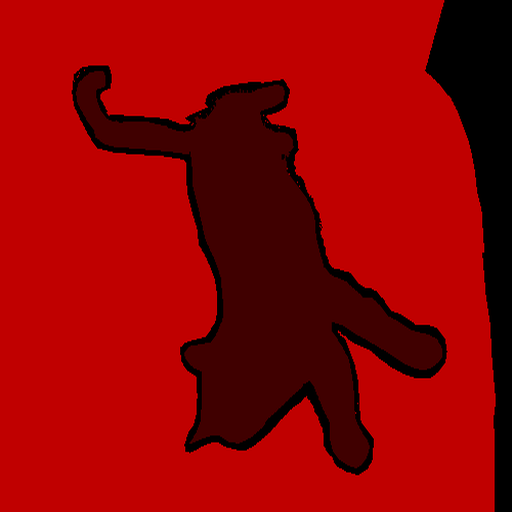} &
    \img{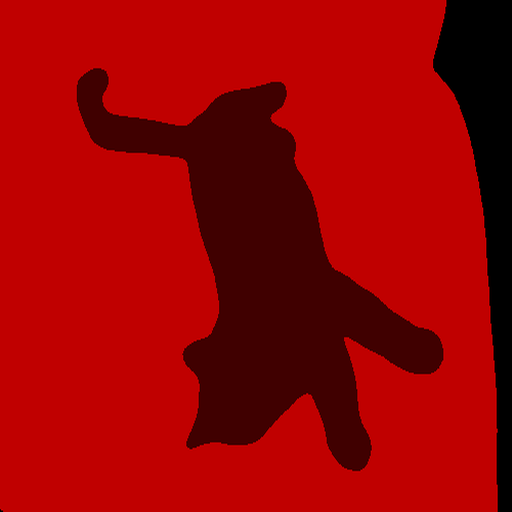} &
    \img{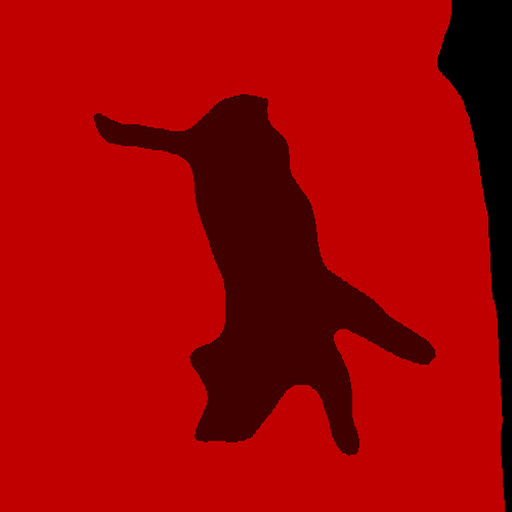} &
    \img{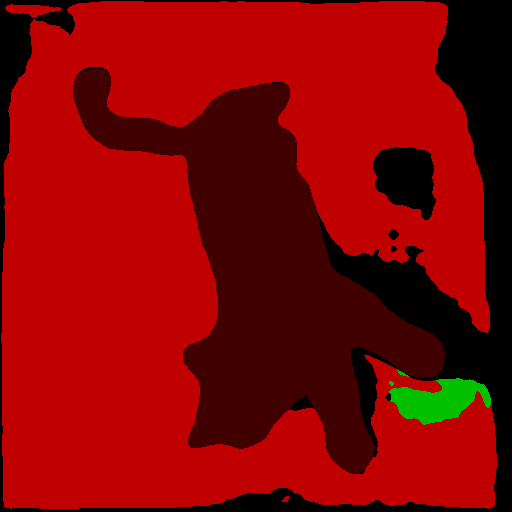} &
    \img{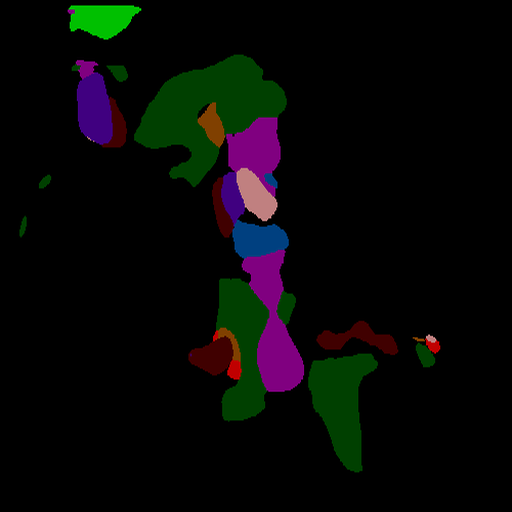} &
    \img{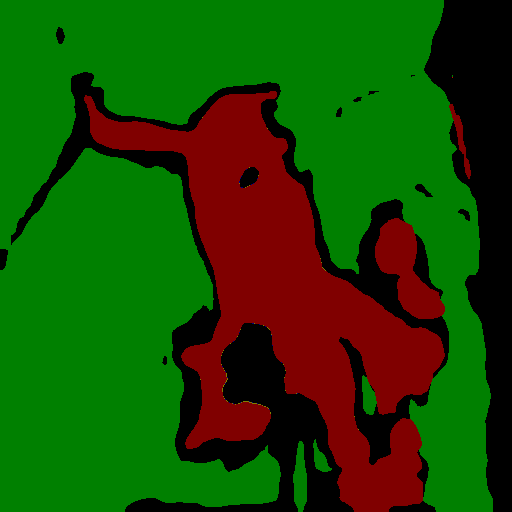} \\[-2pt]

    \img{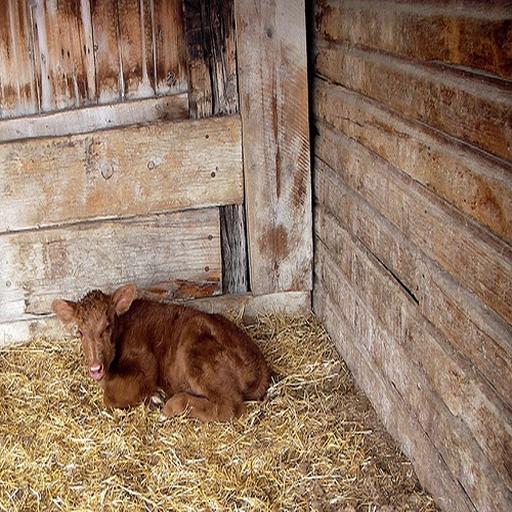} &
    \img{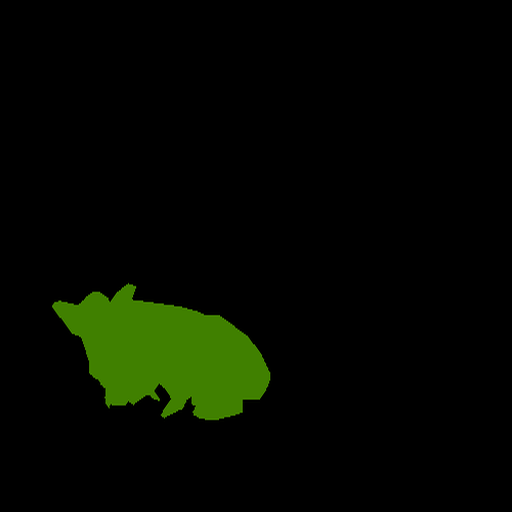} &
    \img{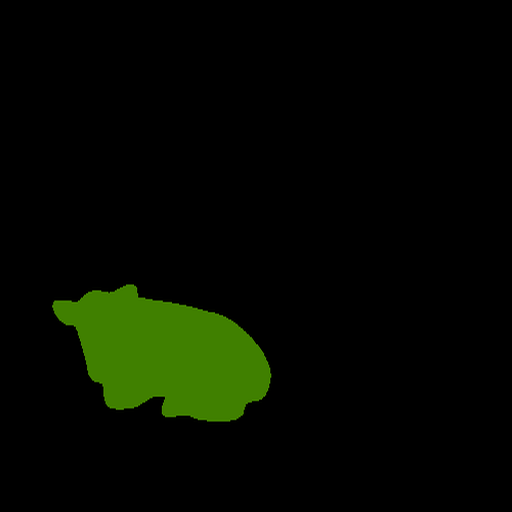} &
    \img{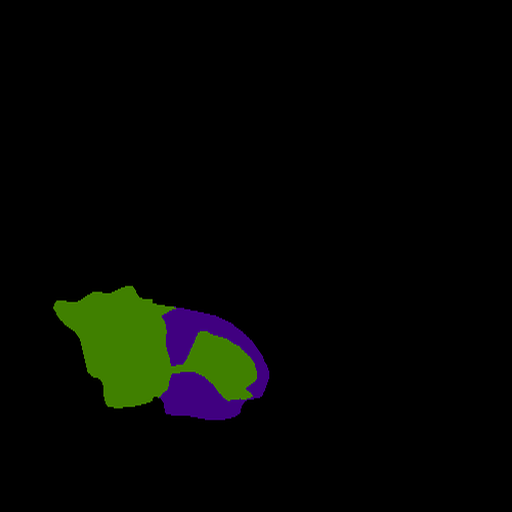} &
    \img{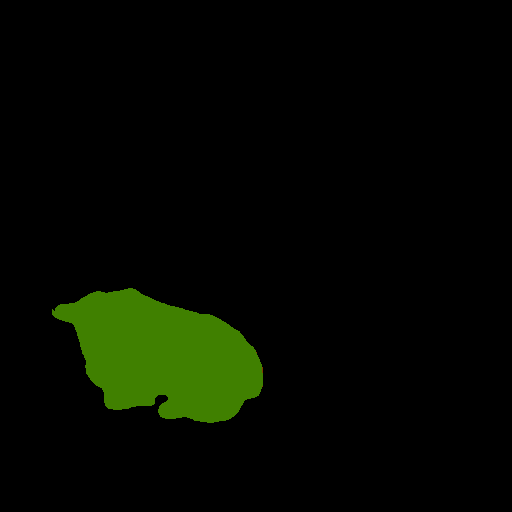} &
    \img{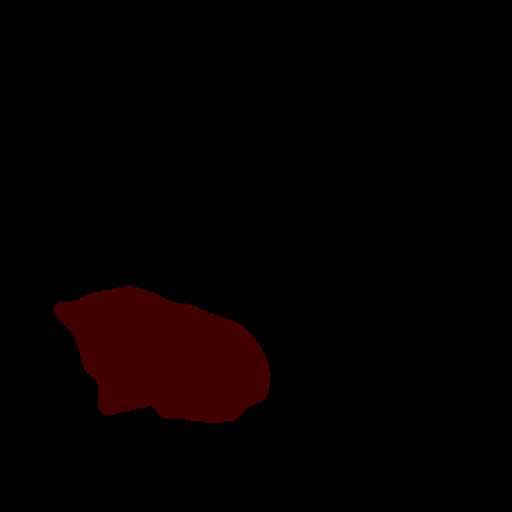} &
    \img{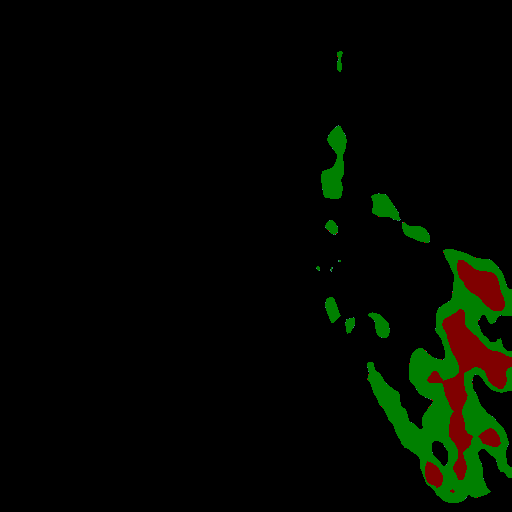} \\[2pt]

    \img{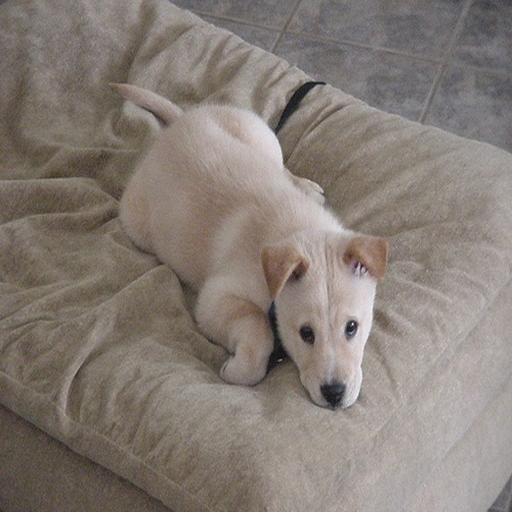} &
    \img{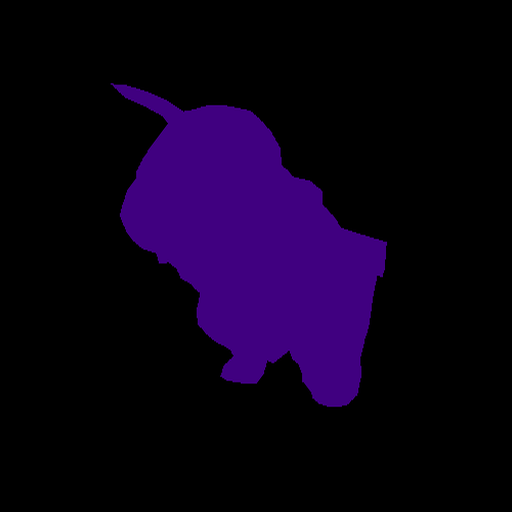} &
    \img{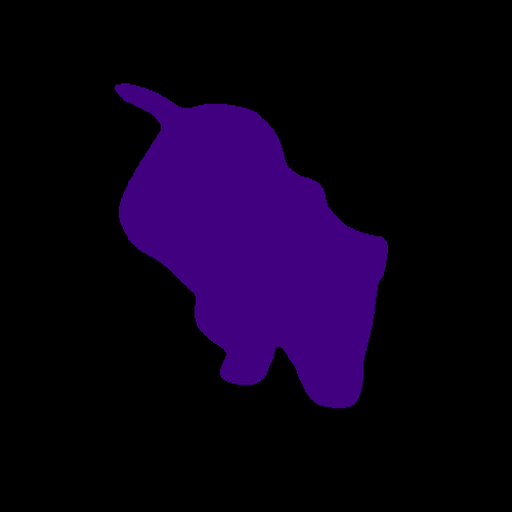} &
    \img{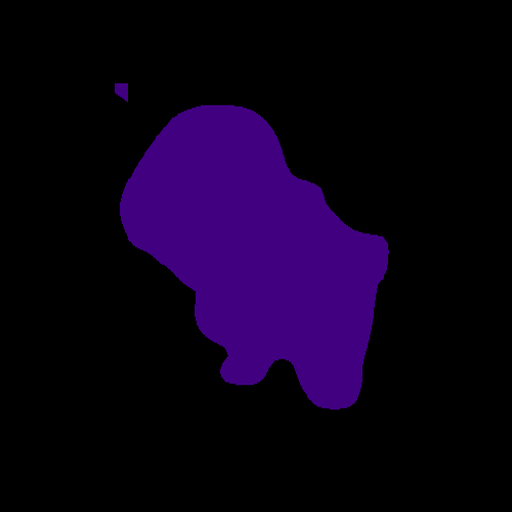} &
    \img{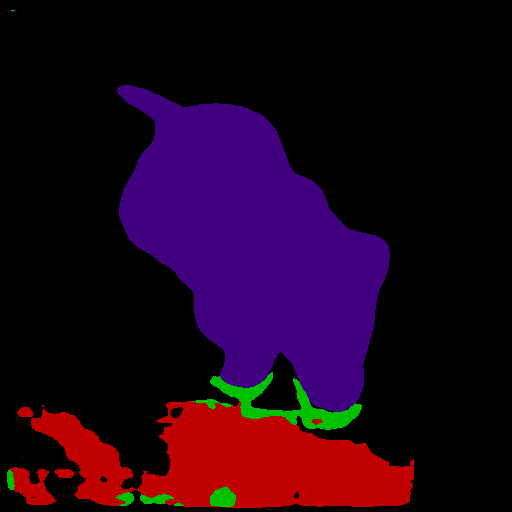} &
    \img{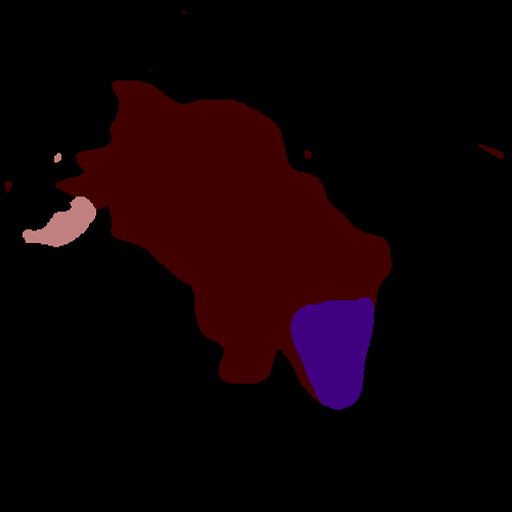} &
    \img{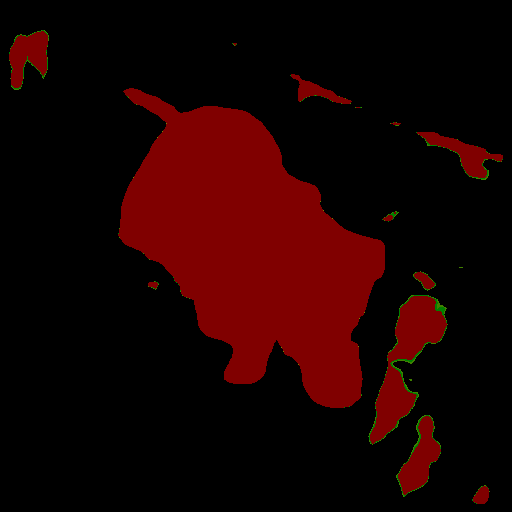} \\[2pt]

    \img{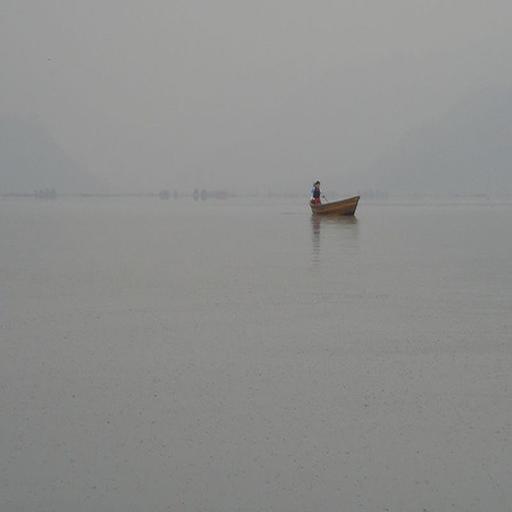} &
    \img{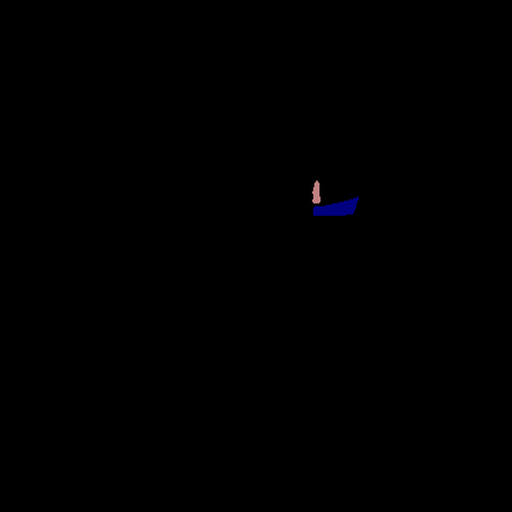} &
    \img{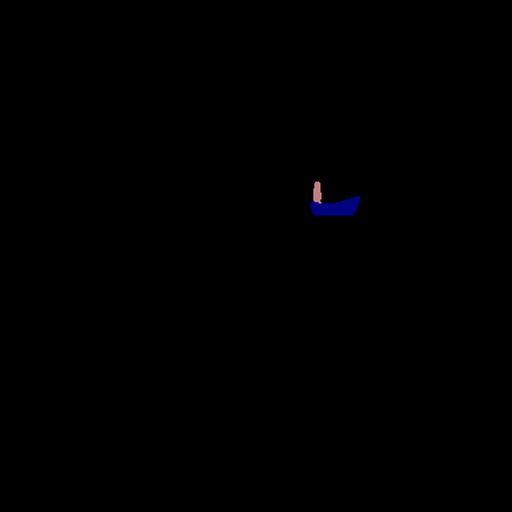} &
    \img{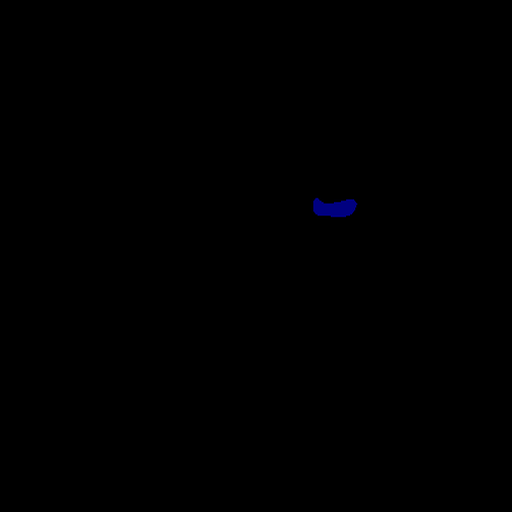} &
    \img{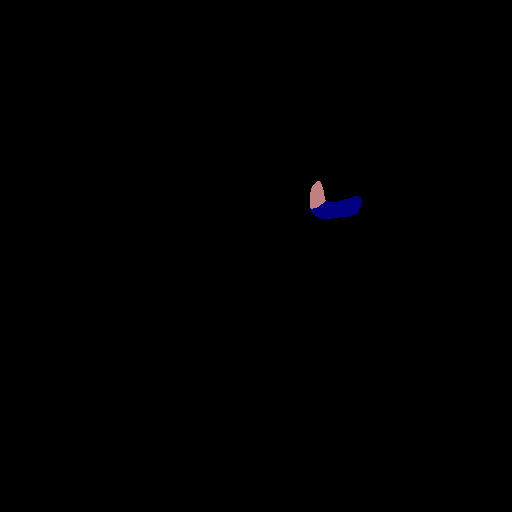} &
    \img{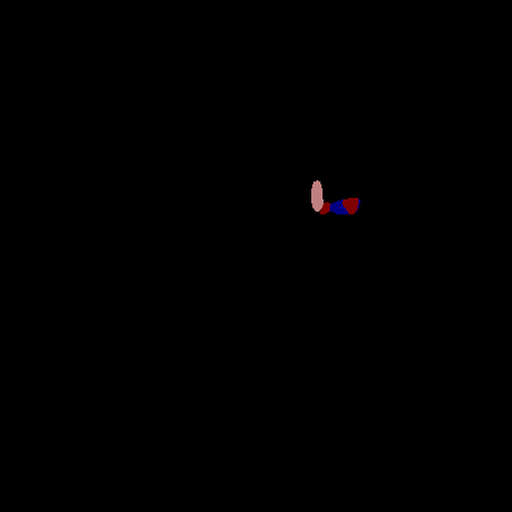} &
    \img{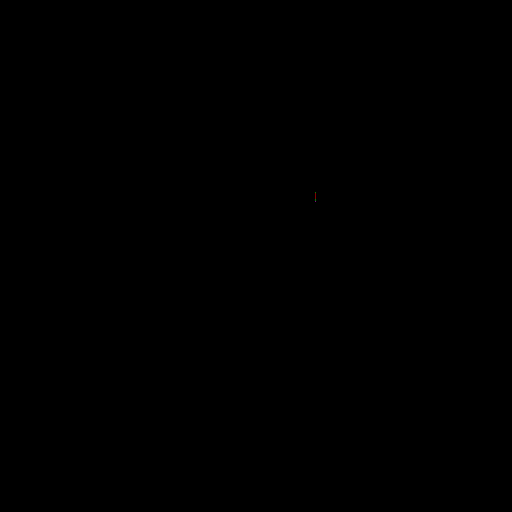} \\[2pt]

    \img{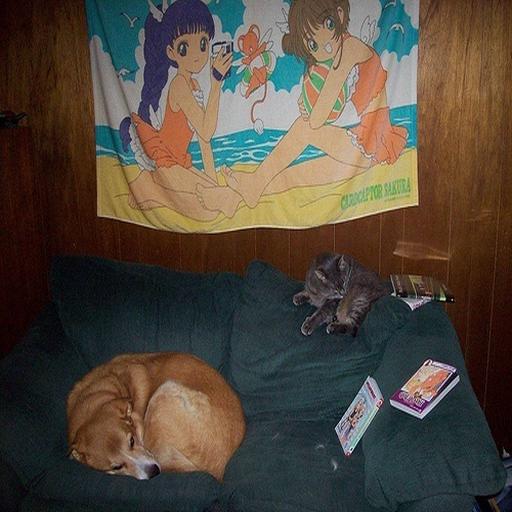} &
    \img{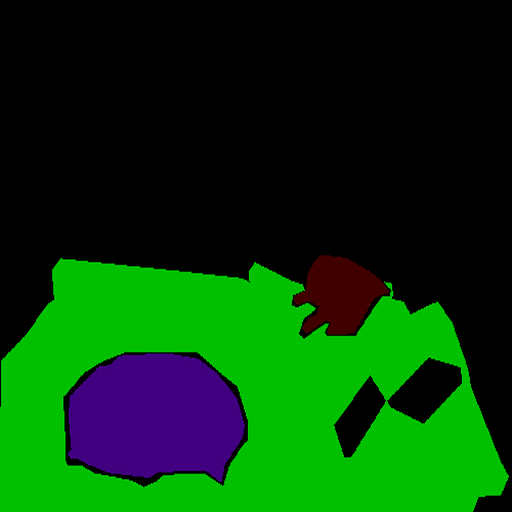} &
    \img{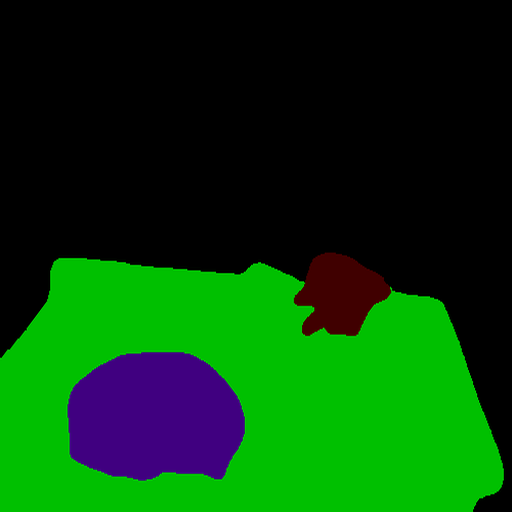} &
    \img{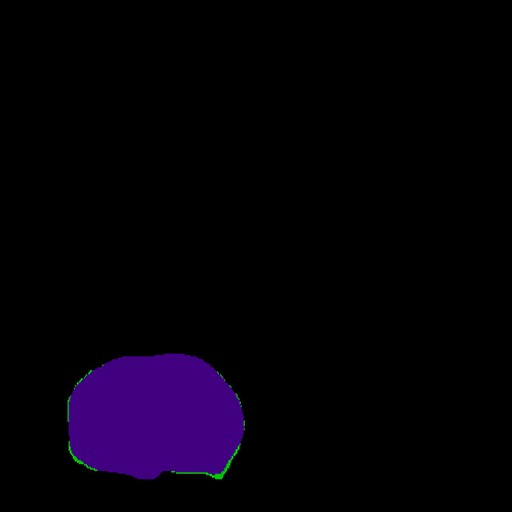} &
    \img{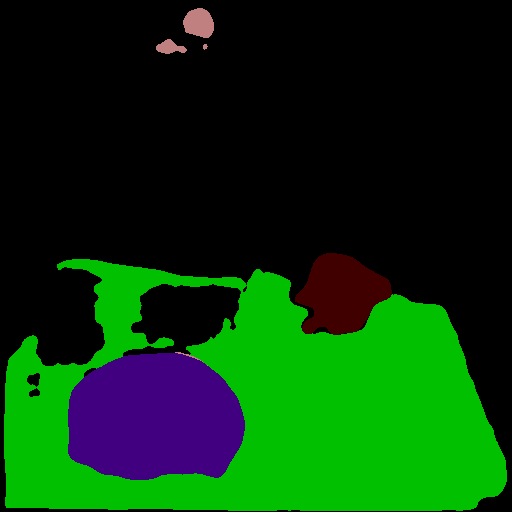} &
    \img{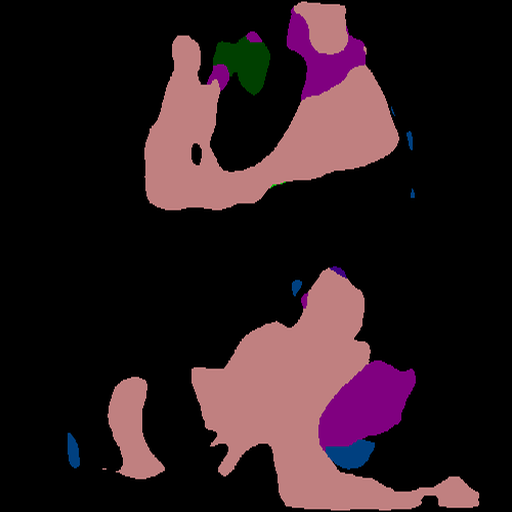} &
    \img{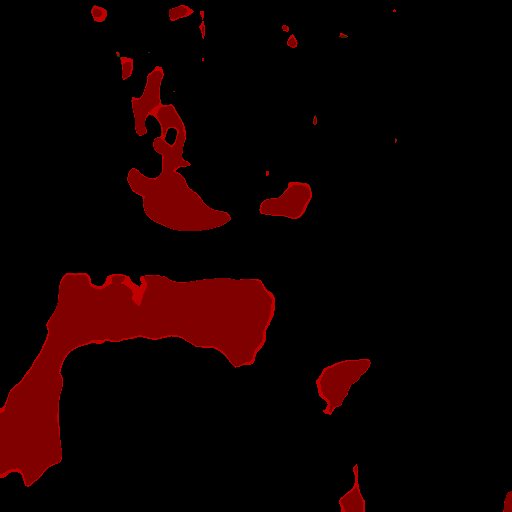} \\[4pt]

    \parbox[c]{\cellw}{\centering\fontsize{6}{6}\selectfont\textbf{\mbox{Image}}} &
    \parbox[c]{\cellw}{\centering\fontsize{6}{6}\selectfont\textbf{\mbox{GT}}} &
    \parbox[c]{\cellw}{\centering\fontsize{6}{6}\selectfont\textbf{\mbox{\method}}} &
    \parbox[c]{\cellw}{\centering\fontsize{6}{6}\selectfont\textbf{\mbox{VPD~\cite{zhao2023vpd}}}} &
    \parbox[c]{\cellw}{\centering\fontsize{6}{6}\selectfont\textbf{\mbox{DINO-v3~\cite{siméoni2025dinov3}}}} &
    \parbox[c]{\cellw}{\centering\fontsize{6}{6}\selectfont\textbf{\mbox{DHF~\cite{luo2024diffusionhyperfeaturessearchingtime}}}} &
    \parbox[c]{\cellw}{\centering\fontsize{6}{6}\selectfont\textbf{\mbox{DatasetDM~\cite{wu2023datasetdmsynthesizingdataperception}}}} \\

  \end{tabular}
  \caption{
    \textbf{Qualitative comparisons on Pascal VOC examples.}
  }
  \label{fig:pascal_qualitative_grid}
\end{figure}
\FloatBarrier

\end{document}